\pdfoutput=1

\documentclass[11pt]{article}

\usepackage[preprint]{EMNLP2023}

\usepackage{wrapfig}
\usepackage{microtype}
\usepackage{hyperref}
\usepackage{url}
\usepackage{booktabs}
\usepackage{subfig}
\usepackage{microtype}
\usepackage{graphicx}
\usepackage{booktabs} 
\usepackage{amsmath}
\usepackage{amssymb}
\usepackage{mathtools}
\usepackage{amsthm}
\usepackage{url}
\usepackage[utf8]{inputenc}
\usepackage[T1]{fontenc}
\usepackage{nicefrac}
\usepackage[table,xcdraw]{xcolor}
\usepackage{listings}
\usepackage{multirow}
\usepackage{colortbl}
\usepackage{xspace}
\usepackage{bbding}
\usepackage{wrapfig}
\usepackage{tcolorbox}
\tcbuselibrary{breakable}
\tcbuselibrary{skins}
\tcbuselibrary{most}
\usepackage{amssymb}
\usepackage{titletoc}
\usepackage{graphicx}

\definecolor{darkred}{rgb}{0.65,0.0,0.0}
\tcbset{
  aibox/.style={
    width=\linewidth,
    top=10pt,
    colback=white,
    colframe=black,
    colbacktitle=black,
    enhanced,
    breakable,
    center,
    fontupper=\linespread{0.92}\selectfont,
    segmentation style={draw=none},
    attach boxed title to top left={yshift=-0.1in,xshift=0.15in},
    boxed title style={boxrule=0pt,colframe=white,},
  }
}
\newtcolorbox{AIbox}[2][]{aibox,title=#2,#1}
\usepackage{tabularx}
\usepackage{longtable}
\usepackage{makecell}
\usepackage{pifont}
\usepackage{wasysym}
\usepackage{tikz}
\usepackage{float}
\usepackage{fancyvrb}
\usepackage{lipsum}
\usepackage{caption}
\usepackage{subcaption}
\usepackage{placeins}
\usepackage{enumitem}
\usepackage{epstopdf}

\usepackage{titletoc}
\usepackage{lineno}
\usepackage{latexsym}

\usepackage[T1]{fontenc}
\usepackage{newtxtext,newtxmath}
\usepackage[utf8]{inputenc}

\usepackage{microtype}
\usepackage{inconsolata}

\definecolor{darkred}{rgb}{0.5, 0.0, 0.0}
\definecolor{skyblue}{RGB}{203, 221, 245}
\definecolor{Secondary}{HTML}{DDEFD8}
\definecolor{gray1}{HTML}{F3F3F3}

\def\benchmark{\texttt{{M$^3$Exam}}\xspace}
\def\method{\texttt{{M$^3$Proctor}}\xspace}

\newcommand{\ie}{\textit{i}.\textit{e}.\xspace}

\newcommand{\eg}{\textit{e}.\textit{g}.\xspace}

\newcommand{\Yes}{\makebox[1em][c]{\textcolor{green!55!black}{\ding{51}}}}
\newcommand{\Part}{\makebox[1em][c]{%
  \textcolor{orange!85!black}{%
    \makebox[0pt][c]{\ding{51}}%
    \makebox[0pt][c]{\raisebox{0.30ex}{\scalebox{0.62}{\ding{55}}}}%
  }%
}}
\newcommand{\NA}{\makebox[1em][c]{\textcolor{cyan!70!black}{$\boldsymbol{\thicksim}$}}}
\newcommand{\No}{\makebox[1em][c]{\textcolor{red!75!black}{\ding{55}}}}
\definecolor{tintgreen}{RGB}{240, 248, 242}
\definecolor{tintred}{RGB}{252, 242, 242}
\definecolor{tintyellow}{RGB}{253, 250, 238}
\newcommand{\takeaway}[2]{
    \begin{tcolorbox}[
        colback=white!90!gray,     
        colframe=teal!60!black,     
        arc=5pt,                    
        boxsep=5pt,                 
        left=10pt,                  
        right=10pt,                 
        top=2pt,                    
        bottom=2pt,                 
        boxrule=0.8pt,              
        drop shadow=gray!50!white,  
        enhanced jigsaw             
    ]
    \vspace{-0.1cm}
        \noindent\textbf{\textbf{\textit{Takeaway #1:}}} #2
    \vspace{-0.1cm}
    \end{tcolorbox}
    \vspace{-0.1cm}
}

\title{\benchmark: Benchmarking Multimodal Memory for Realistic \\User-Agent Interactions}

\author{
  \textbf{Zhengjun Huang}\textsuperscript{1,7}\thanks{Equal contribution.}, \hspace{0.5mm}
  \textbf{Wenxuan Liu}\textsuperscript{2}\footnotemark[\value{footnote}], \hspace{0.5mm}
  \textbf{Zhoujin Tian}\textsuperscript{1}, \hspace{0.5mm}
  \textbf{Wei Chen}\textsuperscript{3,6}, \hspace{0.5mm}
  \textbf{Junle Chen}\textsuperscript{1}, \\
  \textbf{Yuqian Wu}\textsuperscript{3}, \hspace{0.5mm}
  \textbf{Fangyuan Zhang}\textsuperscript{4}, \hspace{0.5mm}
  \textbf{Qintian Guo}\textsuperscript{5$^\dag$}, \hspace{0.5mm}
  \textbf{Xiaofang Zhou}\textsuperscript{1}
  \\
  \textsuperscript{1}The Hong Kong University of Science and Technology,
  \textsuperscript{2}Beijing University of Chemical Technology \\
  \textsuperscript{3}The Hong Kong University of Science and Technology (Guangzhou)\\
  \textsuperscript{4}Harbin Institute of Technology (Shenzhen), 
  \textsuperscript{5}Beijing Institute of Technology (Zhuhai) \\
  \textsuperscript{6}Tencent Hy,
  \textsuperscript{7}Peng Cheng Laboratory \\
  \texttt{\small zhuangff@cse.ust.hk, qtguo@bit.edu.cn, zxf@cse.ust.hk} \\
}

\begin{document}
\maketitle

\begin{abstract}
Language agents are increasingly deployed over accumulating multimodal
information, yet existing benchmarks assume a human--human form with
sparse visuals and straightforward content, evaluating neither
reasoning over authentic multimodal file interaction nor the
interpretation of concealed user information. We therefore introduce \benchmark, a query-centric multimodal
conversational memory benchmark built on realistic user--agent
interaction, with multi-dimensional evaluation spanning cross-modal
grounding and implicit information inference.
Benchmarking MLLMs and memory systems reveals persistent gaps in cross-modal
grounding, cross-session reasoning, and the efficiency cost of
accumulating multimodal context. We further propose \method, a
multimodal memory method that detects query modality bias and
consumes raw visual sources only on demand, improving accuracy by
13\% while cutting index-construction time and retrieved 
tokens by over 70\%\footnote{Code and data are released at \url{https://anonymous.4open.science/r/M-3-Exam-128D}.}.
\end{abstract}

\section{Introduction}
\label{sec:intro}


As language agents~\cite{openai2026,anthropic2026,google2026gemini31pro} move toward real-world deployment, multimodal personal assistance~\cite{cheng2025higher} has emerged as a representative scenario that demands multi-turn interactions. This shift elevates the central challenge from per-turn perception to managing a long-term, heterogeneous multimodal history. A fundamental capability for such agents is durable, well-structured \emph{multimodal memory}: the ability to store, retrieve, and reason over a fragmented trail of text, images, and documents accumulated across multiple sessions~\cite{zhang2025survey}. Because multi-session conversations are the primary medium through which this rich memory is naturally formed and utilized, they serve as the ideal testbed for evaluating an agent's long-term multimodal capabilities.

\begin{figure}[t]
  \centering
  \includegraphics[width=0.88\linewidth]{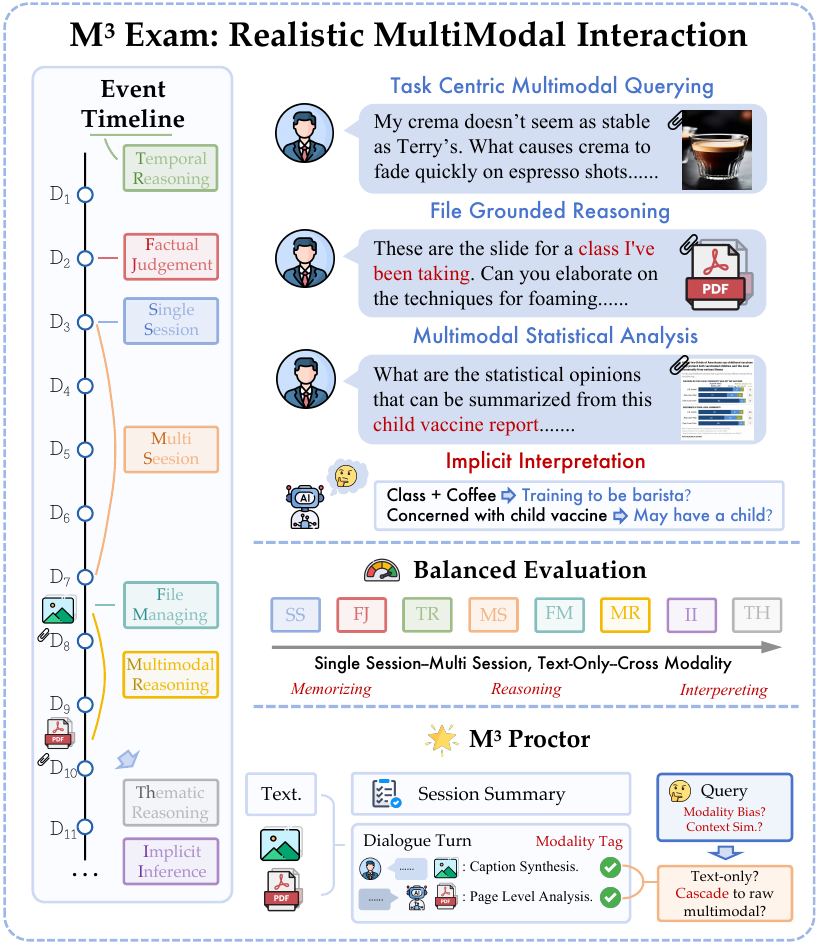}
  \caption{\textbf{Overview of \benchmark.}}
  \label{fig:teaser}
  \vspace{-4mm}
\end{figure}

\begin{table*}[t]
\centering
\caption{Comparison of \benchmark with representative
conversational-memory benchmarks.
\textbf{A.\,Round} = average user--assistant
rounds per session; \textbf{A.\,File} = average images and document
files per session.
\textbf{MR} = multimodal reasoning over visual evidence;
\textbf{SA} = statistical analysis of multimodal graphs;
\textbf{FM} = file management and retrieval;
\textbf{FR} = file reasoning over document content.
\textbf{II} = implicit information inference;
\textbf{TH} = domain-specific thematic reasoning.
\Yes = supported; \Part = partial / restricted; \No = not supported; \NA = N/A.}
\vspace{-1mm}
\small
\newcolumntype{Y}{>{\centering\arraybackslash}X}
\newcolumntype{L}{>{\centering\arraybackslash}p{5.0cm}}
\setlength{\tabcolsep}{3.5pt}
\renewcommand{\arraystretch}{0.95}
\begin{tabularx}{\linewidth}{L|YY|YYYY|YY}
\toprule
\multirow{2}{*}{\centering\textbf{Benchmark}}
 & \multicolumn{2}{c|}{\textbf{Memorizing}}
 & \multicolumn{4}{c|}{\textbf{Reasoning}}
 & \multicolumn{2}{c}{\textbf{Interpreting}} \\
\cmidrule(lr){2-3}\cmidrule(lr){4-7}\cmidrule(lr){8-9}
 & \textbf{A.\,Round} & \textbf{A.\,File}
 & \textbf{MR} & \textbf{SA} & \textbf{FM} & \textbf{FR}
 & \textbf{II} & \textbf{TH} \\
\midrule
LongMemEval~\citep{wu2025longmemeval}     & 5.2  & \NA   & \No   & \No   & \No   & \No   & \No   & \No   \\
MemoryArena~\citep{he2026memoryarena}     & 6.9  & \NA   & \No   & \No   & \No   & \No   & \No   & \No   \\
MemoryAgentBench~\citep{hu2025evaluating} & 9.6  & \NA   & \No   & \No   & \No   & \No   & \Part & \Part \\
LoCoMo~\citep{maharana2024locomo}         & 10.8 & 3.4  & \Part & \No   & \No   & \No   & \No   & \No   \\
MMDialog~\citep{feng2023mmdialog}         & 4.6  & 2.6  & \Part & \No   & \Part & \No   & \No   & \No   \\
MMRC~\citep{xue2025mmrc}                  & 12.9 & 2.9  & \Yes  & \No   & \Yes  & \No   & \No   & \No   \\
Mem-Gallery~\citep{bei2026mem}            & 16.5 & 4.2  & \Yes  & \No   & \Yes  & \No   & \No   & \No   \\
\midrule
\rowcolor{skyblue!50}
\textbf{\benchmark (Ours)}
 & \textbf{12.7} & \textbf{7.5}
 & \Yes & \Yes & \Yes & \Yes
 & \Yes & \Yes \\
\bottomrule
\end{tabularx}
\label{tab:bench-compare}
\vspace{-4mm}
\end{table*}


As illustrated in Figure~\ref{fig:teaser}, realistic multimodal memory unfolds organically over an extended timeline: a user might share espresso photos in one session, upload class slides in another, and later analyze a child vaccine report. Answering such queries mirrors how humans recall and reason over past information: one must \emph{memorize} what was encountered, \emph{reason} over it to connect disparate pieces, and \emph{interpret} information that is implied but never stated outright~\cite{wu2025longmemeval}. A faithful multimodal memory benchmark must therefore capture three corresponding dimensions of difficulty. First, \textit{content complexity} requires agents to persistently track and manage heterogeneous artifacts (\eg, dialogues, images, PDFs) within a growing, multi-session history. Second, \textit{reasoning complexity} for cross-modal inference, like connecting a coffee's visual state to a past PDF. Finally, \textit{intent complexity} to deduce unstated contexts, such as inferring the user is a barista or parent from history.

Existing benchmarks, however, fall short on these evaluation dimensions
(Table~\ref{tab:bench-compare}). Early long-term conversational-memory
suites are text-only and omit visual modalities altogether
\cite{wu2025longmemeval,he2026memoryarena}; subsequent multimodal benchmarks add images but treat them as static snapshots and ignore document-level artifacts \cite{maharana2024locomo,bei2026mem},
while document-QA benchmarks probe rich multimodal reasoning but
restrict it to a single document within a single turn
\cite{tanaka2023slidevqa,ma2024mmlongbench}. As Table~\ref{tab:bench-compare} shows, no prior benchmark jointly stresses realistic multimodal memorizing, reasoning, and interpreting, while nearly all
assume user information is fully stated, overlooking the
\emph{implicit-inference} regime that requires indirect context interpretation.

To close this gap, we introduce \benchmark, a query-centric
\underline{\textbf{m}}ulti\underline{\textbf{m}}odal conversational \underline{\textbf{m}}emory benchmark
specifically targeting realistic multimodal user--agent interaction. 
\benchmark comprises $239$ realistic multi-session conversations spanning
$15$ persona scenarios ($3{,}025$ rounds with $1{,}799$ multimodal
artifacts) paired with $5{,}150$ evaluation questions. Given a query, 
an agent must analyze the accumulating multimodal
history and retrieve the relevant dialogue sessions or multimodal
artifacts to answer it. These demands are reflected in a
reasoning-heavy, balanced question bank that, beyond standard retrieval and
multi-hop questions, adds two types absent from prior work: Thematic
Reasoning (\textsc{TH}), which requires domain knowledge implicit in the
user's context, and Implicit Inference (\textsc{II}), whose answer
hinges on information the history implies but never states outright.
Benchmarking the latest Multimodal Large Language Models (MLLMs) and frontier agent memory systems on \benchmark{} shows they struggle with cross-modal reasoning and implicit interpretation.

\begin{figure*}[t]
  \centering
  \includegraphics[width=\linewidth]{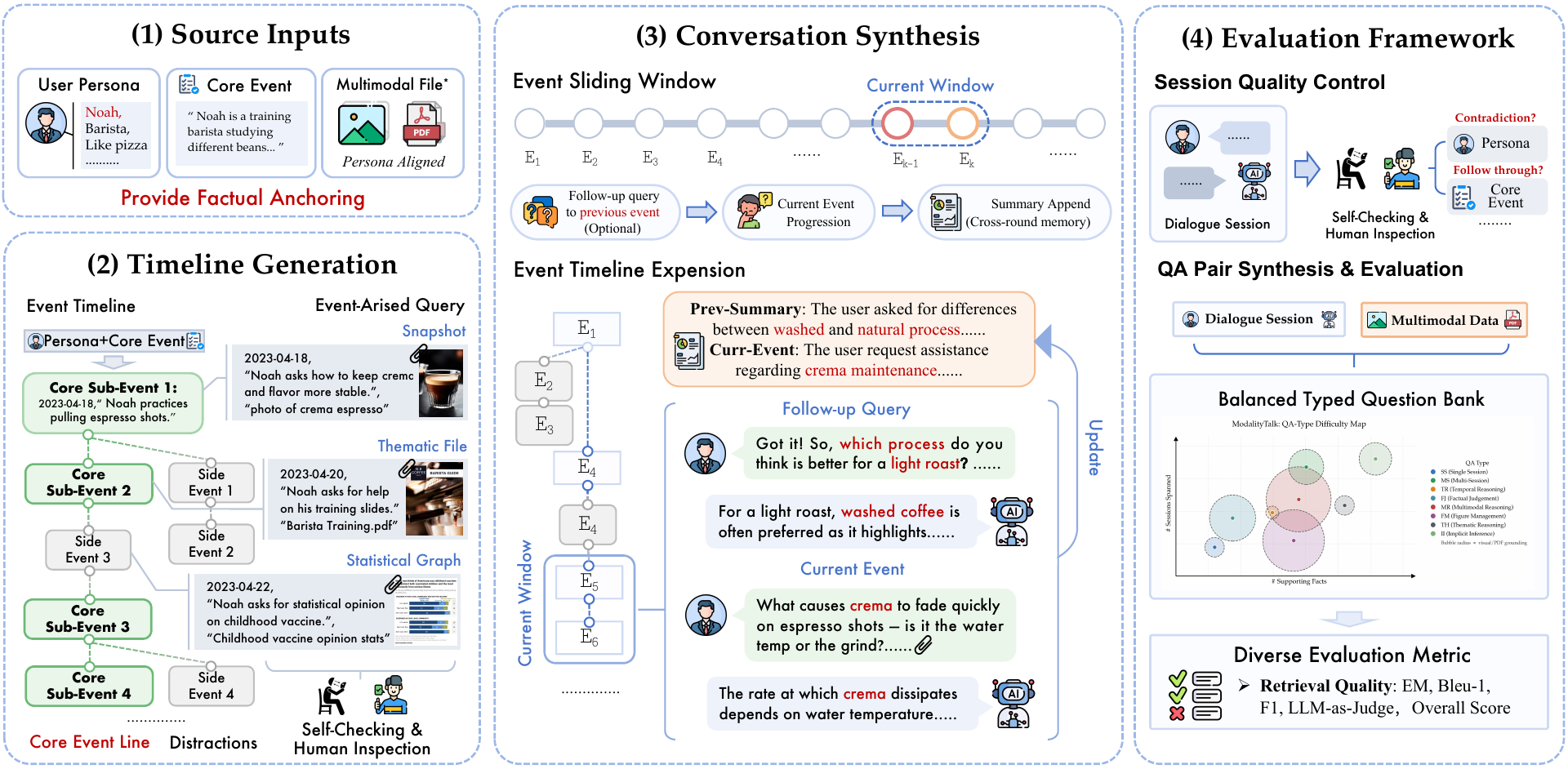}
  \caption{Overall pipeline of \benchmark, designed to evaluate multimodal memory ability in realistic scenarios.}
  \label{fig:overview}
  \vspace{-2mm}
\end{figure*}

We further turn our findings into a method. Experiments show that
existing multimodal memory systems are modality-agnostic. They retrieve
and inject raw visual sources indiscriminately, ignoring whether a
question actually needs them, which both buries the decisive evidence
among irrelevant visuals and inflates the per-query token budget.
We therefore propose \method, a modality-aware memory method that
detects \emph{modality bias} within queries and consumes raw visual
sources only on demand via (1) bias detection, (2) modality-aware
re-ranking and (3) a cost-aware cascade that
escalates to raw visuals only when text proves insufficient. On \benchmark, \method improves accuracy by over 13\% with 70\% less tokens
while cutting index-construction time by 80\% over existing systems. Overall, our main contributions are threefold:
\begin{itemize}[leftmargin=*,itemsep=0.2em]
\item \emph{Proposed Novelty Benchmark.} We introduce \benchmark, a realistic multimodal 
memory benchmark spanning 15 persona scenarios of 239 multi-session
conversations with rich cross-modality reasoning and implicit interpretation evaluations.
\item \emph{Benchmarking \& Findings.} Extensive benchmarking and in-depth analysis are conducted across latest multimodal models and agent memory systems, uncovering their limitations
in visual grounding, complex implicit reasoning, and the efficiency of accumulating multimodal contexts.
\item \emph{Proposed Novelty Baseline.} We propose a novel \method, a modality-aware multimodal memory method that detects
query modality bias and enables modality cascade to consume raw visual sources only on demand. \method improves accuracy by 13\% with significantly less cost.
\end{itemize}

\section{\benchmark: Agent Benchmark}
\label{sec:benchmark}

This section describes \benchmark{} in detail
(Figure~\ref{fig:overview}). We first formalize the task
(\S\ref{sec:taskformulation}), then describe how each instance is built
from the source inputs through timeline generation, multi-turn
conversation synthesis, and question-bank construction
(\S\ref{sec:benchmarkconstruction}), with quality control applied at
every stage. Finally, we review the
resulting benchmark statistics (\S\ref{sec:datasetstatistics}).

\subsection{Task Formulation}
\label{sec:taskformulation}

\benchmark{} casts the agent as a respondent over a multimodal memory
accumulated from a user's past conversations. 
Each instance pairs the resulting
multi-session conversation $\mathcal{D}=(D_1,\dots,D_L)$ with a query
$x$. The sessions are ordered in time and interleave dialogue turns with
attached images and document pages, forming a history $\mathcal{H}$.
Given $\mathcal{H}$ and $x$, the agent produces an answer
$\hat{a}=f(x,\mathcal{H})$, scored against the gold answer $a_0$. Each query is annotated with the
supporting-fact rounds $\mathcal{F}_q$ it depends on, which may span
multiple sessions and modalities. Thus, answering requires
\emph{retrieving} the right evidence from $\mathcal{H}$, \emph{grounding}
it across text, images, and documents, and \emph{interpreting} the
implicit information required by the question.

\subsection{Benchmark Construction}
\label{sec:benchmarkconstruction}

\paragraph{Source Inputs.}
\label{sec:sourceinputs}
Each instance is synthesized from a triple $(\mathcal{P},\mathcal{C},\mathcal{F})$:
\emph{Persona} $\mathcal{P}$, a short biography (name, profession,
interests) that anchors and steers generation; \emph{Core event}
$\mathcal{C}$, a one-paragraph narrative arc the persona pursues over
time; and a \emph{multimodal file pool} $\mathcal{F}$ of charts,
photographs, and PDF documents, provided locally or retrieved by
keyword. Construction proceeds in the following three stages, each followed by a
quality-control inspection.

\paragraph{Timeline Construction.}
\label{sec:timelineconstruction}
Conditioned on $(\mathcal{P},\mathcal{C})$, a text large language model (LLM) generates a
chronologically ordered \emph{core-event line} of $N$ events
$\{e_1,\dots,e_N\}$, each a record $e_i=(\textsc{idx}_i,\textsc{desc}_i,
\textsc{query}_i,\textsc{time}_i,\textsc{kw}_i)$ that pairs a life event
with the visual-grounded query the persona will raise, a strictly
increasing date, and an image-search keyword. To mirror how real users
drift off-topic, we additionally generate $M$ \emph{distractor side
events} from domains orthogonal to $\mathcal{C}$ (errands, hobbies,
casual chatter), which are merged into the core line by date and
renumbered; these stress evidence retrieval amid noise and supply hard
negatives for the find-matching and implicit-inference questions. An LLM
self-check then audits persona violations, core-event drift, and
internal contradictions, repairing only flagged events, before a human
spot-check clears the timeline for synthesis.

\begin{figure*}[t!]
  \centering
  \includegraphics[width=\linewidth]{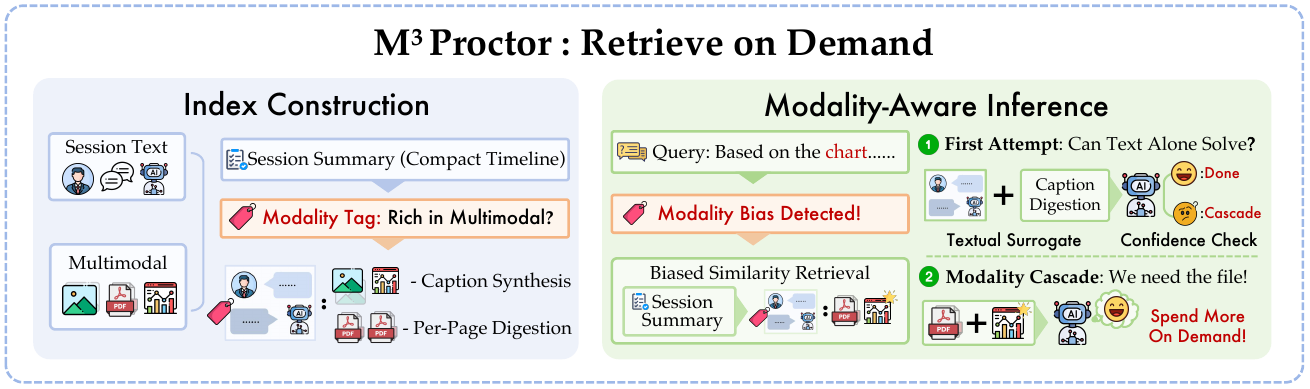}
  \vspace{-6mm}
  \caption{Overview of \method, designed to detect modality bias within queries and retrieve only on demand.}
  \label{fig:method}
  \vspace{-4mm}
\end{figure*}

\paragraph{Multi-Turn Conversation Synthesis.}
The timeline is extended into a conversation session by walking an \emph{event sliding
window} that advances one event per round: round $k$ exposes a
$\delta$-event window $\mathcal{W}_k=(e_{p_k-\delta},e_{p_k})$ whose trailing
event lets the user simulator issue an optional follow-up while the
leading event drives the current query over its freshly attached images,
delivered cumulatively so the assistant at round $k$ sees all images up
to $k$. To carry memory across long horizon, each exchange
is condensed into a key summary $\sigma_k$ appended to a running memory
$S_k=S_{k-1}\cup\{\sigma_k\}$, on which the user simulator $\pi_{\text{user}}$
conditions the next round:
\begin{equation}
  r_{k+1} \sim \pi_{\text{user}}\bigl(\,\cdot \mid \mathcal{W}_{k+1},\, S_k,\, \mathcal{P}\,\bigr).
\end{equation}
The round stream is then partitioned into sessions
$\mathcal{D}=(D_1,\dots,D_L)$. A two-stage session quality control then repairs persona drift
and contradictions: per-chunk audit scans local
windows of rounds, while cross-chunk audit catches errors
spanning chunk boundaries, with a final expert inspection.

\paragraph{Question Bank Synthesis.}
\label{sec:eval-framework}
From the finalized dataset we synthesize a typed question bank by
sampling context windows and prompting an LLM to emit items, each a
tuple $q=(x,A,\mathcal{F}_q,\tau)$ with question $x$, an ordered answer
list $A=(a_0,\dots,a_m)$ whose head $a_0$ is the gold and whose tail
gives progressively coarser alternatives for graded scoring,
supporting-fact rounds $\mathcal{F}_q$, and type $\tau$. The eight types
span three capabilities: \emph{memorizing}---single-session retrieval
(\textsc{ss}), file management (\textsc{fm}), and factual judgment
(\textsc{fj}); \emph{reasoning}---multi-session analysis (\textsc{ms}),
multimodal reasoning (\textsc{mr}), and temporal reasoning
(\textsc{tr}); and \emph{interpreting}---implicit inference
(\textsc{ii}) and thematic reasoning (\textsc{th}). Quality validation
pass any item that is malformed or cites evidence outside the
shown context, so every retained question is answerable from the
evidence it cites. Per-type definitions are in
Appendix~\ref{sec:appendix-examples}.

\subsection{Dataset Statistics}
\label{sec:datasetstatistics}

\benchmark{} comprises $239$ multi-session conversations spanning $15$
persona scenarios, totaling $3{,}025$ user--assistant rounds paired with
$5{,}150$ evaluation questions. The conversations are long-horizon with each
persona spaning $14$ to $18$ sessions with a mean of $12.7$ rounds per
session, and the conversations together accumulate $1{,}799$ multimodal
artifacts. The question bank is reasoning-heavy and balanced across personas,
deliberately weighted toward cases that stress multimodal memory:
$1{,}796$ questions ($34.9\%$) are cross-modal (\textsc{fm}, \textsc{mr}),
and $769$ ($14.9\%$) demand interpretation (\textsc{th}, \textsc{ii}) from
unstated context. Breakdown details are shown in
Appendix~\ref{app:detaildata}.

\section{\method: Proxy Baseline}
\label{sec:ourmethod}

Prior multimodal memory methods struggle with the complex, multi-session nature of \benchmark because they are inherently modality-agnostic. Indiscriminately injecting raw visual sources buries decisive evidence and unnecessarily inflates token budgets for queries resolvable by text alone. To facilitate future research, we introduce \method (Figure~\ref{fig:method}), a baseline that elevates modality bias (\ie, the specific modality on which a query depends) to a first-class signal. During index construction (\S\ref{sec:method-index}), it projects raw modalities into searchable textual surrogates equipped with modality tags. For modality-aware inference (\S\ref{sec:method-infer}), it dynamically detects query bias and employs a cascaded strategy, escalating to raw visual sources only on demand.

\subsection{Multimodal Index Construction}
\label{sec:method-index}

To keep the memory searchable by a text retriever while preserving the evidence
locked inside images and documents, we project each raw modality into a
textual \emph{surrogate} attached to the round that introduced it:
images carry a caption, informative visuals such as charts and tables additionally
carry a number-rich transcription of their axes, series, and values, and
each document page pairs its text layer with a one-sentence digest. 
Each chunk also carries a modality tag
$\mathbb{1}_m(c)\in\{0,1\}$ over $\mathcal{M}=\{\mathrm{I},\mathrm{P},\mathrm{C}\}$
(image, document, chart), read from the surrogate rather than the
pixels, which is the zero-cost hook the downstream modules act on. We further attach a compact session-summary chunk to each session which acts as a
miniature timeline that preserves the cross-session signal otherwise
fragmented by round-level chunking. Together, the text surrogates and modality tags establish the foundation for the answering cascade (Section\ref{sec:method-cascade}).
Because a chart's surrogate already encodes its underlying quantities,
many figure-reading questions are more answerable from textual content alone, allowing
the cascade to forgo re-rendering the raw source unless the modality tag
and bias signal indicate otherwise.

\subsection{Modality-Aware Inference}
\label{sec:method-infer}

At inference time, \method exploits the indexed modality tags and the
question's \emph{modality bias} to determine, for each retrieved chunk,
whether its low-cost textual surrogate suffices or the raw source must
be consulted. This process carries out in three coupled steps: \method first detects the question's modality bias and applies it to
re-rank the retrieved evidence, surfacing the most relevant dialogues
irrespective of modality. It then attempts an answer from the textual
surrogates of the dialogues alone, resorting to the raw visual or document sources only
when this text-surrogate attempt is deemed unreliable.

\paragraph{Bias detection and bias-aware re-ranking.}
We make modality demand explicit by representing the bias of a question
as a binary vector indexed by $\mathcal{M}$,
\begin{equation}
b(q) = \bigl(b_m(q)\bigr)_{m\in\mathcal{M}} \in \{0,1\}^{|\mathcal{M}|},
\label{eq:bias}
\end{equation}
where $b_m(q)=1$ iff answering $q$ requires inspecting a source of
modality $m$. A single instruction-tuned LLM call returns $b(q)$, and
the result is cached so this overhead is paid at most once per
question---far cheaper than a recurring multimodal forward pass.
Crucially, $b(q)$ is a \emph{shared} signal that both re-ranking and
answering consume, keeping modality decisions consistent end to end.
Re-ranking is its first use. Dense retrieval ranks chunks by semantic
similarity alone, which can bury a round that holds the decisive image
but is only moderately similar to $q$; we therefore use $b(q)$ to
correct the ranking \emph{within} the candidate pool. We first retrieve
a pool of the chunks with highest cosine similarity $\cos(q,c)$, then
re-rank each candidate by a bias-modulated score as follows
\begin{equation}
s(q,c) = \cos(q,c)
       + \sum_{m\in\mathcal{M}} \lambda_m\, b_m(q)\,\mathbb{1}_m(c),
\label{eq:rerank}
\end{equation}
and keep the $k$ highest-scoring chunks. The bonus $\lambda_m$ applies
only when the question is biased toward $m$ ($b_m(q)=1$) while the
chunk carries that modality ($\mathbb{1}_m(c)=1$); we set
$\lambda_{\mathrm{P}}\!>\!\lambda_{\mathrm{I}}\!>\!\lambda_{\mathrm{C}}$
since document-bearing chunks are rarest and most diagnostic, and add a
lightweight diversity constraint so the retained set covers the
session-summary view and spans more than one session, protecting
multi-session and temporal questions
(details in Appendix~\ref{sec:appendix-config}). Because the bonuses are
small relative to the cosine range, re-ranking breaks ties in favor of
the right modality but cannot promote an unrelated chunk, and it
operates entirely on index-time tags and surrogates, so improving the
modality composition of the top-$k$ incurs \emph{no} multimodal
cost with raw sources staying untouched at this stage.

\paragraph{Cost-aware modality cascading.}
\label{sec:method-cascade}
Re-ranking decides which evidence to read, then the cascade decides
how much to pay for it. Since the decisive cost in a multimodal
assistant is incurred exactly when raw sources enter the
prompt, yet a chunk's textual surrogate often already suffices, we make answering a cascade that prefers the cheapest
sufficient representation and pays for raw modalities only on demand.
\underline{\text{Stage~1}} answers $q$ from text only, using generated surrogates including captions,
transcriptions, and file digests presented in the retrieved
context. A confidence test $\Phi$ then decides whether this answer can be
trusted, escalating ($\Phi{=}1$) when the text answer is unreliable
($u(\hat{a}_1){=}1$) or when the modality bias is unsatisfied by text
($\beta(q){=}1$):
\begin{equation}
\Phi(\hat{a}_1,q)=\mathbb{1}\!\bigl[\,u(\hat{a}_1) \,\vee\, \beta(q)\,\bigr].
\label{eq:phi}
\end{equation}
If reaches escalation threshold, \underline{\text{Stage~2}} attaches the raw source of modality $m$
under a per-modality gate driven by a fused modality-evidence score
\begin{equation}
\small
e_m(q) = w_r\,\frac{1}{k}\!\sum_{c\in\mathcal{R}_k}\!\mathbb{1}_m(c)
       + w_b\, b_m(q)
       + w_s\, \sigma_m(q),
\label{eq:escore}
\end{equation}
which fuses, respectively, the fraction of retrieved chunks carrying
modality $m$, the detected bias, and a surface lexical cue
$\sigma_m(q)$; the raw source is supplied only when this evidence clears
the threshold.


The common case of a text-answerable question terminates after one cheap
call, while the modality gate ensures that raw modality
sources are consumed only when fused evidence, rather than a single brittle cue, indicates they are necessary. This yields the
trade-off we target: questions that genuinely require inspection still
escalate and retain accuracy, whereas text-answerable questions avoid the
multimodal premium, lowering the average token cost without
sacrificing correctness.

\begin{table*}[t!]
\small
\centering
\caption{Overall performance comparison on \benchmark. All memory systems are implemented with Qwen-2.5-VL-7B as the default backbone. Best and second-best are marked in \textbf{bold} and \underline{underline}: Frontier MLLMs are ranked separately, while text-based and
multimodal agentic-memory systems are ranked as a single group.}
\vspace{-2mm}

\definecolor{lcblue}{RGB}{229, 238, 250}
\definecolor{refgray}{HTML}{ECECEC}
\newcommand{\icn}[1]{\raisebox{-0.25ex}{\includegraphics[height=1.0em]{figures/icons/#1.png}}\,}
\newcolumntype{C}{>{\scriptsize\centering\arraybackslash}X}
\newcolumntype{W}{>{\scriptsize\centering\arraybackslash}p{2.9cm}}
\newcolumntype{S}{>{\scriptsize\centering\arraybackslash}X}  
\setlength{\tabcolsep}{3pt}
\renewcommand{\arraystretch}{1.15}
\setlength{\fboxsep}{1pt}

\begin{tabularx}{\linewidth}{W|CC|CC|CC|CC|CC|CC|C|C|S}

\toprule
\multirow{2}{*}{\textbf{Agent Type}} & \multicolumn{2}{c}{\scriptsize\textbf{SS}} & \multicolumn{2}{c}{\scriptsize\textbf{MS}} & \multicolumn{2}{c}{\scriptsize\textbf{TR}} & \multicolumn{2}{c}{\scriptsize\textbf{MR}} & \multicolumn{2}{c}{\scriptsize\textbf{TH}} & \multicolumn{2}{c}{\scriptsize\textbf{II}} & \scriptsize\textbf{FM} & \scriptsize\textbf{FJ} & \scriptsize{\textbf{Overall}}\\
 & \scriptsize F1 & \scriptsize LLM & \scriptsize F1 & \scriptsize LLM & \scriptsize F1 & \scriptsize LLM & \scriptsize F1 & \scriptsize LLM & \scriptsize F1 & \scriptsize LLM & \scriptsize F1 & \scriptsize LLM & \scriptsize EM & \scriptsize EM & \scriptsize{Score} \\
\hline
\multicolumn{16}{c}{\cellcolor{lcblue!60}\textit{{Frontier MLLMs}}} \\
  \icn{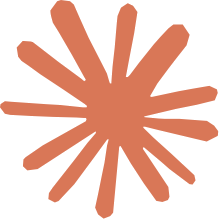}Claude-Opus-4.6 & 0.2593 & 0.6050 & 0.3678 & 0.5687 & 0.3754 & \underline{0.5813} & 0.5354 & \textbf{0.7306} & 0.2387 & \underline{0.6083} & 0.2412 & 0.6501 & \underline{0.2203} & 0.9443 & 0.4937 \\
  \icn{openai-icon}GPT-5.4 & 0.5257 & \underline{0.7623} & 0.3276 & \textbf{0.6274} & 0.4975 & 0.4950 & 0.4477 & 0.6389 & 0.1683 & 0.4333 & 0.3227 & \textbf{0.7125} & 0.1890 & 0.9621 & 0.4823 \\
  \icn{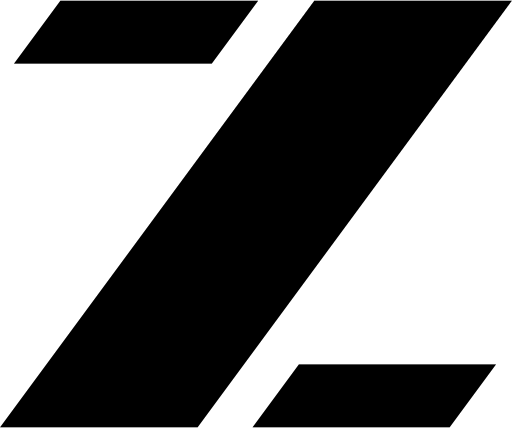}GLM-5.1 & \underline{0.5409} & 0.6510 & \textbf{0.6361} & 0.6066 & 0.5142 & 0.5115 & 0.5515 & 0.6550 & \textbf{0.5843} & \textbf{0.6193} & \textbf{0.3735} & 0.5667 & 0.1667 & 0.9332 & \textbf{0.5493} \\
  \icn{qwen-icon}Qwen3.6-Plus & 0.4713 & 0.6354 & 0.5907 & 0.6004 & \underline{0.5218} & 0.4799 & 0.5167 & 0.6406 & 0.5325 & 0.5227 & \underline{0.3347} & 0.3833 & 0.1520 & \textbf{0.9800} & \underline{0.5049} \\
  \icn{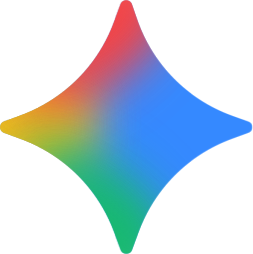}Gemini-3.1-Pro-Preview & 0.5320 & \textbf{0.7693} & 0.2991 & 0.5437 & 0.4926 & 0.5404 & 0.3789 & 0.5111 & 0.3161 & 0.4417 & 0.2870 & \underline{0.6875} & \textbf{0.2403} & \underline{0.9666} & 0.4668 \\
  \icn{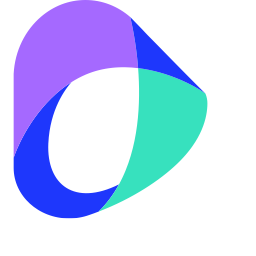}Doubao-Seed-2.0-Pro & \textbf{0.5638} & 0.2917 & \underline{0.6330} & 0.3607 & \textbf{0.6514} & \textbf{0.6437} & \textbf{0.7204} & 0.6586 & \underline{0.5480} & 0.1364 & 0.3005 & 0.4103 & 0.1950 & 0.9532 & 0.4938 \\
  \icn{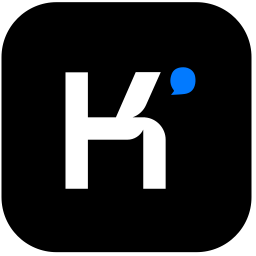}Kimi-k2.5 & 0.4269 & 0.6304 & 0.3410 & 0.4536 & 0.4642 & 0.5050 & \underline{0.5614} & 0.6556 & 0.1987 & 0.4351 & 0.2487 & 0.5589 & 0.1308 & 0.8667 & 0.4493 \\
  \icn{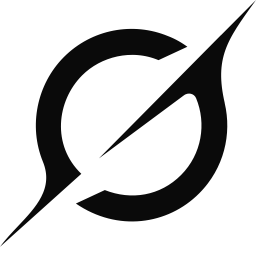}Grok-4 & 0.4383 & 0.6211 & 0.3129 & \underline{0.6125} & 0.3259 & 0.5150 & 0.3912 & \underline{0.7034} & 0.1450 & 0.3667 & 0.2322 & 0.6875 & 0.1871 & 0.7923 & 0.4357 \\
\hline\hline
  \rowcolor{refgray!60}\icn{qwen-icon}Qwen-2.5-VL-7B & 0.2742 & 0.5468 & 0.4587 & 0.3471 & 0.4378 & 0.5063 & 0.5425 & 0.4682 & 0.2461 & 0.0635 & 0.1672 & 0.0523 & 0.2070 & 0.7904 & 0.4179 \\
\hline
  \multicolumn{16}{c}{\cellcolor{Secondary!60}\textit{{Text-based Agentic-Memory Systems}}} \\
  NaiveRAG & 0.4430 & 0.6547 & 0.3117 & 0.4429 & 0.2349 & 0.4167 & 0.2080 & 0.2572 & 0.1321 & 0.2806 & 0.1263 & 0.3202 & 0.2569 & 0.7840 & 0.3249 \\
  A-Mem & \textbf{0.5921} & \underline{0.7247} & 0.3998 & 0.5345 & 0.1298 & 0.1697 & 0.2403 & 0.3727 & 0.1361 & 0.3493 & 0.1227 & 0.2872 & 0.2875 & 0.9243 & 0.3450 \\
  Mem0 (w/o vision) & 0.5343 & 0.6957 & 0.3837 & 0.5448 & 0.1361 & 0.1853 & 0.2471 & 0.3787 & 0.1346 & 0.3843 & 0.0962 & 0.3000 & 0.1544 & 0.9243 & 0.3410 \\
  MemoryOS & 0.5410 & 0.7047 & 0.3399 & 0.5145 & 0.1248 & 0.3977 & 0.2146 & 0.3299 & 0.1274 & 0.3327 & 0.1162 & 0.3175 & 0.3150 & \underline{0.9733} & 0.3470 \\
\hline
  \multicolumn{16}{c}{\cellcolor{tintred!60}\textit{{Multimodal Agentic-Memory Systems}}} \\
  Universal-RAG & \underline{0.5858} & \textbf{0.7328} & \textbf{0.4349} & \textbf{0.5769} & 0.3903 & 0.4431 & 0.2938 & 0.4369 & 0.2691 & \underline{0.4872} & 0.1586 & 0.5559 & 0.5367 & 0.8241 & 0.4366 \\
  RAG-Anything & 0.5849 & 0.6418 & 0.3940 & 0.5474 & \underline{0.4353} & 0.4465 & 0.3863 & 0.4970 & \underline{0.2782} & 0.4390 & \underline{0.1873} & \underline{0.6201} & 0.4572 & 0.8508 & 0.4481 \\
  Mem0 (with vision) & 0.4798 & 0.6457 & \underline{0.4181} & 0.5548 & 0.1511 & 0.1837 & 0.4459 & 0.5607 & 0.1485 & 0.3527 & 0.1286 & 0.5973 & 0.1942 & 0.9176 & 0.3969 \\
  MemVerse & 0.4850 & 0.6406 & 0.3895 & 0.5662 & 0.3229 & 0.4096 & \underline{0.4591} & 0.5613 & 0.2518 & 0.4223 & 0.1849 & 0.6048 & 0.5474 & 0.9042 & 0.4481 \\
  NGM & 0.3559 & 0.5266 & 0.3453 & 0.4660 & 0.1716 & 0.1601 & \textbf{0.4708} & 0.5479 & 0.2288 & 0.3905 & 0.1807 & 0.6060 & \textbf{0.6208} & 0.9042 & 0.3813 \\
  MIRIX & 0.4954 & 0.6378 & 0.2853 & 0.4875 & \textbf{0.4564} & \textbf{0.4949} & 0.4076 & \underline{0.5775} & \textbf{0.2999} & \textbf{0.5917} & 0.1054 & 0.5375 & 0.3807 & 0.8664 & \underline{0.4560} \\
  \rowcolor{refgray!60}
  \textbf{\method} (Ours) & 0.4993 & 0.6924 & 0.3697 & \underline{0.5667} & 0.3952 & \underline{0.4877} & 0.4305 & \textbf{0.6063} & 0.2202 & 0.4786 & \textbf{0.3021} & \textbf{0.6515} & \underline{0.5688} & \textbf{0.9844} & \textbf{0.4838} \\
\bottomrule
\end{tabularx}
\label{tab:main-results}
\end{table*}

\begin{figure*}
  \centering
  \includegraphics[width=\linewidth]{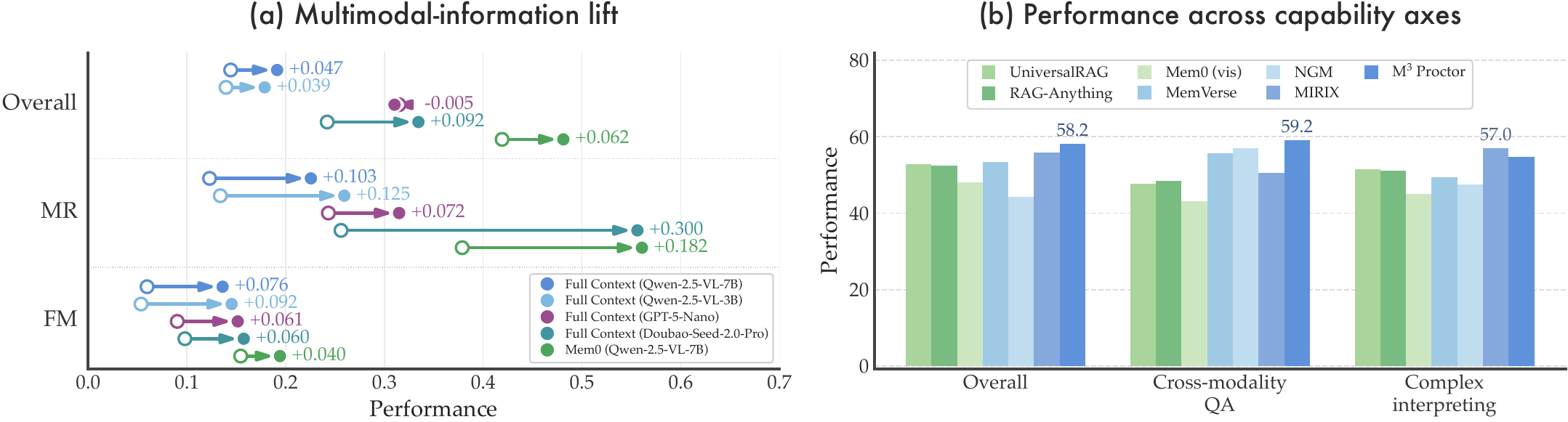}
  \caption{Multimodal influence and capability profile on \benchmark{}. \emph{Cross-modality QA} involve \textsc{mr}, \textsc{fm} and other queries labeled as modality-coupled, while \emph{Complex Interpreting} involve \textsc{th} and \textsc{ii}.}
  \label{fig:combined}
\end{figure*}

\section{Benchmarking Analysis}
\label{sec:experiments}

\subsection{Experimental Setup}

\paragraph{Models and Baselines.}
We compare two families of systems on \benchmark{}.
\emph{Frontier MLLMs} include the latest strong closed-source
models---Claude-Opus-4.6, GPT-5.4, GLM-5.1, Qwen3.6-Plus, 
Gemini-3.1-Pro-Preview, Doubao-Seed-2.0-Pro, Kimi-k2.5 and Grok-4---answering directly from the 
labeled supporting evidence without external memory module, serving as a no-memory reference
(details in Appendix~\ref{sec:appendix-mllm}).
\emph{Agent-memory systems}
instead manage an external memory and retrieve from it; we group them
into text-based methods: NaiveRAG~\citep{lewis2020rag},
A-Mem~\citep{xu2025amem}, Mem0~\citep{chhikara2025mem0}, and
MemoryOS~\citep{kang2025memory}, and multimodal
methods: UniversalRAG~\citep{yeo2025universalrag},
RAG-Anything~\citep{guo2025raganything},
MemVerse~\citep{liu2025memverse}, NGM~\citep{fisher2025ngm},
MIRIX~\citep{wang2025mirix}, and Mem0 with vision enabled. All
agentic-memory systems, including \method, are deployed on the same
Qwen-2.5-VL-7B~\citep{bai2025qwen25vl} backbone (base results highlighted). Implementation details are deferred to Appendix~\ref{sec:appendix-config}.

\paragraph{Evaluation Metrics.}
\label{sec:eval-metrics}


We evaluate different model answers under four complementary metrics: exact match (EM), token-level F1 score, BLEU-1, and the LLM-as-a-Judge (LLM-J) adapted from Mem-Gallery~\citep{bei2026mem} (using Qwen2.5-VL-32B-Instruct as the judge) to report representative metrics. Since \textsc{fm} (image identifiers) and \textsc{fj} (single-letter
choices) have closed-form answers, we only report the EM metric. We further summarize each model with an
\emph{Overall Score}, a weighted comprehensive metric over all evaluated
answer-quality signals. Complete scoring definitions and implementation details can be found in Appendix~\ref{sec:appendix-metrics}.


\subsection{Main Results}
\label{sec:main-results}

Table~\ref{tab:main-results} reports the main results with result breakdown and alternative backbone evaluations in Appendix~\ref{sec:backboneaaa}. We read it
along three questions: how far current systems are from solving
\benchmark, where they fail, and whether \method closes the gap.

\noindent\ding{182}\
\textit{M$^3$Exam poses a significant challenge.}
The strongest frontier MLLM (GLM-5.1) reaches only $0.549$ overall,
\emph{even when relieved of retrieval} (full-context variant
shown in Appendix~\ref{app:fullcontext} degrades further still).
Agentic-memory systems (built on Qwen-2.5-VL-7B) improve over the base
model but generally trail frontier MLLMs, with multimodal-enhanced
variants scoring higher. Taken together, these results
indicate that \benchmark{} remains far from solved, exposing
capabilities that current models and memory systems have yet to
acquire.

\noindent\ding{183}\
\textit{Cross-modal and implicit-intent questions are the bottleneck.}
Results show that systems handle single-session reasoning well but degrade sharply on the
capabilities \benchmark{} targets. The base model collapses on complex
interpretation tasks ($0.064$ on \textsc{th} and $0.052$ on \textsc{ii})
and remains weak on cross-modal reasoning ($0.468$ on \textsc{mr} and
$0.207$ on \textsc{fm}), yet external memory recovers much of this
shortfall, approaching and at times surpassing frontier MLLMs. These
results suggest that the bottleneck of \benchmark lies in realistic multimodal
reasoning and implicit interpretation, which stems from inadequate
evidence management rather than a fundamental backbone ceiling and is
therefore addressable.

\noindent\ding{184}\
\textit{M$^3$Proctor advances the agentic-memory frontier.}
\method attains the best Overall Score among agentic-memory systems
($0.484$ vs.\ $0.456$ for the prior best), with the gains concentrated
where the benchmark is hardest: it leads on \textsc{mr} ($0.606$
LLM-J), \textsc{ii} ($0.652$), and \textsc{fm} ($0.569$ EM). Most
notably, despite running on a modest $7$B backbone and retrieving its
own evidence, \method surpasses several retrieval-relieved frontier
MLLMs and trails only the strongest proprietary models. This positions \method as a competitive agentic-memory system that matches
frontier MLLMs in multimodal memory while operating at a fraction of
their scale.

\takeaway{}{\benchmark{} remains far from solved, with cross-modal
grounding and implicit inference as the core challenge, but \method
offers an efficient open-source baseline.}

\begin{figure*}
  \centering
  \includegraphics[width=\linewidth]{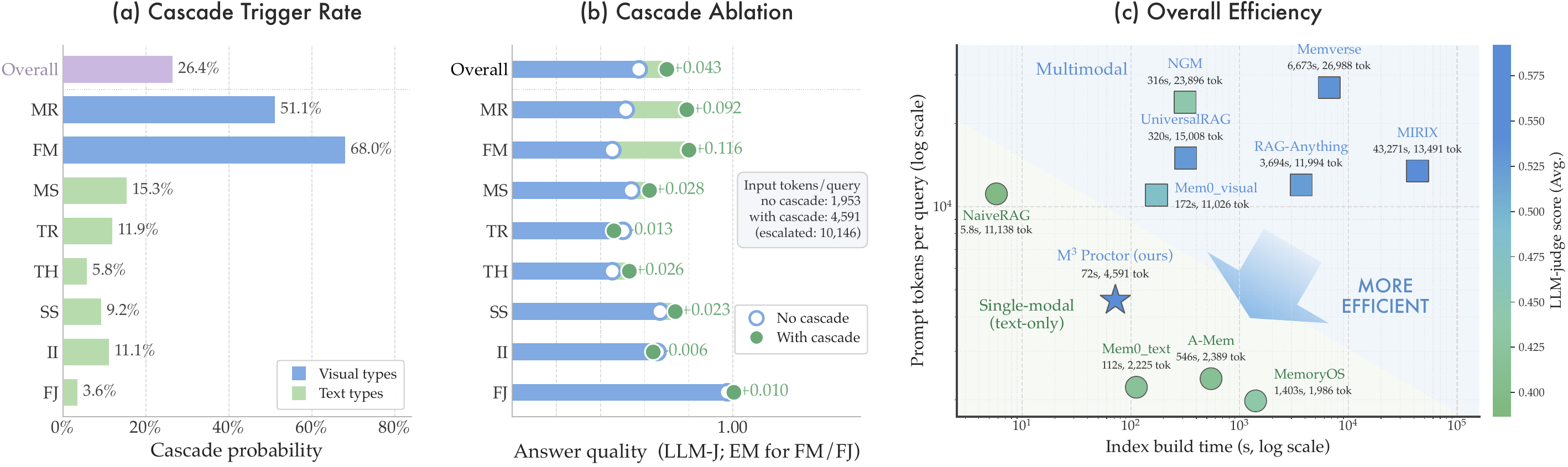}
  \caption{In-depth analysis of cascade ablation and \method efficiency performance.}
  \label{fig:depth}
\end{figure*}

\section{In-Depth Analysis}
\label{sec:in-depth}
We now further analyze the multimodal dependence of \benchmark{} (Figure~\ref{fig:combined}), and the cascade behavior and efficiency performance of \method (Figure~\ref{fig:depth}).

\noindent\ding{182}\
\textit{M$^3$Exam is tightly coupled to its visual evidence.}
Figure~\ref{fig:combined} (a) measures the multimodal-information lift:
the performance gain a system obtains from accessing raw multimodal
evidence rather than its generated textual surrogates alone. 
The lift is modest overall but pronounced on the strongly multimodal subsets, reaching up
to $+0.300$ on \textsc{mr} and $+0.116$ on \textsc{fm}, indicating that a textual
surrogate cannot stand in for the visual source on these questions. 
The capability profile in Figure~\ref{fig:combined}(b) is consistent:
systems differ most on cross-modality QA, where \method leads at
$0.592$ and remain
separated on complex interpreting ($0.570$), confirming that the
benchmark's hardest questions genuinely hinge on multimodal evidence and
implicit inference rather than lexical matching over the dialogue text.

\noindent\ding{183}\
\textit{The modality cascade fires where, and only where, vision is
needed.}
Figure~\ref{fig:depth} (a) shows that the cascade triggers far more
often on visual types ($68.0\%$ on \textsc{fm}, $51.1\%$ on \textsc{mr})
than on text types ($\leq 15\%$),
indicating that modality-bias detection reliably identifies which
queries require a raw source. Figure~\ref{fig:depth} (b) confirms the
escalation pays off precisely there: activating the cascade lifts
answer quality by $0.116$ on \textsc{fm} and $0.092$ on \textsc{mr}
while leaving text types essentially unchanged. The cascade thus supplies
multimodal evidence selectively, rather than re-feeding it indiscriminately.

\noindent\ding{184}\
\textit{M$^3$Proctor matches multimodal accuracy at text-only cost.}
Figure~\ref{fig:depth}~(c) plots index-build time against per-query
token consumption, where cascade escalation calls are taken into
account. Results show that although \method is a multimodal memory, 
its footprint falls within the \emph{text-only} regime---$72$\,s
to build the index and $4{,}591$ tokens per query---one to three orders
of magnitude below multimodal baselines such as MIRIX ($43{,}271$\,s)
and NGM ($26{,}988$ tokens), since raw visuals enter the prompt only on
the $26.4\%$ of queries that the cascade escalates. \method nonetheless
attains the highest performance of any system, demonstrating that
modality-aware escalation---rather than indiscriminate visual
injection---is what delivers multimodal accuracy.

\takeaway{}{\benchmark{} relies heavily on its visual evidence, and
\method's modality cascade supplies that evidence only when needed---%
matching multimodal level accuracy at significantly reduced cost.}

\section{Related Work}
\label{sec:related}

\paragraph{Benchmarks for Long-term Multimodal Memory.}
Early efforts ground multi-turn QA in
documents, with MMLongBench-Doc~\cite{ma2024mmlongbench} over rendered
pages and MultiDoc2Dial~\cite{feng2021multidoc2dial} yet treat documents as static inputs rather
than accumulating memory. Later benchmarks shift to evolving
dialogue histories: LongMemEval~\cite{wu2025longmemeval} targets
retrieval across very long histories, while
MemoryArena~\cite{he2026memoryarena} and
MemoryAgentBench~\cite{hu2025evaluating} probe fact update,
consolidation, and retrieval, but remain text-only. LoCoMo~\cite{maharana2024locomo}
introduced images into long-term conversational memory, and recent
multimodal benchmarks like MMDialog~\cite{feng2023mmdialog},
MMRC~\cite{xue2025mmrc}, Mem-Gallery~\cite{bei2026mem} push
toward multi-session multimodal memory, yet treat images only as
topical anchors and leave out implicit intent. \emph{\benchmark unifies
these strands, uniquely evaluating implicit-intent inference over
realistic accumulating document-level multimodal conversations.}

\paragraph{Systems for Long-term Multimodal Memory.}
To overcome fixed context windows, external memory systems store
past interactions and retrieve relevant evidence on demand. Textual
memory systems like A-Mem~\cite{xu2025amem} links
interactions into a Zettelkasten-style note network,
while Mem0~\cite{chhikara2025mem0} consolidates salient facts into a
compact store, and MemoryOS~\cite{kang2025memory} unifies parametric, activation, and
plaintext memory under an OS abstraction.
The inspired multimodal memory systems like Universal-RAG~\cite{yeo2025universalrag}
and RAG-Anything~\cite{guo2025raganything}which retrieve across text,
images, and tables; MemVerse~\cite{liu2025memverse} and neural graph
memory~\cite{fisher2025ngm} encode cross-modal associations; and
MIRIX~\cite{wang2025mirix} manages multimodal experience through
specialized components. However current systems conduct global uniform retrieval, 
regardless of whether a query actually needs visual evidence grounding. 
\emph{\method predicts a query's multimodal intent and returns text surrogates
or raw artifacts accordingly, recovering evidence at minimal cost.}
\section{Conclusion}
\label{sec:conclusion}

We introduce \benchmark, a query-centric benchmark evaluating how well
multimodal conversational agents memorize, reason, and interpret
over long-horizon histories of dialogue, images, and documents. Our
evaluation reveals a clear gap: existing memory systems and frontier
MLLMs handle single-session recall but struggle with cross-modal
grounding, cross-session reasoning, and implicit-intent inference.
We therefore propose \method, a modality-aware method that
infers a query's modality bias and escalates to raw sources only on
demand, improving accuracy while substantially cutting tokens and
index-construction time.

\section*{Limitations}

Although \benchmark targets realistic multimodal user--agent interactions, the current benchmark mainly focuses on single-turn QA over accumulated history. Real-world interactions often involve long-horizon multi-turn conversations with evolving intents and iterative memory updates, which remain underexplored in all existing multimodal memory benchmarks. We plan to further extend \benchmark toward more dynamic multi-turn memory evaluation in future work.
In addition, while \method improves modality-aware multimodal retrieval and reasoning efficiency, its capability on interpreting-oriented questions remains limited. Better modeling implicit user states and performing deeper contextual interpretation over long-term multimodal interactions remain important directions for future work.

\section*{Ethics Statement}

\paragraph{Data provenance and privacy.}
\benchmark{} contains no real user data: every persona, core event,
timeline, and dialogue turn is synthetically generated by large
language models from hand-authored seeds
(Table~\ref{tab:persona-core}), so the conversations describe fictional
individuals. The personas are occupationally varied yet avoid sensitive
attributes and real-world identities; any resemblance to an actual
person is unintended. Being fully synthetic, the benchmark raises no
consent, de-identification, or PII concerns.

\paragraph{Multimodal artifacts.}
The visual and document evidence attached to each conversation is 
an assembled persona-aligned file pool
of charts, photographs, and PDF documents. We use only artifacts that
are obtained under terms that permit research
use, and we do not redistribute copyrighted third-party material. We
manually screened the generated content to remove any unsafe,
offensive, or privacy-leaking material before release.

\bibliography{custom}

@inproceedings{feng2023mmdialog,
  title     = {{MMDialog}: A Large-Scale Multi-Turn Dialogue Dataset towards Multi-Modal Open-Domain Conversation},
  author    = {Feng, Jiazhan and Sun, Qingfeng and Xu, Can and Zhao, Pu and Yang, Yaming and Tao, Chongyang and Zhao, Dongyan and Lin, Qingwei},
  booktitle = {Proceedings of the 61st Annual Meeting of the Association for Computational Linguistics (ACL)},
  year      = {2023}
}

@inproceedings{maharana2024locomo,
  title     = {Evaluating Very Long-Term Conversational Memory of {LLM} Agents},
  author    = {Maharana, Adyasha and Lee, Dong-Ho and Tulyakov, Sergey and Bansal, Mohit and Barbieri, Francesco and Fang, Yuwei},
  booktitle = {Proceedings of the 62nd Annual Meeting of the Association for Computational Linguistics (ACL)},
  year      = {2024}
}

@article{qwen2025vl,
  title={Qwen2.5-VL Technical Report},
  author={{Qwen Team} and Bai, Shuai and Chen, Keqin and Liu, Xuejing and Wang, Jialin and Ge, Wenbin and Song, Sibo and Dang, Kai and Wang, Peng and Wang, Shijie and Tang, Jun and others},
  journal={arXiv preprint arXiv:2502.13923},
  year={2025},
  url={https://arxiv.org/abs/2502.13923}
}

@article{xue2025mmrc,
  title   = {{MMRC}: A Large-Scale Benchmark for Understanding Multimodal Large Language Model in Real-World Conversation},
  author  = {Xue, Haochen and Tang, Feilong and Hu, Ming and Liu, Yexin and Huang, Qidong and Li, Yulong and Liu, Chengzhi and Xu, Zhongxing and Zhang, Chong and Feng, Chun-Mei and Xie, Yutong and Razzak, Imran and Ge, Zongyuan and Su, Jionglong and He, Junjun and Qiao, Yu},
  journal = {arXiv preprint arXiv:2502.11903},
  year    = {2025}
}

@inproceedings{xu2025amem,
  title     = {{A-MEM}: Agentic Memory for {LLM} Agents},
  author    = {Xu, Wujiang and Liang, Zujie and Mei, Kai and Gao, Hang and Tan, Juntao and Zhang, Yongfeng},
  booktitle = {Advances in Neural Information Processing Systems (NeurIPS)},
  year      = {2025}
}

@article{chhikara2025mem0,
  title   = {Mem0: Building Production-Ready {AI} Agents with Scalable Long-Term Memory},
  author  = {Chhikara, Prateek and Khant, Dev and Aryan, Saket and Singh, Taranjeet and Yadav, Deshraj},
  journal = {arXiv preprint arXiv:2504.19413},
  year    = {2025}
}

@inproceedings{wu2025longmemeval,
  title     = {{LongMemEval}: Benchmarking Chat Assistants on Long-Term Interactive Memory},
  author    = {Wu, Di and Wang, Hongwei and Yu, Wenhao and Zhang, Yuwei and Chang, Kai-Wei and Yu, Dong},
  booktitle = {International Conference on Learning Representations (ICLR)},
  year      = {2025}
}

@inproceedings{feng2021multidoc2dial,
  title     = {{MultiDoc2Dial}: Modeling Dialogues Grounded in Multiple Documents},
  author    = {Feng, Song and Patel, Siva and Wan, Hui and Joshi, Sachindra},
  booktitle = {Proceedings of the 2021 Conference on Empirical Methods in Natural Language Processing (EMNLP)},
  year      = {2021}
}

@article{bai2025qwen25vl,
  title   = {{Qwen2.5-VL} Technical Report},
  author  = {Bai, Shuai and Chen, Keqin and Liu, Xuejing and Wang, Jialin and Ge, Wenbin and others},
  journal = {arXiv preprint arXiv:2502.13923},
  year    = {2025}
}

@inproceedings{lewis2020rag,
  title     = {Retrieval-Augmented Generation for Knowledge-Intensive {NLP} Tasks},
  author    = {Lewis, Patrick and Perez, Ethan and Piktus, Aleksandra and Petroni, Fabio and Karpukhin, Vladimir and Goyal, Naman and K{\"u}ttler, Heinrich and Lewis, Mike and Yih, Wen-tau and Rockt{\"a}schel, Tim and Riedel, Sebastian and Kiela, Douwe},
  booktitle = {Advances in Neural Information Processing Systems (NeurIPS)},
  year      = {2020}
}

@article{yeo2025universalrag,
  title   = {{UniversalRAG}: Retrieval-Augmented Generation over Corpora of Diverse Modalities and Granularities},
  author  = {Yeo, Woongyeong and Kim, Kangsan and Jeong, Soyeong and Baek, Jinheon and Hwang, Sung Ju},
  journal = {arXiv preprint arXiv:2504.20734},
  year    = {2025}
}

@article{guo2025raganything,
  title   = {RAG-Anything: All-in-One RAG Framework},
  author  = {Guo, Zirui and Ren, Xubin and Xu, Lingrui and Zhang, Jiahao and Huang, Chao},
  journal = {arXiv preprint arXiv:2510.12323},
  year    = {2025}
}

@article{liu2025memverse,
  title   = {MemVerse: Multimodal Memory for Lifelong Learning Agents},
  author  = {Liu, Junming and others},
  journal = {arXiv preprint arXiv:2512.03627},
  year    = {2025}
}

@misc{fisher2025ngm,
  title  = {Neural Graph Memory: A Structured Approach to Long-Term Memory in Multimodal Agents},
  author = {Fisher, Matthew},
  year   = {2025}
}

@article{wang2025mirix,
  title   = {MIRIX: Multi-Agent Memory System for LLM-Based Agents},
  author  = {Wang, Yu and Chen, Xi},
  journal = {arXiv preprint arXiv:2507.07957},
  year    = {2025}
}

@inproceedings{rajpurkar2016squad,
  title     = {{SQuAD}: 100,000+ Questions for Machine Comprehension of Text},
  author    = {Rajpurkar, Pranav and Zhang, Jian and Lopyrev, Konstantin and Liang, Percy},
  booktitle = {Proceedings of the 2016 Conference on Empirical Methods in Natural Language Processing (EMNLP)},
  pages     = {2383--2392},
  year      = {2016}
}

@inproceedings{papineni2002bleu,
  title     = {{BLEU}: a Method for Automatic Evaluation of Machine Translation},
  author    = {Papineni, Kishore and Roukos, Salim and Ward, Todd and Zhu, Wei-Jing},
  booktitle = {Proceedings of the 40th Annual Meeting of the Association for Computational Linguistics (ACL)},
  pages     = {311--318},
  year      = {2002}
}

@article{cheng2025higher,
  title={Higher Satisfaction, Lower Cost: A Technical Report on How LLMs Revolutionize Meituan's Intelligent Interaction Systems},
  author={Cheng, Xuxin and Zeng, Ke and Cao, Zhiquan and Dai, Linyi and Gao, Wenxuan and Han, Fei and Jian, Ai and Hong, Feng and Hu, Wenxing and Huang, Zihe and others},
  journal={arXiv preprint arXiv:2510.13291},
  year={2025}
}

@article{he2026memoryarena,
  title={Memoryarena: Benchmarking agent memory in interdependent multi-session agentic tasks},
  author={He, Zexue and Wang, Yu and Zhi, Churan and Hu, Yuanzhe and Chen, Tzu-Ping and Yin, Lang and Chen, Ze and Wu, Tong Arthur and Ouyang, Siru and Wang, Zihan and others},
  journal={arXiv preprint arXiv:2602.16313},
  year={2026}
}

@article{bei2026mem,
  title={Mem-gallery: Benchmarking multimodal long-term conversational memory for mllm agents},
  author={Bei, Yuanchen and Wei, Tianxin and Ning, Xuying and Zhao, Yanjun and Liu, Zhining and Lin, Xiao and Zhu, Yada and Hamann, Hendrik and He, Jingrui and Tong, Hanghang},
  journal={arXiv preprint arXiv:2601.03515},
  year={2026}
}

@inproceedings{tanaka2023slidevqa,
  title={Slidevqa: A dataset for document visual question answering on multiple images},
  author={Tanaka, Ryota and Nishida, Kyosuke and Nishida, Kosuke and Hasegawa, Taku and Saito, Itsumi and Saito, Kuniko},
  booktitle={Proceedings of the AAAI Conference on Artificial Intelligence},
  volume={37},
  number={11},
  pages={13636--13645},
  year={2023}
}

@article{ma2024mmlongbench,
  title={Mmlongbench-doc: Benchmarking long-context document understanding with visualizations},
  author={Ma, Yubo and Zang, Yuhang and Chen, Liangyu and Chen, Meiqi and Jiao, Yizhu and Li, Xinze and Lu, Xinyuan and Liu, Ziyu and Ma, Yan and Dong, Xiaoyi and others},
  journal={Advances in Neural Information Processing Systems},
  volume={37},
  pages={95963--96010},
  year={2024}
}

@article{zhang2025survey,
  title={A survey on the memory mechanism of large language model-based agents},
  author={Zhang, Zeyu and Dai, Quanyu and Bo, Xiaohe and Ma, Chen and Li, Rui and Chen, Xu and Zhu, Jieming and Dong, Zhenhua and Wen, Ji-Rong},
  journal={ACM Transactions on Information Systems},
  volume={43},
  number={6},
  pages={1--47},
  year={2025},
  publisher={ACM New York, NY}
}

@article{hu2025evaluating,
  title={Evaluating memory in llm agents via incremental multi-turn interactions},
  author={Hu, Yuanzhe and Wang, Yu and McAuley, Julian},
  journal={arXiv preprint arXiv:2507.05257},
  year={2025}
}

@misc{openai2026,
  author       = {{OpenAI}},
  title        = {{GPT-5.5} Instant: Smarter, clearer, and more personalized},
  howpublished = {\url{https://openai.com/index/gpt-5-5-instant/}},
  year         = {2026},
  month        = {May}
}

@misc{anthropic2026,
  author       = {{Anthropic}},
  title        = {Introducing {Claude Opus 4.7}},
  howpublished = {\url{https://www.anthropic.com/news/claude-opus-4-7}},
  year         = {2026},
  month        = {April}
}

@misc{google2026gemini31pro,
  author       = {{Google DeepMind}},
  title        = {{Gemini 3.1 Pro}: Best for complex tasks and bringing creative concepts to life},
  howpublished = {\url{https://deepmind.google/models/gemini/pro/}},
  year         = {2026},
  month        = {February}
}

@misc{anthropic2026opus46,
  author       = {{Anthropic}},
  title        = {Introducing {Claude Opus 4.6}},
  howpublished = {\url{https://www.anthropic.com/news/claude-opus-4-6}},
  year         = {2026},
  month        = {February}
}

@misc{openai2026gpt54,
  author       = {{OpenAI}},
  title        = {Introducing {GPT-5.4}},
  howpublished = {\url{https://openai.com/index/introducing-gpt-5-4/}},
  year         = {2026},
  month        = {March}
}

@misc{zhipu2026glm51,
  author       = {{Zhipu AI}},
  title        = {{GLM-5.1}},
  howpublished = {\url{https://z.ai/blog/glm-5.1}},
  year         = {2026},
  month        = {April}
}

@misc{alibaba2026qwen36,
  author       = {{Qwen Team}},
  title        = {{Qwen3.6-Plus}},
  howpublished = {\url{https://qwen.ai/blog?id=qwen3.6}},
  year         = {2026}
}

@misc{bytedance2026doubao,
  author       = {{ByteDance}},
  title        = {{Doubao-Seed-2.0}},
  howpublished = {\url{https://seed.bytedance.com/en/seed2}},
  year         = {2026}
}

@misc{moonshot2026kimi25,
  author       = {{Moonshot AI}},
  title        = {{Kimi K2.5}},
  howpublished = {\url{https://www.kimi.com/blog/kimi-k2-5}},
  year         = {2026},
  month        = {January}
}

@misc{xai2025grok4,
  author       = {{xAI}},
  title        = {{Grok 4}},
  howpublished = {\url{https://x.ai/news/grok-4}},
  year         = {2025}
}

@inproceedings{kang2025memory,
  title={Memory os of ai agent},
  author={Kang, Jiazheng and Ji, Mingming and Zhao, Zhe and Bai, Ting},
  booktitle={Proceedings of the 2025 Conference on Empirical Methods in Natural Language Processing},
  pages={25972--25981},
  year={2025}
}

\appendix










\clearpage
\begin{center}
    \Large{\sc\Huge Appendix\\\small \benchmark: Benchmarking Multimodal Memory for Realistic User-Agent Interactions}\\
\end{center}
\vskip 4mm
\startcontents[sections]
\vbox{\sc\Large Table of Contents}
\vspace{5mm}
\hrule height .8pt
\vspace{-2mm}
\printcontents[sections]{}{1}{\setcounter{tocdepth}{2}}
\vspace{4mm}
\hrule height .8pt
\vskip 10mm


\section{Dataset Details and Statistics}
\label{sec:appendix-data}

\subsection{Dataset Statistics}
\label{app:detaildata}

\begin{figure*}[t]
  \centering
  \includegraphics[width=\linewidth]{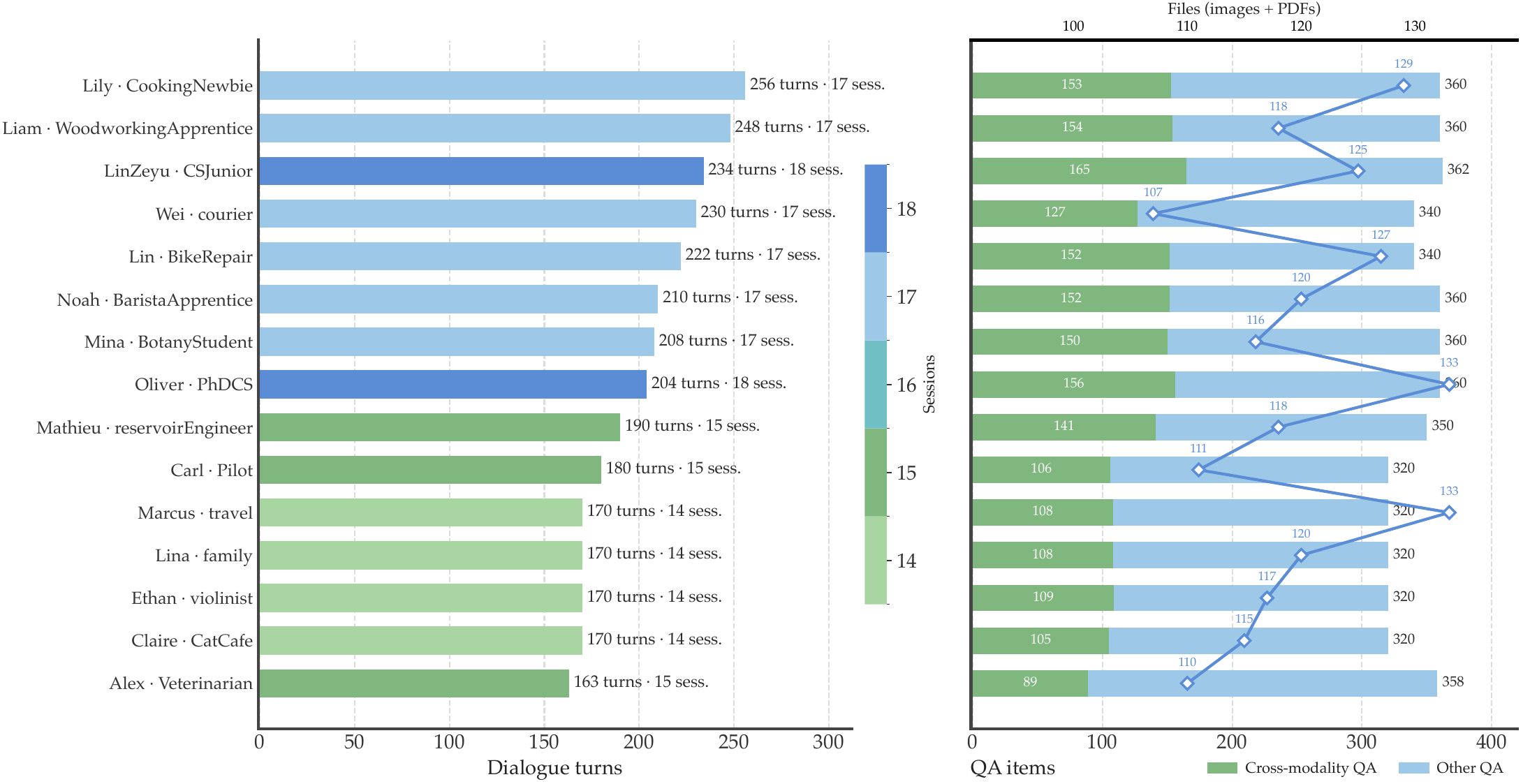}
  \caption{\textbf{Per-persona scale of \benchmark.}
\textbf{(A)}~\emph{Conversation size.} Dialogue turns per persona, with
bar colour encoding the number of sessions.
\textbf{(B)}~\emph{QA and asset footprint.} Total QA items per persona,
split into cross-modality QA (green; \textsc{fm} and \textsc{mr}) and
the remaining QA (blue); the overlaid line (top axis) reports attached
files per persona.}
  \label{fig:datasetoverview}
\end{figure*}

Figure~\ref{fig:datasetoverview} summarizes the per-persona scale of
the benchmark. In aggregate, \benchmark{} comprises $239$ multi-session
conversations spanning $3{,}025$ user--assistant rounds, $1{,}799$
attached images and multimodal artifacts, and $5{,}150$ evaluation items,
of which $1{,}975$ ($38.3\%$) are cross-modality questions whose answers
require grounding in an image or file. The corpus is deliberately
long-horizon: each persona holds $14$--$18$ sessions and $163$--$256$
dialogue turns (Panel~\textbf{A}), so answering a question typically
demands recalling content deposited across many earlier sessions rather
than within a single recent context window. The footprint is also
balanced across personas (Panel~\textbf{B}): every persona contributes a
comparable number of QA items ($320$--$362$) and attached files
($107$--$133$), and the nine document-bearing personas (bottom block)
carry a markedly higher share of cross-modality questions
($\geq 141$ each) than the six dialogue-only personas, reflecting the
extra chart- and PDF-grounded items their later sessions support. This
even, large-scale coverage ensures that no capability is measured on
only a few scenarios.

\subsection{Data Distribution}
\label{sec:appendix-data_distribution}

\begin{figure*}[t]
  \centering
  \includegraphics[width=\linewidth]{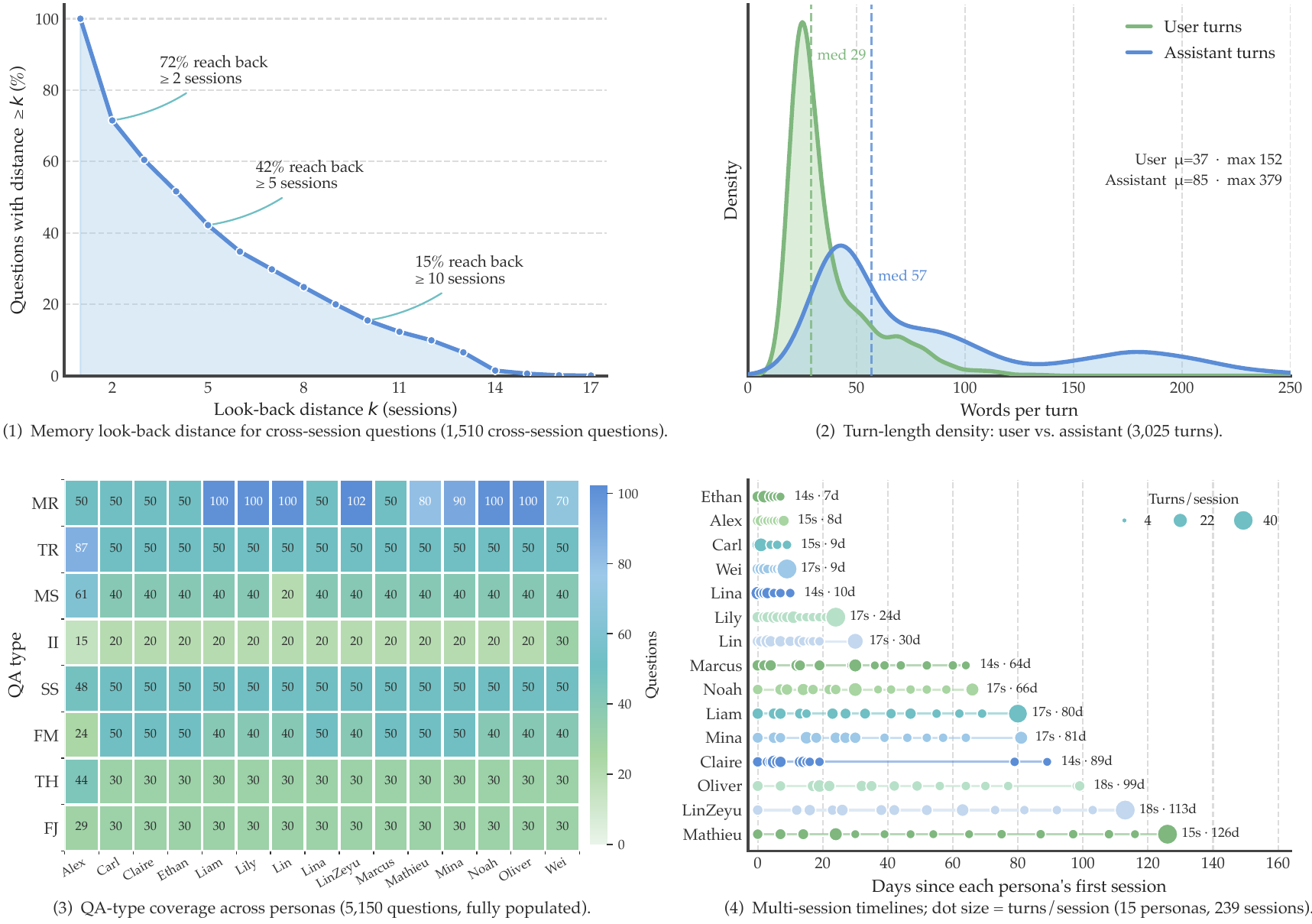}
  \caption{\textbf{Distributional properties of \benchmark.}
\textbf{(1)}~\emph{Memory look-back distance.} For each cross-session
question, the session-index gap between the earliest and latest round it
cites in \texttt{supporting\_facts}; the curve plots the fraction of
questions whose evidence spans at least $k$ sessions.
\textbf{(2)}~\emph{Turn-length density.} Gaussian-kernel density of word
counts for user versus assistant turns.
\textbf{(3)}~\emph{QA-type coverage across personas.} Number of
questions of each type (rows) per persona (columns).
\textbf{(4)}~\emph{Multi-session timelines.} One track per persona, each
dot a session at its real calendar date (normalized to day~$0$), with
dot size encoding turns per session.}
  \label{fig:data_distribution}
\end{figure*}

Figure~\ref{fig:data_distribution} characterizes \benchmark
along four axes that motivate its use as a cross-modal memory benchmark.

\paragraph{Answers are non-local and reach far back in the history.}
Panel~\textbf{(1)} quantifies how far apart in the conversation a
question's supporting evidence lies. Of the $1{,}510$ cross-session
questions, $72\%$ require evidence from rounds at least two sessions
apart, $42\%$ span five or more sessions, and the distribution carries a
long tail out to a maximum of $17$ sessions. Because each persona holds
$14$--$18$ sessions, a model cannot answer these items from recent
context alone: it must retain and retrieve information deposited many
sessions earlier. This long-range structure is the property that
text-summarisation or recency-biased baselines fail to capture, and it
complements the high \emph{Cross-Modality QA} counts in
Figure~\ref{fig:data_distribution}.

\paragraph{Dialogues exhibit a realistic, asymmetric exchange.}
Panel~\textbf{(2)} shows that user and assistant turns occupy clearly
different length regimes. User turns are short and tightly concentrated
(median $29$ words, mean $37$), reflecting focused questions and
follow-ups, whereas assistant turns are substantially longer and
heavier-tailed (median $57$, mean $85$, up to $379$ words), reflecting
detailed explanations and multi-step procedures. This give-and-take
mirrors genuine assistant usage rather than balanced chit-chat, and the
length of assistant turns is what makes single-session recall
(\textsc{ss}) and thematic (\textsc{th}) questions non-trivial.

\paragraph{Coverage is systematic across all scenarios.}
Panel~\textbf{(3)} reports the per-persona breakdown of the eight
question types. The grid is fully populated---every one of the $15$
personas contributes every type---so no capability is evaluated on only
a handful of scenarios. The heatmap also makes the benchmark's
reasoning emphasis visible: multimodal reasoning (\textsc{mr}) is the
largest column for the document-bearing personas (up to $100$--$102$
items each), consistent with the construction pipeline allocating extra
chart- and PDF-grounded questions to personas whose later sessions
include documents.

\paragraph{Sessions are spread irregularly over real time.}
Panel~\textbf{(4)} plots each persona's sessions on a normalized
calendar axis. Rather than being back-to-back, the $239$ sessions are
distributed unevenly---with bursts and multi-week gaps---over spans
ranging from $8$ to $127$ days (a mean of $12.7$ turns per session).
This temporal spread is precisely what gives the timeline-grounded
temporal-reasoning questions (\textsc{tr}) their difficulty, since
``when did $X$ happen'' cannot be answered by turn order alone.


\subsection{Question Types and Synthesis}
\label{sec:appendix-examples}

\begin{figure*}[t]
  \centering
  \includegraphics[width=\linewidth]{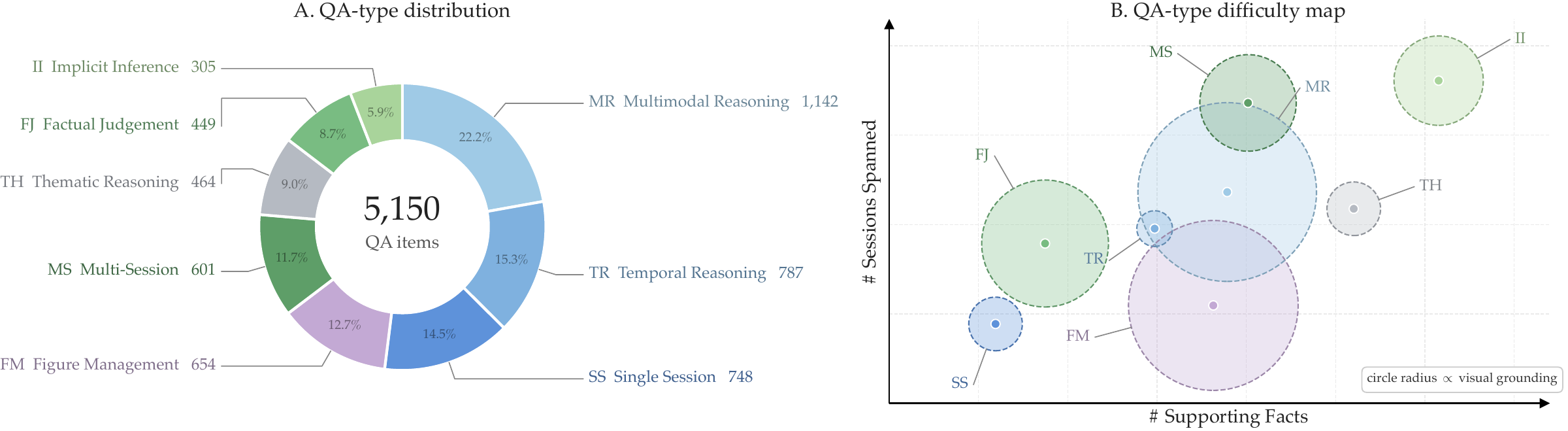}
  \caption{\textbf{QA-type composition and difficulty of \benchmark.}
\textbf{(A)}~\emph{Distribution.} Number and share of the $5{,}150$ QA
items across the eight question types.
\textbf{(B)}~\emph{Difficulty map.} Each type is placed by its average
evidence load ($x$: number of supporting facts) and memory span
($y$: number of distinct sessions the supporting facts span), with
bubble radius growing with the fraction of items requiring visual or PDF
grounding. Colors denote the question type and are shared across both
panels.}
  \label{fig:distribution}
\end{figure*}

Figure~\ref{fig:distribution} summarizes the composition and difficulty
of the $5{,}150$ evaluation items. Panel~\textbf{(A)} shows that the
benchmark is reasoning-heavy: multimodal reasoning (\textsc{mr},
$1{,}142$ items, $22.2\%$), temporal reasoning (\textsc{tr}, $787$,
$15.3\%$), and single-session recall (\textsc{ss}, $748$, $14.5\%$) are
the three largest types, followed by figure management (\textsc{fm},
$654$, $12.7\%$) and multi-session (\textsc{ms}, $601$, $11.7\%$); the
interpreting types---thematic reasoning (\textsc{th}, $464$, $9.0\%$),
factual judgment (\textsc{fj}, $449$, $8.7\%$), and implicit inference
(\textsc{ii}, $305$, $5.9\%$)---are deliberately rarer but the hardest
to construct, since they hinge on unstated context. No single type
dominates, so accuracy on the benchmark cannot be inflated by mastering
one easy category. Panel~\textbf{(B)} places each type by its average
evidence load ($x$: number of supporting facts) and memory span
($y$: number of distinct sessions the evidence spans), with bubble
radius growing with the fraction of items needing visual or PDF
grounding. The types separate along both axes: \textsc{ss} sits in the
low-load, single-session corner, whereas \textsc{ii} and \textsc{ms}
demand the widest memory span and \textsc{mr} and \textsc{fm} carry the
strongest visual grounding (largest bubbles). This spread confirms that
\benchmark{} jointly stresses long-range memory and cross-modal
grounding rather than any single dimension.

\benchmark generates eight typed question categories on
each finalized dataset. Each item carries a question, an ordered list
of answers (the first entry is the gold; the remaining entries are
progressively coarser alternatives used by the graded EM metric;
\textit{see} Section~\ref{sec:eval-metrics}), and round-level
\texttt{supporting\_facts} that identify the dialogue rounds an
evaluator must surface for the model. All types share the same outer
loop: we sample a dialogue context (one or more session windows), fill
a type-specific prompt template, call an LLM to emit a JSON array of
items, validate each item against the schema in
Section~\ref{sec:eval-framework}, and append the surviving items to
\texttt{question.json}; generation resumes from any partial bank until
a per-type target count is met. Below we describe each type along two
axes---\emph{what it tests} and \emph{how it is synthesized}, the
latter giving the sampling strategy and the construction-time
constraint that defines the type. The illustrative items are drawn
verbatim from the \texttt{Alex\_Veterinarian} dataset
(\textit{persona}: a senior veterinary clinician); the corresponding
verbatim prompt templates follow in
Section~\ref{sec:appendix-prompt-templates}.

\paragraph{SS — Single-Session.}
SS questions test detailed recall from a single dialogue session: the
gold answer is grounded in one round, and a model that has not read
that session must not be able to answer correctly.
\emph{Synthesis.} We sample a contiguous window of $4$ to $N$ rounds
from a single session and ask the generator to write questions
answerable \emph{only} from that excerpt. The construction constraint
is locality: the supporting fact must cite exactly one round, and the
gold answer is kept to a short phrase (${\le}3$ words) so the item is
unambiguous and recall-focused.

\paragraph{MS — Multi-Session.}
MS questions require connecting or comparing information across
\emph{at least two} sessions; a question answerable from a single
session alone is rejected at generation time.
\emph{Synthesis.} We sample $2$ to $3$ distinct sessions and draw a
window from each. The generator must write a question whose answer
requires fusing facts from at least two different sessions, so the
supporting facts cite $\ge\!2$ rounds drawn from different sessions
(e.g., \texttt{D2:5, D16:9}).

\paragraph{TR — Temporal Reasoning.}
TR questions hinge on dates, ordering, or time spans drawn from the
event timeline and session headers.
\emph{Synthesis.} We sample $2$ to $4$ sessions and sort them
chronologically; questions hinge on absolute dates, event ordering, or
time-range counts taken from the timeline and session headers. To
absorb formatting variation, the gold answer is stored in multiple
date formats (e.g.,
\texttt{["2023-10-10", "October 10, 2023", "Oct 10 2023"]}) so that
partially correct phrasings receive partial credit under the graded
EM, and the supporting facts include the recap round bearing the
absolute date.

\paragraph{MR — Multimodal Reasoning (chart-grounded).}
MR questions require reading a number, label, rank, or simple
arithmetic relation from a visible chart.
\emph{Synthesis.} We sample a single session centred on image-bearing
rounds; for PDF sessions, the relevant pages are rendered and attached
so the vision model sees them directly. The defining constraint is
that the answer is visible in the attached image or page but
\emph{not} present in the dialogue text, which forces genuine visual
grounding rather than text-only recall. The supporting facts cite the
image-bearing round.

\paragraph{FM — Find-Matching Image.}
FM questions ask the model to identify which image in the dataset
matches a textual description; the answer is the image file name (in
\texttt{img\_\textless number\textgreater} format), testing
cross-modal retrieval over the dataset's flat image folder.
\emph{Synthesis.} We sample a round that exposes at least two candidate
images and ask the generator to write a description that uniquely
matches exactly one of them. The answer list places the matching image
identifier first, followed by the competing candidates (e.g.,
\texttt{["img\_60", "img\_61"]}), so the task is cross-modal retrieval
under hard negatives.

\paragraph{TH — Thematic Reasoning.}
TH questions test recall of multi-step procedures, decision rationales,
or case-management protocols that the assistant introduced during the
dialogue; they typically have a structured gold answer (e.g., a
5-pillar treatment plan) and reward partial recall via the answer-list
gradient.
\emph{Synthesis.} We sample a single session with a wider window ($5$
to $8$ rounds) and ask the generator to bundle at least three decisions
or steps from one case thread into one structured answer, stored
hierarchically (full $\rightarrow$ short $\rightarrow$ keyword) so the
graded metric rewards partial recall of the procedure.

\paragraph{II — Implicit Inference.}
II questions require inferring a fact that is never stated directly, by
composing implicit signals from multiple rounds (e.g., role,
relationship, or motivation); supporting facts cite the rounds that
together justify the inference.
\emph{Synthesis.} We sample up to five sessions with short windows
each. The generator must write a question whose answer is \emph{never
stated outright} but can be inferred by composing $\ge\!2$ clues spread
across the history; the answer is a qualitative noun phrase, and
frequency-count answers are disallowed to keep the item
inference-driven rather than tally-driven. Detailed case study shown in
Appendix~\ref{sec:casestudy}.

\paragraph{FJ — Factual Judgment.}
FJ questions are multiple-choice with one correct option and three
plausible distractors, covering yes/no judgments and
choose-the-better-recommendation decisions, and are scored under EM
with a single-letter gold answer.
\emph{Synthesis.} We sample a multi-session excerpt and embed four
options (A--D) directly in the question text, with distractors drawn
from the same corpus to make them plausible. The generator is
instructed to output only the gold letter, yielding an unambiguous
multiple-choice item.

\subsection{Quality Control.}
\label{sec:quality-control}

To prevent errors and hallucinations from propagating downstream, we
apply quality control at every stage, combining an automatic LLM self-check
with expert verification. For timeline generation, the self-check audits
persona violations, core-event drift, offensive conetents and internal contradictions
(e.g., date-order or mutually exclusive states). For conversation synthesis, we audit at two granularities: a per-chunk
audit scans windows of rounds for inconsistencies, while a cross-chunk
audit catches those spanning chunk boundaries, and each detected flawed round is
rewritten by a repair pass conditioned on neighboring rounds. For the question bank,
a validation pass enforces well-formedness and evidence consistency,
discarding any question whose supporting facts fall outside the sampled
context window so that every retained question is answerable from the
evidence it cites. These edits made across these stages are documented, and a final expert
inspection of the records verifies persona fidelity and core-event
coverage. Prompts and the implement details are given
in Appendix~\ref{sec:appendix-examples}.

\subsection{Persona and Core-Event Specifications}
\label{sec:appendix-persona-spec}

Each persona scenario in \benchmark{} is seeded from a \emph{persona}
(who the user is and what their interactions cover) and a
\emph{core event} (the storyline that the multi-session timeline
elaborates). Table~\ref{tab:persona-core} summarizes both seeds for all
fifteen subsets; these are the inputs to the timeline-construction
prompts described in Section~\ref{sec:appendix-prompt-templates}.

\section{Experimental Details}
\label{sec:appendix-experiments}

\subsection{MLLM Details}
\label{sec:appendix-mllm}

\paragraph{Implementation}
The frontier MLLMs in the main results (Table~\ref{tab:main-results}) 
have no external memory module. They
answer directly from the evidence placed in their context. For a fair
comparison that tests reasoning rather than retrieval, we do not dump
the entire conversation history into the window. Instead, for each
question we follow its annotated supporting facts to the sessions the
answer depends on and supply those sessions together with their
attached images and rendered PDF pages, so every model sees the same
gold evidence. This implementation removes the
retrieval burden and upper-bounds what each model could achieve given
perfect memory.

\paragraph{Long-context buffer baseline.}
Modern language models often support extended context windows ranging
from $128$K to over $1$M tokens, which suggests a simple memory-free
strategy: keep a buffer of the most recent tokens. In a model with a
$128$K-token limit, the agent concatenates incoming chunks until the
total exceeds the window size; once the limit is reached, the earliest
chunks are evicted in a first-in-first-out (FIFO) manner. Such an agent
relies solely on positional recency and assumes the model can attend
effectively over whatever currently fits in its window. We include this
buffer behavior as a reference point for the no-memory regime, in
contrast to the oracle-context feeding above.

\paragraph{Models.}
We evaluate the following frontier closed-source MLLMs as answering
models. \emph{Claude-Opus-4.6}~\citep{anthropic2026opus46} is
Anthropic's flagship reasoning model. \emph{GPT-5.4}~\citep{openai2026gpt54}
is OpenAI's latest general-purpose multimodal model.
\emph{Gemini-3.1-Pro}~\citep{google2026gemini31pro} is Google's multimodal
model with a very long native context window. \emph{GLM-5.1} (Zhipu)%
~\citep{zhipu2026glm51} and \emph{Qwen3.6-Plus} (Alibaba)%
~\citep{alibaba2026qwen36} are strong open-weight-lineage multimodal
models. The arena additionally includes \emph{Doubao-Seed-2.0}
(ByteDance)~\citep{bytedance2026doubao}, \emph{Kimi-k2.5} (Moonshot)%
~\citep{moonshot2026kimi25}, and \emph{Grok-4} (xAI)~\citep{xai2025grok4}.
All API-served models are accessed through OpenAI-compatible1121
remote endpoints and exposed to the same function-calling interface. 
For the full agentic-memory baseline experiments, we additionally evaluate all
memory systems on four answering backbones spanning families and
scales. The lightweight/open-weight group includes Qwen2.5-VL-7B and Qwen2.5-VL-3B%
~\citep{qwen2025vl}, and two proprietary models, GPT-5-nano%
~\citep{openai2026gpt54} and Doubao-Seed-2.0-Pro, to confirm that our findings are not an artifact of a single backbone.

\subsection{Metrics}
\label{sec:appendix-metrics}

We split evaluation into two metric families: \emph{answer-quality}
metrics that score the model's predicted answer against the gold,
and \emph{retrieval-quality} metrics that score how well a
retrieval-augmented baseline surfaces the supporting-fact rounds
annotated in our question bank. Let $\hat{a}$ denote the predicted
answer and $\mathcal{A} = (a_0, a_1, \ldots)$ the ordered gold answer
list, where $a_0$ is the canonical gold (identical to the
\texttt{label} field in the dataset) and each $a_i$ for $i{>}0$ is a
progressively coarser alternative phrasing. Because exact and soft
matching call for different text normalization, we use two operators:
$\mathrm{norm}_{\mathrm{em}}(\cdot)$ is a \emph{sentence-level}
normalization (lowercasing, collapsing runs of whitespace to a single
space, and stripping leading/trailing non-alphanumeric-underscore
characters) used for whole-string equality, while
$\mathrm{tok}(\cdot)$ is the SQuAD-style normalization (lower-casing,
punctuation stripping, article removal) followed by whitespace
tokenization, used for F1 and BLEU-1.

\paragraph{\textbf{Graded Exact Match (EM).}}
We follow long-term-memory benchmarks
\citep{maharana2024locomo,bei2026mem} that use exact-match
scoring on short factual answers, and extend it into a gradient over
the ordered gold list. Under sentence-level normalization we take the
earliest matching position
$i^{*} = \min\{i : \mathrm{norm}_{\mathrm{em}}(\hat{a}) =
\mathrm{norm}_{\mathrm{em}}(a_i)\}$ and assign a discounted score:

{\small
\begin{equation}
  \mathrm{EM}(\hat{a}, \mathcal{A}) =
  \begin{cases}
    2^{-i^{*}} & \text{if } i^{*}\ \text{exists},\\[2pt]
    0 & \text{otherwise.}
  \end{cases}
\end{equation}
}

so predicting the canonical gold scores $1.0$, predicting the
first alternative scores $0.5$, and so on. Compared with
binary EM, this rewards near-correct but less specific phrasings in
proportion to how close they are to the canonical answer, while still
preferring the canonical form.

\paragraph{\textbf{Type-specific EM adaptations.}}
Three question types use specialised rules for EM and its companion
soft metrics. \textbf{(1) Judgement (\textsc{fj}).} Only the graded EM
above enters the main results; token-level F1, BLEU-1, and LLM-J are
not applied to this type (recorded as $0$, with no judge call) to
avoid mixing scales with a non-textual answer form.
\textbf{(2) Image-filename matching (\textsc{fm} whose gold parses to
an \texttt{img\_<n>} identifier).} Instead of the graded EM over
$\mathcal{A}$, we use a binary image-id match against the canonical
gold $a_0$. Writing $g$ for the normalised image id of $a_0$ and
$\mathrm{ids}(\hat{a})$ for the set of all \texttt{img\_<n>} tokens
extracted from the prediction,

{\small
\begin{equation}
  \mathrm{EM}_{\mathrm{img}}(\hat{a}, a_0) =
  \begin{cases}
    1 & g \in \mathrm{ids}(\hat{a}),\\[2pt]
    0 & \text{otherwise.}
  \end{cases}
\end{equation}
}

\noindent This subtype computes no F1, BLEU-1, or LLM-J against $a_0$
(all recorded as $0$), since the shared \texttt{img} prefix would
otherwise inflate token overlap; set-based precision/recall/F1 over
predicted and gold image ids are logged per item for diagnostics only
and do not enter the main table.
\textbf{(3) Text matching (\textsc{fm} with no parseable
\texttt{img\_<n>} in the gold).} EM scores $1.0$ if
$\mathrm{norm}_{\mathrm{em}}(\hat{a}) = \mathrm{norm}_{\mathrm{em}}(a_i)$
for any $i$ and $0$ otherwise (a full-string hit on any equivalent
phrasing earns full credit, with no $2^{-i}$ discount). This subtype
still computes F1, BLEU-1, and LLM-J against $a_0$, but those soft
scores are excluded from the TOTAL macro-average together with all
\textsc{fm} and \textsc{fj} items.

\paragraph{\textbf{Token-Level F1.}}
We use the SQuAD-style token-level F1~\citep{rajpurkar2016squad}
between the prediction and the canonical gold $a_0$. Let
$P = \mathrm{tok}(\hat{a})$ and $G = \mathrm{tok}(a_0)$ be the
predicted and gold token multisets, and $|P \cap G|$ the number of
overlapping tokens (with multiplicity). Then

{\small
\begin{align}
  \mathrm{Prec} &= \frac{|P \cap G|}{|P|},
  \quad
  \mathrm{Rec}  = \frac{|P \cap G|}{|G|}, \notag\\[2pt]
  \mathrm{F1}   &= \frac{2 \cdot \mathrm{Prec} \cdot \mathrm{Rec}}
                       {\mathrm{Prec} + \mathrm{Rec}}.
\end{align}
}

When $P$ and $G$ are both empty we set $\mathrm{F1}=1$; when exactly
one is empty we set $\mathrm{F1}=0$. F1 captures partial lexical
overlap that strict EM ignores and is the standard metric for
short-answer QA evaluation. In the main results, the TOTAL F1 is an
equal-weight macro-average over the items remaining after
\textbf{excluding all \textsc{fm} and \textsc{fj}} questions.

\paragraph{\textbf{BLEU-1.}}
We report unigram BLEU with brevity penalty as introduced
by~\citep{papineni2002bleu}, computed against the canonical gold
$a_0$. Let $c_{\mathrm{clip}}(w) = \min\bigl(c_P(w),\, c_G(w)\bigr)$
denote the clipped count of unigram $w$ in the prediction, where
$c_P$ and $c_G$ are the unigram counts in $P$ and $G$. Then

{\small
\begin{align}
  \mathrm{BP}            &= \min\!\bigl(1,\ \exp(1 - |G|/|P|)\bigr), \notag\\[2pt]
  \mathrm{BLEU\text{-}1} &= \mathrm{BP} \cdot
    \frac{\sum_{w \in P} c_{\mathrm{clip}}(w)}{|P|}.
\end{align}
}

Compared with F1, BLEU-1 imposes an additional penalty on overly short
predictions through the brevity penalty $\mathrm{BP}$; the empty-set
boundary is handled as for F1. Its macro-average rule is the same as
F1's (TOTAL excludes \textsc{fm} and \textsc{fj}).

\paragraph{\textbf{LLM-as-a-Judge (LLM-J).}}

\begin{table*}[h]
\centering
\small
\newcolumntype{Y}{>{\centering\arraybackslash}X}
\newcolumntype{J}{>{\centering\arraybackslash}p{3.4cm}}
\setlength{\tabcolsep}{3pt}
\renewcommand{\arraystretch}{1.05}
\begin{tabularx}{\linewidth}{J|YY|YY|YY}
\toprule
\multirow{2}{*}{\textbf{Judge}}
 & \multicolumn{2}{c|}{\textbf{Text}}
 & \multicolumn{2}{c|}{\textbf{Image}}
 & \multicolumn{2}{c}{\textbf{PDF}} \\
\cmidrule(lr){2-3}\cmidrule(lr){4-5}\cmidrule(lr){6-7}
 & \textbf{Count} & \textbf{Acc.\,(\%)}
 & \textbf{Count} & \textbf{Acc.\,(\%)}
 & \textbf{Count} & \textbf{Acc.\,(\%)} \\
\midrule
Human Verification                     & 50 / 50 & 100 & 50 / 50 & 100 & 50 / 50 & 100 \\
Qwen2.5-VL-3B                          & 52 / 48 & 96  & 50 / 50 & 96  & 48 / 52 & 96  \\
DeepSeek-V3.2                          & 49 / 51 & 95  & 51 / 49 & 97  & 50 / 50 & 98  \\
\rowcolor{skyblue} Qwen2.5-VL-32B (Ours) & 51 / 49 & 97  & 50 / 50 & 100 & 48 / 52 & 98  \\
\bottomrule
\end{tabularx}
\caption{Agreement of each judge with human verification on 50
sampled items per task type. ``Count'' reports the correct\,/\,incorrect
split as scored by that judge; ``Acc.'' reports agreement (\%) with
human verification. Numbers are close across judges, indicating
that LLM-J is stable across judge models.}
\label{tab:judge-verification}
\end{table*}

Following Mem-Gallery~\citep{bei2026mem}, we use
Qwen2.5-VL-32B-Instruct~\citep{bai2025qwen25vl} as an impartial judge
and ask it to score $\hat{a}$ against $a_0$ on a five-level rubric
$\{0, 0.25, 0.5, 0.75, 1.0\}$ covering \emph{incorrect}, \emph{poor /
tangential}, \emph{partial / vague}, \emph{good / minor imperfection},
and \emph{correct / exact}. The verbatim judge prompt is reproduced in
Appendix~\ref{sec:appendix-prompt-templates}. The judge may only emit
one of the five levels; a parse failure is scored $0$. LLM-J captures
semantic equivalence that strict-token metrics miss (e.g., paraphrased
correct answers or equivalent entity mentions). It is not applied to
\textsc{fj} or to the \textsc{fm} image subtype (no judge call,
\texttt{llm\_score} set to $0$); the remaining types, including the
\textsc{fm} text subtype, are still scored against $a_0$, but the
TOTAL LLM-J macro-averages only after \textbf{excluding all
\textsc{fm} and \textsc{fj}} items.

A graded LLM-as-a-Judge can in principle be sensitive to borderline
correctness and answer phrasing. To check that our scoring is not
biased by the choice of judge, we conduct a small-scale human
verification on three reasoning-heavy task types — text-only,
image-grounded, and PDF-grounded reasoning. For each task type we manually
sample 50 correct answered and 50 wrong model generations together with their queries and gold
answers, and re-score them with two alternative judges: a smaller
in-family model (Qwen2.5-VL-3B-Instruct) and an out-of-family
strong model (DeepSeek-V3.2). We then measure agreement with human verification on the same items. We binarize each judgment by treating scores of $0.75$--$1.0$ as correct
and $0$--$0.5$ as incorrect, and report both the judgment counts and
the resulting accuracy against the human-annotated sample.

Results are summarized in Table~\ref{tab:judge-verification}. The
default judge (Qwen2.5-VL-32B-Instruct) achieves high agreement with
human verification across all three task types, and both alternative
judges produce closely matching accuracy. The consistency across
judge families indicates that our LLM-J scores are not driven by an
idiosyncratic preference of a single judge model, supporting the
robustness of the LLM-as-a-Judge protocol used throughout the paper.
Variance is further reduced in the main results (Table~\ref{tab:QA performance qwen 7b})
by averaging over repeated runs with independent judging.

\paragraph{\textbf{Overall Score.}}
To rank systems with a single number, we report a composite
\emph{Overall Score} that combines the four answer-quality metrics,
\begin{equation}
\small
\mathrm{Overall} = 0.5\,\mathrm{LLM\text{-}J}
  + 0.5\cdot\tfrac{1}{3}\bigl(\mathrm{EM}+\mathrm{F1}+\mathrm{BLEU\text{-}1}\bigr),
\label{eq:overall}
\end{equation}
where each constituent is the per-type score macro-averaged over
question types weighted by their item counts. The half weight on LLM-J,
with the remaining half split equally among EM, F1, and BLEU-1, is
deliberate: \benchmark{} answers are often compositional---bundling
several elements such as multi-step procedures or cross-modal
facts---rather than short spans, so the surface-matching metrics
(EM, F1, BLEU-1) only partially credit a correct-but-rephrased
response. We therefore give the semantically-aware LLM-J the largest
single weight, while still retaining the lexical metrics as a check
against judge leniency. Only the Overall Score and the three component
metrics EM, F1, and LLM-J appear in the main paper; BLEU-1 enters
solely through this composite.

\begin{figure*}[t]
  \centering
  \includegraphics[width=\linewidth]{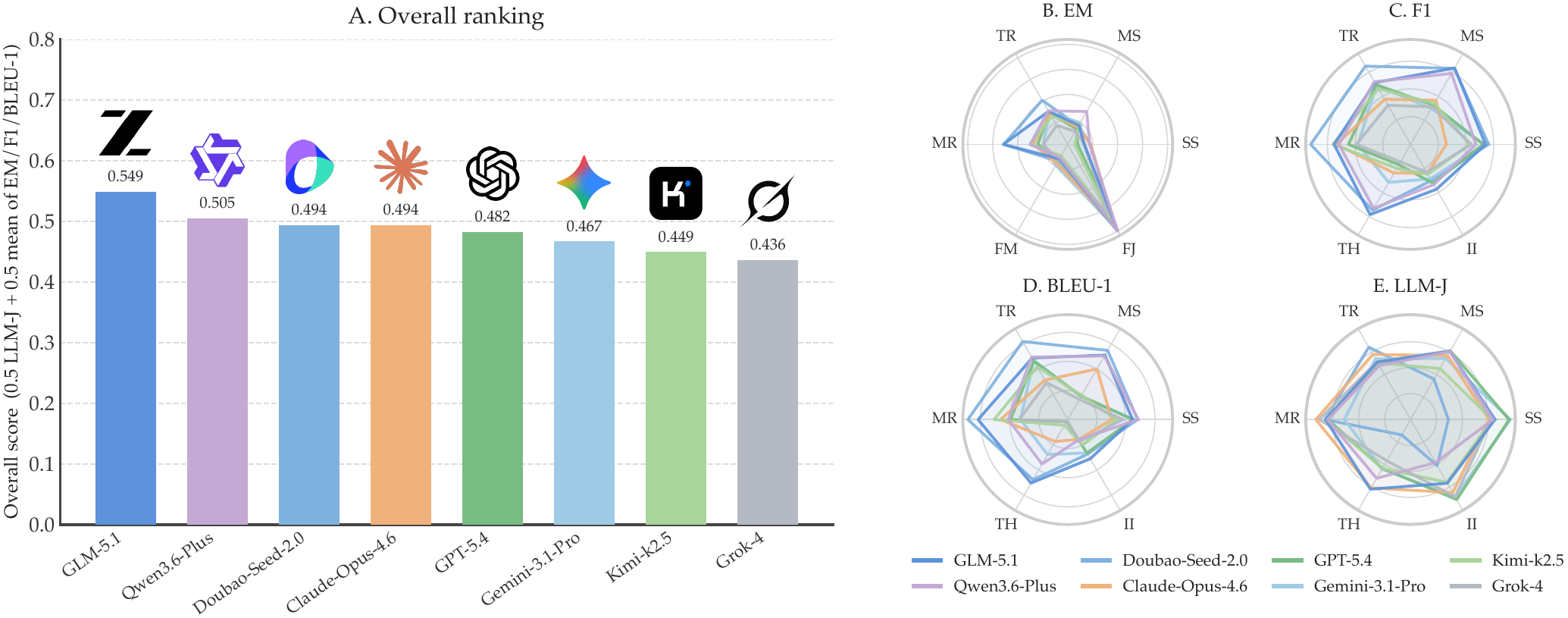}
  \caption{\textbf{Frontier-MLLM arena on \benchmark.} We evaluate eight
  frontier closed-source multimodal LLMs---GLM-5.1, Qwen3.6-Plus, Claude-Opus-4.6,
  Doubao-Seed-2.0, GPT-5.4, Gemini-3.1-Pro, Kimi-k2.5 and Grok-4---as the
  answering model on \benchmark, scoring every model under the four metrics
  (EM, F1, BLEU-1 and LLM-J).
  \textbf{Left:} overall ranking, where each model is
  scored under a unified standard combining the four
  metrics.
  \textbf{Right:} per-metric breakdown, one radar per metric, where each
  hexagon spans the six question types that the metric is defined on and each
  model is drawn as one colored polygon.}
  \label{fig:modelarena}
\end{figure*}

\subsection{Baseline Descriptions}
\label{sec:appendix-baselines}

In this appendix section, we provide detailed descriptions of the advanced memory methods 
evaluated in Table~\ref{tab:QA performance qwen 7b}. To enable a fair comparison between text-only and multimodal
regimes, the text-only group is provided with high-quality image
captions generated by Qwen2.5-VL-32B as a text-equivalent of the
visual evidence the multimodal group sees.

\paragraph{Base (Full / Session, Text / Multimodal).} The four
\textit{Base} configurations feed prior dialogue directly into the
backbone without any retrieval or external memory module.
\textit{Base (Full)} concatenates the entire dialogue history,
including all multimodal memory, into the context. To respect the
backbone's context-token limit, it estimates the token consumption of
each image using a predefined per-image token cost and truncates the
input once the running total would exceed the limit, so the longest
histories are clipped rather than dropped. \textit{Base (Session)}
restricts the context to the single session
in which the question is grounded; each is run in two regimes, where
the text-only regime supplies captions of attached visuals while the
multimodal regime supplies the original images (capped at ten per
question). Comparing across the four configurations isolates the
contribution of cross-session context (Full vs.\ Session) and of
visual input (text vs.\ multimodal).

\paragraph{NaiveRAG}\citep{lewis2020rag} encodes prior
dialogue turn with a dense sentence encoder, indexes the resulting
vectors in a FAISS store, and at inference time retrieves the
top-$k$ chunks whose embeddings are nearest to the question.
The retrieved chunks are concatenated into the model's prompt
without further reranking or summarization.
\url{https://github.com/gusye1234/nano-graphrag}

\paragraph{A-Mem}\citep{xu2025amem} is an agentic memory system
for LLM agents that dynamically organises memories by drawing
inspiration from the Zettelkasten method to build interconnected
knowledge networks. When adding a new memory, A-Mem generates
structured notes with contextual descriptions, keywords, and tags,
then identifies connections with historical memories to establish
links, while enabling memory evolution through updates to existing
representations as new information is integrated.
\url{https://github.com/WujiangXu/A-mem}

\paragraph{Mem0 / Mem0 (with vision).}~\citep{chhikara2025mem0} is
a production-style memory layer that extracts atomic facts from
each dialogue turn, deduplicates them against the existing store via
LLM-judged equivalence, and retrieves a top-$K$ subset at inference
time using dense semantic search; an optional graph variant adds
typed entity relations between extracted facts. We evaluate two
configurations: the text-only \textit{Mem0} stores fact strings
derived from captions of attached visuals, while \textit{Mem0 (with
vision)} additionally embeds the original images alongside the
textual facts and retrieves both modalities through the same
consolidation-and-recall pipeline.
\url{https://github.com/mem0ai/mem0}

\paragraph{MemoryOS.}~\citep{kang2025memory} treats long-term memory as
an operating system over a unified abstraction (\textit{MemCube})
that wraps parametric, activation, and plaintext memory under one
lifecycle manager covering allocation, eviction, migration, and
access control. At inference time, queries are dispatched through a
scheduler that surfaces the most relevant MemCubes across the three
tiers; we use the plaintext-memory variant for fair comparison
against the other text-based baselines.
\url{https://github.com/BAI-LAB/MemoryOS}

\paragraph{UniversalRAG.}~\citep{yeo2025universalrag} maintains a
modality- and granularity-aware corpus in which text passages,
images, and clip-level chunks are each indexed separately, and
trains a lightweight router that selects the modality and
granularity most likely to contain the answer for a given query.
Retrieval is then executed within the selected sub-index, and the
top-$K{=}10$ results are concatenated into the prompt.
\url{https://github.com/wgcyeo/UniversalRAG}

\paragraph{RAG-Anything.}~\citep{guo2025raganything} is an
all-in-one RAG framework that indexes heterogeneous content---text,
images, tables, and equations---under a single shared embedding
space and supports cross-modal top-$K{=}10$ retrieval through a
unified query encoder. We use the framework's default cross-modal
configuration without task-specific tuning.
\url{https://github.com/HKUDS/RAG-Anything}

\paragraph{MIRIX.}~\citep{wang2025mirix} is a multi-agent memory
system that partitions long-term memory into six specialised
stores---core, episodic, semantic, procedural, resource, and
knowledge-vault---each managed by a dedicated agent that decides
whether incoming content belongs to its store. At inference time, a
router agent dispatches the query to the relevant store(s) and
aggregates the retrieved items before prompting the backbone.
\url{https://github.com/Mirix-AI/MIRIX}

\paragraph{MemVerse.}~\citep{liu2025memverse} is a multimodal
memory layer for lifelong-learning agents that maintains visual and
textual memories under explicit lifecycle operations---store,
recall, update, and forget---each implemented as an LLM-driven
policy over an embedding-indexed multimodal store. New observations
trigger update or store operations, while retrieval at inference
time issues a recall over the indexed embeddings.
\url{https://github.com/KnowledgeXLab/MemVerse}

\paragraph{NGM (Neural Graph Memory).}~\citep{fisher2025ngm}
represents long-term memory as a heterogeneous graph in which
nodes correspond to dialogue facts and visual entities and edges
encode co-occurrence, temporal, and cross-modal relations. New
turns are converted into subgraphs that are merged into the global
graph; retrieval performs neighbourhood expansion from query-anchor
nodes and returns the induced subgraph as context.
\url{https://github.com/StuckInTheNet/Neural-Graph-Memory-NGM}

\begin{figure}[t]
  \centering
  \includegraphics[width=\linewidth]{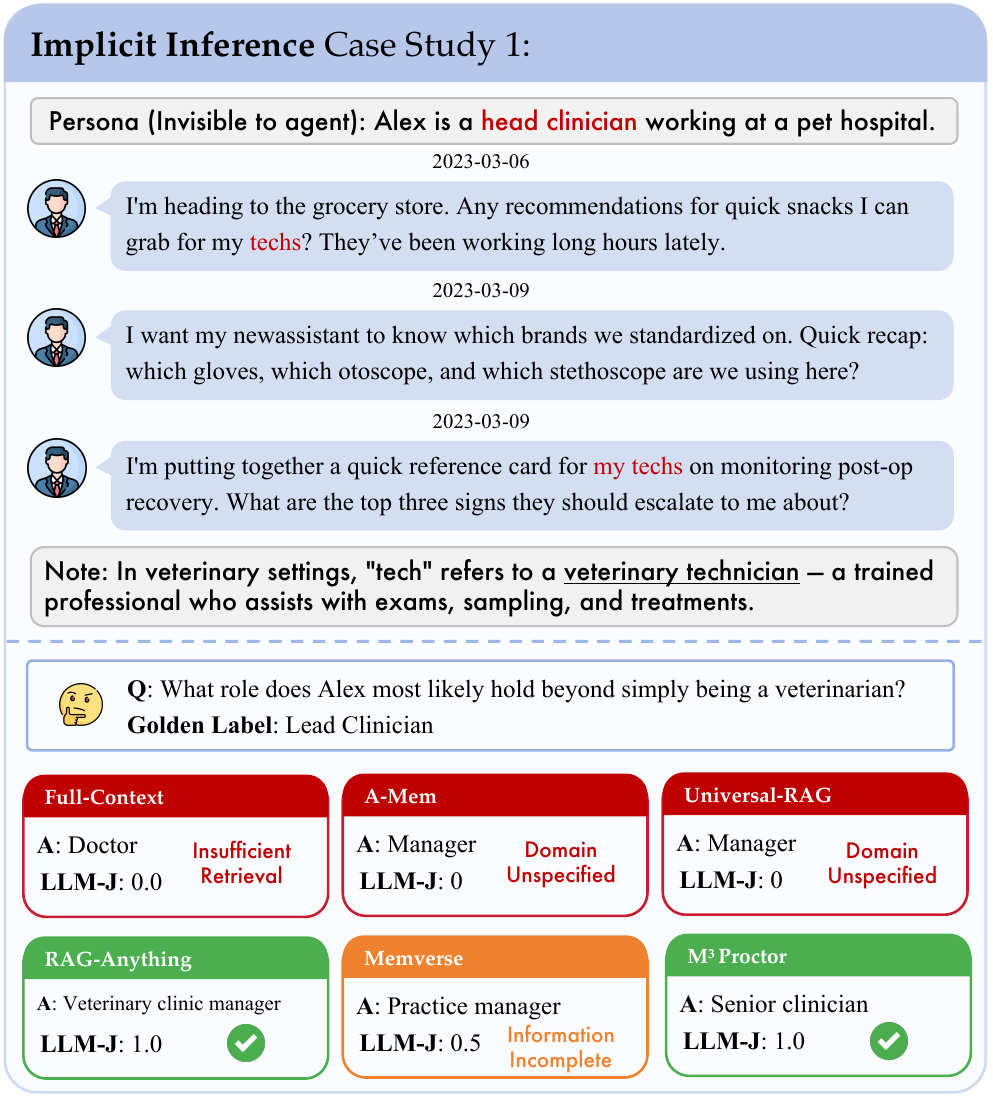}
      \caption{Case study with Implicit Inference Example 1 (Alex, Veterinarian).}
  \label{fig:case1}
\end{figure}

\begin{figure}[t]
  \centering
  \includegraphics[width=\linewidth]{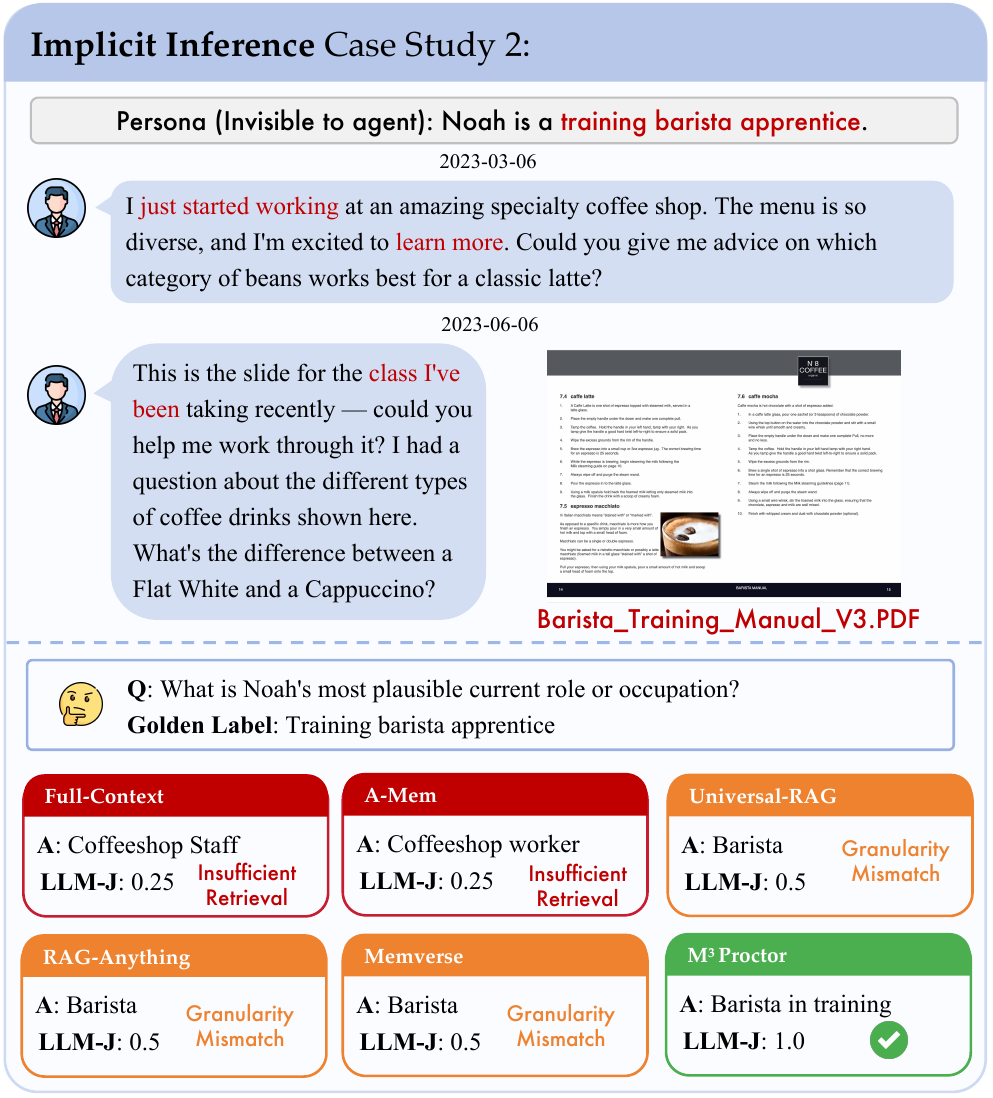}
  \caption{Case study with Implicit Inference Example 2 (Noah, Barista Apprentice).}
  \label{fig:case2}
\end{figure}

\section{Extended Experiments}
\label{sec:appendix-extend}

\subsection{Alternative MLLM Backbone}
\label{sec:backboneaaa}

For space, the main-paper table reports a condensed view of the
results---two metrics per type and a single composite score. Here we
provide the \emph{complete} per-type, four-metric breakdown
(EM, F1, BLEU-1, and LLM-J) for every memory system, and we further
repeat the full evaluation across \emph{four answering backbones} of
different families and scales to confirm that our findings are not an
artifact of one backbone: two open-source MLLMs
(Qwen2.5-VL-7B, Table~\ref{tab:QA performance qwen 7b}; and the smaller
Qwen2.5-VL-3B, Table~\ref{tab:QA performance qwen 3b}) and two strong
proprietary models (GPT-5-Nano, Table~\ref{tab:QA performance gpt};
and Doubao-Seed-2.0-Pro, Table~\ref{tab:QA performance doubao}). All
agentic-memory systems are deployed identically on each backbone, so
differences within a table reflect memory design and differences across
tables reflect the backbone.

\paragraph{Findings are consistent across backbones.}
The conclusions drawn in the main paper hold on all four backbones.
\textbf{(1) Memory is essential.} The memoryless \emph{Full Context}
baseline sits far below every memory system on each backbone (overall
LLM-J of only $0.14$--$0.33$), confirming that raw long-context access
cannot substitute for structured retrieval.
\textbf{(2) Multimodal memory beats text-only memory.} On every
backbone the multimodal systems occupy the top of the table while the
text-only group trails (e.g., on Qwen2.5-VL-7B the best text-only system
reaches $0.44$ LLM-J versus $0.53$--$0.58$ for the multimodal group),
reaffirming that \benchmark{}'s questions genuinely require visual
evidence.
\textbf{(3) \method leads regardless of backbone.} \method attains the
best overall LLM-J on all four backbones---$0.582$ (Qwen2.5-VL-7B),
$0.476$ (Qwen2.5-VL-3B), $0.597$ (GPT-5-Nano), and $0.552$
(Doubao-Seed-2.0-Pro)---ahead of the strongest prior system (MIRIX) in
every case, with the gain again concentrated on the cross-modal and
implicit-intent types. Backbone scale shifts the absolute numbers (the
$3$B backbone lowers all systems, and the proprietary backbones raise
the whole field) but never the ordering: \method on the modest $3$B
backbone ($0.476$) still outperforms most baselines run on the larger
$7$B backbone, indicating that its advantage stems from
modality-aware evidence management rather than backbone capacity.

We further ask how far the current frontier of closed-source MLLMs is
from solving \benchmark on its own. The main paper already reports five
such models alongside the memory systems; here we broaden the panel
into a head-to-head arena of \emph{eight} leading models---GLM-5.1,
Qwen3.6-Plus, Claude-Opus-4.6, Doubao-Seed-2.0, GPT-5.4, Gemini-3.1-Pro,
Kimi-k2.5, and Grok-4---each used as the answering model and scored
under all four metrics
(Figure~\ref{fig:modelarena}). To isolate \emph{reasoning}
ability from long-context handling, we do not run these models over the
full conversation history. Instead, for each question we use its
annotated supporting facts to locate the exact session(s)---together with
their attached images and PDF pages---that the answer depends on, and
provide only that evidence to the model. This oracle-context setting
removes the retrieval burden entirely and upper-bounds what each model
could achieve given perfect memory, so any remaining error reflects a
reasoning or grounding limitation rather than a failure to recall. To
summarise each model with a single indicator, we additionally report a
unified score that weights semantic correctness against lexical
overlap: $0.5\,\text{LLM-J} + 0.5\,\text{mean(EM, F1, BLEU-1)}$.
Because many answers in \benchmark{} are compositional---bundling
several elements such as multi-step procedures or cross-modal
facts---rather than short spans, the surface-matching metrics (EM, F1,
BLEU-1) only partially credit a correct-but-rephrased response; we
therefore give the semantically-aware LLM-J the largest single weight
($0.5$) and let the three lexical metrics share the other half.

\paragraph{Even frontier MLLMs are far from solving \benchmark{}.}
Under this favourable setting, no model exceeds an overall score of
$0.55$: the eight models cluster tightly between $0.445$ and $0.550$
(Figure~\ref{fig:modelarena}A), with GLM-5.1 narrowly on top and the
remaining models---including the latest GPT, Gemini, Claude, and
Grok---separated by only a few points. The arena is also not won by a
single model: the per-metric radars (Figure~\ref{fig:modelarena}B--E)
show different leaders on different axes, and every model follows the
same uneven profile---relatively strong on single-session recall
(\textsc{ss}) and factual judgement (\textsc{fj}) but dropping sharply
on multimodal reasoning (\textsc{mr}) and the interpreting types
(thematic \textsc{th} and implicit inference \textsc{ii}), where the
answer must be composed across modalities or inferred from unstated
context. Two observations follow. First, even when the gold evidence is
handed to the model directly, frontier MLLMs still fail on a large
fraction of items, exposing genuine deficits in cross-modal grounding
and implicit-intent reasoning rather than mere context-length limits;
the consistency of this weakness across eight independent models shows
it is a property of the task, not of any one system. Second, the
oracle-context score is an \emph{upper bound} that a deployed agent
without access to gold supporting facts cannot reach; closing the gap
to it requires a memory system that surfaces the right multimodal
evidence on its own, which motivates the retrieval-augmented methods
studied in the main paper.

\subsection{Full Context Evaluation}
\label{app:fullcontext}

\begin{figure}[t]
  \centering
  \includegraphics[width=\linewidth]{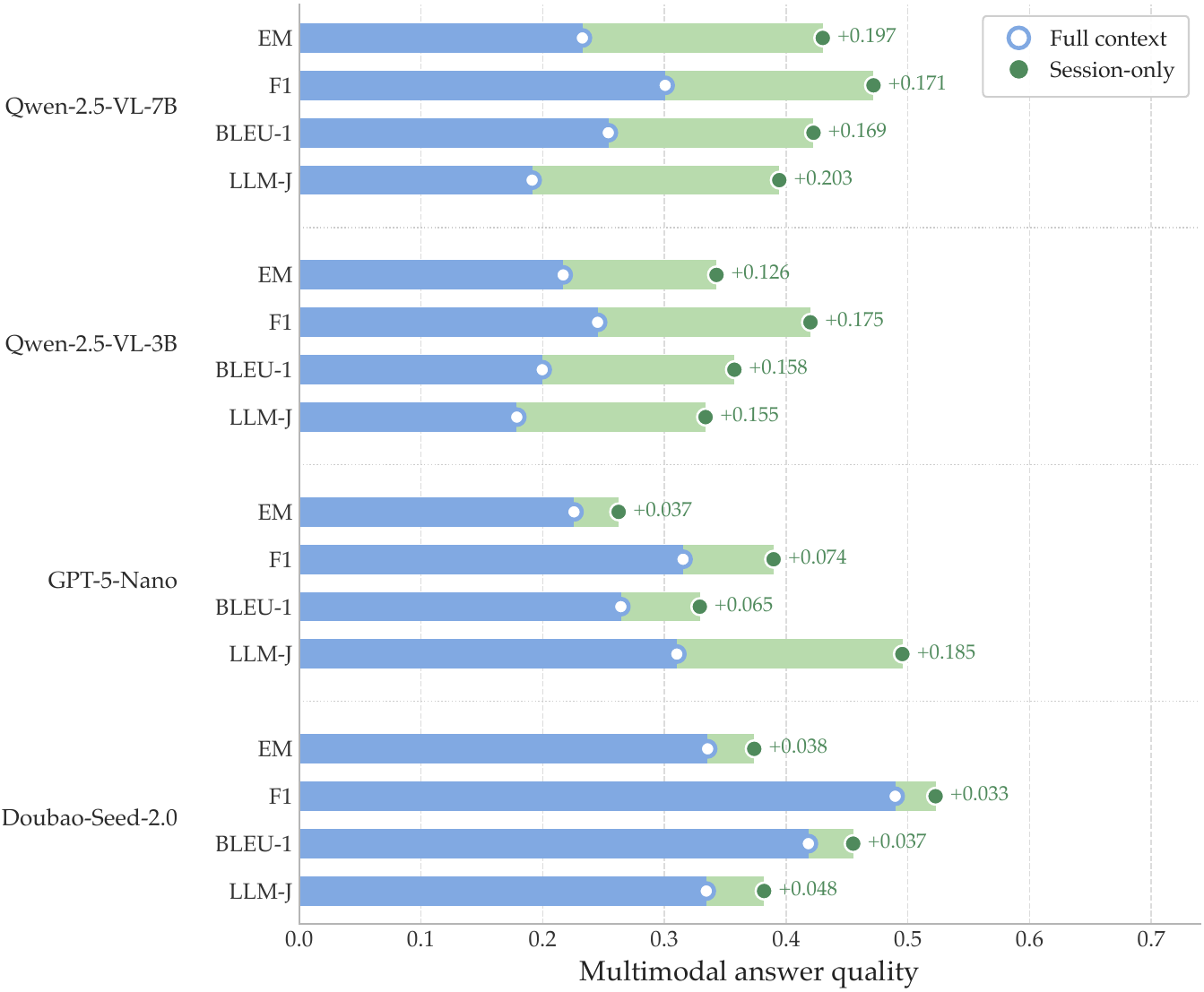}
  \caption{\textbf{The full history adds context and retrieval pressure.}
For each backbone and metric in the multimodal regime, we compare
feeding only the gold supporting session(s) (\emph{Session-only},
retrieval relieved) against feeding the entire conversation
(\emph{Full context}). Bars are overlapped on a shared baseline; the
green tail and its label give the
\emph{Session-only}$-$\emph{Full context} gap, i.e.\ the accuracy lost
to processing the full history.}
  \label{fig:session-vs-full}
\end{figure}

The main paper evaluates each backbone in a \emph{session-only}
setting: for every question we feed only the gold supporting
session(s), which relieves the model of retrieval and lets us measure
its raw answering ability in isolation. A natural question is whether
this simplification is necessary---could a backbone simply ingest the
\emph{full} conversation and find the evidence itself? To answer this,
we re-run the same backbones with the entire history in context and
compare the two regimes in the multimodal setting
(Figure~\ref{fig:session-vs-full}), where the gap reported is
\emph{Session-only}$-$\emph{Full context}.

\paragraph{Feeding the full history imposes both context and retrieval
pressure.} All sixteen comparisons (four backbones $\times$ four
metrics) favour the session-only setting: handing a model the whole
conversation \emph{lowers} answer quality in every case, because it
must now both withstand a much longer context and locate the relevant
rounds amid many distractors. The penalty is largest on the open
Qwen-2.5-VL backbones---up to $+0.197$/$+0.203$ EM/LLM-J on the $7$B
model and $+0.126$ to $+0.175$ across metrics on the $3$B model, around
a fifth of their score. The proprietary models tolerate the longer
input better but still degrade throughout: Doubao-Seed-2.0 by
$+0.033$ to $+0.048$, and GPT-5-Nano by only $+0.037$ to $+0.074$ on
the lexical metrics yet $+0.185$ on LLM-J, showing the extra context
chiefly eroded the \emph{semantic} quality of its answers. Because even
frontier MLLMs lose accuracy once retrieval and context pressure are
added back, the full-history regime is not a viable substitute for
structured memory---which is precisely why the main paper isolates
answering ability with the session-only setting and why a memory that
surfaces the right evidence on its own is needed in deployment.


\subsection{Case Study: Implicit Inference}
\label{sec:casestudy}

Figures~\ref{fig:case1} and~\ref{fig:case2} illustrate why
implicit-inference questions expose the limits of current systems. In
both, the user's true role is never stated and must be composed from
scattered cues---Alex repeatedly instructing his \emph{techs} and
standardizing clinic equipment (Figure~\ref{fig:case1}), or Noah
describing himself as newly hired and studying a training manual
(Figure~\ref{fig:case2}). Systems fail in distinct ways. Memoryless and
text-only baselines (Full-Context, A-Mem) retrieve too little context
and answer generically (``Doctor'', ``Manager'', ``Coffeeshop staff''),
while multimodal baselines that retrieve more evidence still misjudge
the granularity of the inference, over- or under-specifying the role
(``Veterinary clinic manager'', plain ``Barista''). \method, in
contrast, both surfaces the dispersed supporting evidence and calibrates
the inference to the right level (``Senior clinician'', ``Barista in
training''), matching the gold label. These cases show that answering
correctly requires not just recalling evidence but interpreting
unstated, domain-grounded intent that prior systems systematically miss.

\section{Persona and Core-Event Specifications}
\label{sec:appendix-persona-spec}

\begin{table*}[t!]
\small
\centering
\caption{QA performance of \benchmark on \textbf{Qwen-2.5-VL-7B}. For each metric, the best and second-best scores are marked in \textbf{bold} and with \underline{underline}, respectively.}
\vspace{-2mm}

\newcolumntype{C}{>{\scriptsize\centering\arraybackslash}X}
\newcolumntype{M}{>{\scriptsize\centering\arraybackslash}X}
\newcolumntype{W}{>{\scriptsize\centering\arraybackslash}p{1.8cm}}
\setlength{\tabcolsep}{3pt}
\renewcommand{\arraystretch}{0.6}
\setlength{\fboxsep}{1pt}

\begin{tabularx}{\linewidth}{W|M|CCCCCCCC|C}

\toprule
\multicolumn{2}{c|}{\raisebox{-0.2ex}{\includegraphics[height=1.1em]{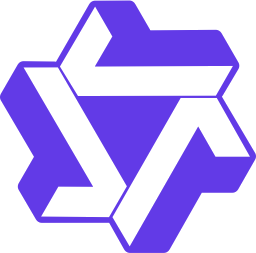}}\,\textbf{Qwen-2.5-VL-7B}} & \textbf{SS} & \textbf{MS} & \textbf{TR} & \textbf{MR} & \textbf{FM} & \textbf{FJ} & \textbf{TH} & \textbf{II} & \textbf{Avg.}\\
\midrule
\multicolumn{11}{c}{\cellcolor{Secondary}\textit{\textbf{Text-Only}}} \\
\midrule

\multirow{4}{*}{\makecell{Full Context\\(without memory)}}
  & EM     & 0.1638 & 0.1148 & 0.1865 & 0.1067 & 0.0596 & 0.5189 & -- & -- & 0.1671 \\
  & F1     & 0.3342 & 0.2598 & 0.2947 & 0.1801 & -- & -- & 0.1458 & 0.1572 & 0.2370 \\
  & BLEU-1 & 0.2898 & 0.1885 & 0.2752 & 0.1603 & -- & -- & 0.0677 & 0.1112 & 0.1964 \\
  & LLM-J  & 0.1938 & 0.1592 & 0.2407 & 0.1229 & -- & -- & 0.0123 & 0.0263 & 0.1443 \\
\midrule
\multirow{4}{*}{\makecell{NaiveRAG}}
  & EM     & 0.2507 & 0.1849 & 0.3613 & 0.1522 & 0.2569 & 0.7840 & -- & -- & 0.2914 \\
  & F1     & 0.4430 & 0.3117 & 0.2349 & 0.2080 & -- & -- & 0.1321 & 0.1263 & 0.2572 \\
  & BLEU-1 & 0.3945 & 0.2323 & 0.2151 & 0.1815 & -- & -- & 0.0373 & 0.0798 & 0.2107 \\
  & LLM-J  & 0.6547 & 0.4429 & 0.4167 & 0.2572 & -- & -- & 0.2806 & 0.3202 & 0.3967 \\
\midrule
\multirow{4}{*}{\makecell{A-Mem}}
  & EM     & \textbf{0.3347} & \underline{0.2887} & 0.1532 & 0.1844 & 0.2875 & 0.9243 & -- & -- & 0.3100 \\
  & F1     & \textbf{0.5921} & 0.3998 & 0.1298 & 0.2403 & -- & -- & 0.1361 & 0.1227 & 0.2867 \\
  & BLEU-1 & \textbf{0.5246} & 0.2996 & 0.1255 & 0.2145 & -- & -- & 0.0321 & 0.0509 & 0.2339 \\
  & LLM-J  & \underline{0.7247} & 0.5345 & 0.1697 & 0.3727 & -- & -- & 0.3493 & 0.2872 & 0.4132 \\
\midrule
\multirow{4}{*}{\makecell{Mem0\\(w/o vision)}}
  & EM     & 0.2947 & \textbf{0.2950} & 0.1893 & 0.1765 & 0.1544 & 0.9243 & -- & -- & 0.2886 \\
  & F1     & 0.5343 & 0.3837 & 0.1361 & 0.2471 & -- & -- & 0.1346 & 0.0962 & 0.2746 \\
  & BLEU-1 & 0.4735 & 0.2895 & 0.1341 & 0.2193 & -- & -- & 0.0310 & 0.0452 & 0.2254 \\
  & LLM-J  & 0.6957 & 0.5448 & 0.1853 & 0.3787 & -- & -- & 0.3843 & 0.3000 & 0.4191 \\
\midrule
\multirow{4}{*}{\makecell{MemoryOS}}
  & EM     & \underline{0.3027} & 0.2414 & 0.1453 & 0.1508 & 0.3150 & \underline{0.9733} & -- & -- & 0.2970 \\
  & F1     & 0.5410 & 0.3399 & 0.1248 & 0.2146 & -- & -- & 0.1274 & 0.1162 & 0.2587 \\
  & BLEU-1 & 0.4779 & 0.2522 & 0.1235 & 0.1855 & -- & -- & 0.0281 & 0.0459 & 0.2088 \\
  & LLM-J  & 0.7047 & 0.5145 & 0.3977 & 0.3299 & -- & -- & 0.3327 & 0.3175 & 0.4392 \\

\midrule
\multicolumn{11}{c}{\cellcolor{Secondary}\textit{\textbf{Multimodality}}} \\
\midrule

\multirow{4}{*}{\makecell{Full Context\\(without memory)}}
  & EM     & 0.1587 & 0.1372 & 0.2258 & 0.2182 & 0.1361 & 0.6726 & -- & -- & 0.2326 \\
  & F1     & 0.3703 & 0.3247 & 0.3375 & 0.2982 & -- & -- & 0.1825 & 0.1752 & 0.3006 \\
  & BLEU-1 & 0.3122 & 0.2326 & 0.3201 & 0.2774 & -- & -- & 0.1018 & \underline{0.1254} & 0.2539 \\
  & LLM-J  & 0.1976 & 0.1958 & 0.2951 & 0.2255 & -- & -- & 0.0211 & 0.0294 & 0.1913 \\
\midrule
\multirow{4}{*}{\makecell{Universal\\-RAG}}
  & EM     & 0.2914 & 0.1777 & 0.3002 & 0.2035 & 0.5367 & 0.8241 & -- & -- & 0.3457 \\
  & F1     & \underline{0.5858} & \textbf{0.4349} & 0.3903 & 0.2938 & -- & -- & 0.2691 & 0.1586 & 0.3745 \\
  & BLEU-1 & \underline{0.5174} & \textbf{0.3165} & 0.3798 & 0.2621 & -- & -- & 0.1517 & 0.0894 & 0.3146 \\
  & LLM-J  & \textbf{0.7328} & \textbf{0.5769} & 0.4431 & 0.4369 & -- & -- & \underline{0.4872} & 0.5559 & 0.5283 \\
\midrule
\multirow{4}{*}{\makecell{RAG\\-Anything}}
  & EM     & 0.2977 & 0.1308 & 0.3597 & 0.2844 & 0.4572 & 0.8508 & -- & -- & 0.3630 \\
  & F1     & 0.5849 & 0.3940 & \underline{0.4353} & 0.3863 & -- & -- & \underline{0.2782} & \underline{0.1873} & \textbf{0.4063} \\
  & BLEU-1 & 0.5137 & 0.2881 & \underline{0.4193} & 0.3562 & -- & -- & \underline{0.1673} & 0.1041 & \underline{0.3468} \\
  & LLM-J  & 0.6418 & 0.5474 & 0.4465 & 0.4970 & -- & -- & 0.4390 & \underline{0.6201} & 0.5241 \\
\midrule
\multirow{4}{*}{\makecell{Mem0\\(with vision)}}
  & EM     & 0.2547 & 0.2727 & 0.2027 & 0.3332 & 0.1942 & 0.9176 & -- & -- & 0.3272 \\
  & F1     & 0.4798 & \underline{0.4181} & 0.1511 & 0.4459 & -- & -- & 0.1485 & 0.1286 & 0.3327 \\
  & BLEU-1 & 0.4288 & \underline{0.3058} & 0.1473 & 0.4140 & -- & -- & 0.0356 & 0.0548 & 0.2783 \\
  & LLM-J  & 0.6457 & 0.5548 & 0.1837 & 0.5607 & -- & -- & 0.3527 & 0.5973 & 0.4811 \\
\midrule
\multirow{4}{*}{\makecell{MemVerse}}
  & EM     & 0.2451 & 0.1411 & 0.2606 & \underline{0.3589} & 0.5474 & 0.9042 & -- & -- & \underline{0.3760} \\
  & F1     & 0.4850 & 0.3895 & 0.3229 & \underline{0.4591} & -- & -- & 0.2518 & 0.1849 & 0.3826 \\
  & BLEU-1 & 0.4243 & 0.2805 & 0.3146 & \underline{0.4293} & -- & -- & 0.1485 & 0.0933 & 0.3265 \\
  & LLM-J  & 0.6406 & 0.5662 & 0.4096 & 0.5613 & -- & -- & 0.4223 & 0.6048 & 0.5345 \\
\midrule
\multirow{4}{*}{\makecell{NGM}}
  & EM     & 0.1808 & 0.1448 & 0.1198 & \textbf{0.3894} & \textbf{0.6208} & 0.9042 & -- & -- & 0.3591 \\
  & F1     & 0.3559 & 0.3453 & 0.1716 & \textbf{0.4708} & -- & -- & 0.2288 & 0.1807 & 0.3231 \\
  & BLEU-1 & 0.3121 & 0.2496 & 0.1650 & \textbf{0.4511} & -- & -- & 0.1398 & 0.0996 & 0.2777 \\
  & LLM-J  & 0.5266 & 0.4660 & 0.1601 & 0.5479 & -- & -- & 0.3905 & 0.6060 & 0.4427 \\
\midrule
\multirow{4}{*}{\makecell{MIRIX}}
  & EM     & 0.2959 & 0.0688 & \underline{0.3980} & 0.2900 & 0.3807 & 0.8664 & -- & -- & 0.3527 \\
  & F1     & 0.4954 & 0.2853 & \textbf{0.4564} & 0.4076 & -- & -- & \textbf{0.2999} & 0.1054 & 0.3800 \\
  & BLEU-1 & 0.4467 & 0.1486 & \textbf{0.4520} & 0.3694 & -- & -- & \textbf{0.2572} & 0.0481 & 0.3299 \\
  & LLM-J  & 0.6378 & 0.4875 & \textbf{0.4949} & \underline{0.5775} & -- & -- & \textbf{0.5917} & 0.5375 & \underline{0.5578} \\
\midrule
\multirow{4}{*}{\makecell{\small\textbf{\method}\\\scriptsize(Our Method)}}
  & \cellcolor{skyblue}EM & \cellcolor{skyblue}{0.2541} & \cellcolor{skyblue}{0.1395} & \cellcolor{skyblue}{\textbf{0.4042}} & \cellcolor{skyblue}{0.3572} & \cellcolor{skyblue}{\underline{0.5688}} & \cellcolor{skyblue}{\textbf{0.9844}} & \cellcolor{skyblue}-- & \cellcolor{skyblue}-- & \cellcolor{skyblue}{\textbf{0.4140}} \\
  & \cellcolor{skyblue}F1 & \cellcolor{skyblue}{0.4993} & \cellcolor{skyblue}{0.3697} & \cellcolor{skyblue}{0.3952} & \cellcolor{skyblue}{0.4305} & \cellcolor{skyblue}-- & \cellcolor{skyblue}-- & \cellcolor{skyblue}{0.2202} & \cellcolor{skyblue}{\textbf{0.3021}} & \cellcolor{skyblue}{\underline{0.3935}} \\
  & \cellcolor{skyblue}BLEU-1 & \cellcolor{skyblue}{0.4378} & \cellcolor{skyblue}{0.2466} & \cellcolor{skyblue}{0.4101} & \cellcolor{skyblue}{0.4125} & \cellcolor{skyblue}-- & \cellcolor{skyblue}-- & \cellcolor{skyblue}{0.1235} & \cellcolor{skyblue}{\textbf{0.2823}} & \cellcolor{skyblue}{\textbf{0.3491}} \\
  & \cellcolor{skyblue}LLM-J & \cellcolor{skyblue}{0.6924} & \cellcolor{skyblue}{\underline{0.5667}} & \cellcolor{skyblue}{\underline{0.4877}} & \cellcolor{skyblue}{\textbf{0.6063}} & \cellcolor{skyblue}-- & \cellcolor{skyblue}-- & \cellcolor{skyblue}{0.4786} & \cellcolor{skyblue}{\textbf{0.6515}} & \cellcolor{skyblue}{\textbf{0.5820}} \\

\bottomrule
\end{tabularx}
\label{tab:QA performance qwen 7b}
\end{table*}

\begin{table*}[t!]
\small
\centering
\caption{QA performance of \benchmark on \textbf{Qwen-2.5-VL-3B}. For each metric, the best and second-best scores are marked in \textbf{bold} and with \underline{underline}, respectively.}
\vspace{-2mm}

\newcolumntype{C}{>{\scriptsize\centering\arraybackslash}X}
\newcolumntype{M}{>{\scriptsize\centering\arraybackslash}X}
\newcolumntype{W}{>{\scriptsize\centering\arraybackslash}p{1.8cm}}
\setlength{\tabcolsep}{3pt}
\renewcommand{\arraystretch}{0.6}
\setlength{\fboxsep}{1pt}

\begin{tabularx}{\linewidth}{W|M|CCCCCCCC|C}

\toprule
\multicolumn{2}{c|}{\raisebox{-0.2ex}{\includegraphics[height=1.1em]{figures/icons/qwen-icon.png}}\,\textbf{Qwen-2.5-VL-3B}} & \textbf{SS} & \textbf{MS} & \textbf{TR} & \textbf{MR} & \textbf{FM} & \textbf{FJ} & \textbf{TH} & \textbf{II} & \textbf{Avg.}\\
\midrule
\multicolumn{11}{c}{\cellcolor{Secondary}\textit{\textbf{Text-Only}}} \\
\midrule

\multirow{4}{*}{\makecell{Full Context\\(without memory)}}
  & EM     & 0.1612 & 0.1567 & 0.1118 & 0.1287 & 0.0535 & 0.6904 & -- & -- & 0.1814 \\
  & F1     & 0.2408 & 0.2812 & 0.2118 & 0.1862 & -- & -- & 0.1388 & 0.1313 & 0.2058 \\
  & BLEU-1 & 0.2185 & 0.2387 & 0.2001 & 0.1662 & -- & -- & 0.0605 & 0.0822 & 0.1748 \\
  & LLM-J  & 0.1755 & 0.1747 & 0.1958 & 0.1338 & -- & -- & 0.0347 & 0.0188 & 0.1396 \\
\midrule
\multirow{4}{*}{\makecell{NaiveRAG}}
  & EM     & 0.2667 & 0.2027 & \underline{0.3773} & 0.2494 & 0.3394 & 0.8976 & -- & -- & 0.3488 \\
  & F1     & 0.3657 & 0.3318 & 0.2183 & 0.2961 & -- & -- & 0.1205 & 0.0859 & 0.2632 \\
  & BLEU-1 & 0.3320 & 0.2252 & 0.2070 & 0.2738 & -- & -- & 0.0244 & 0.0439 & 0.2184 \\
  & LLM-J  & 0.5509 & 0.3693 & 0.4274 & 0.4342 & -- & -- & 0.2380 & 0.3331 & 0.4147 \\
\midrule
\multirow{4}{*}{\makecell{A-Mem}}
  & EM     & \textbf{0.3947} & \textbf{0.2977} & 0.2373 & 0.2133 & 0.3563 & 0.9109 & -- & -- & \underline{0.3530} \\
  & F1     & \textbf{0.5585} & \textbf{0.4221} & 0.1531 & 0.2691 & -- & -- & 0.1293 & 0.0729 & 0.2919 \\
  & BLEU-1 & \textbf{0.5012} & \underline{0.3063} & 0.1483 & 0.2345 & -- & -- & 0.0322 & 0.0227 & 0.2385 \\
  & LLM-J  & 0.6183 & \textbf{0.4282} & 0.2484 & 0.4075 & -- & -- & 0.2839 & 0.3331 & 0.3988 \\
\midrule
\multirow{4}{*}{\makecell{Mem0\\(w/o vision)}}
  & EM     & \underline{0.3867} & 0.2677 & 0.2027 & 0.2133 & 0.3150 & \textbf{0.9443} & -- & -- & 0.3386 \\
  & F1     & \underline{0.5394} & \underline{0.4147} & 0.1666 & 0.2780 & -- & -- & 0.1450 & 0.0899 & 0.2955 \\
  & BLEU-1 & \underline{0.4820} & 0.3001 & 0.1623 & 0.2406 & -- & -- & 0.0394 & 0.0407 & 0.2407 \\
  & LLM-J  & \underline{0.6215} & 0.4176 & 0.2503 & 0.3915 & -- & -- & 0.2881 & \textbf{0.3554} & 0.3959 \\
\midrule
\multirow{4}{*}{\makecell{MemoryOS}}
  & EM     & 0.3787 & \underline{0.2723} & 0.2507 & 0.1934 & 0.2783 & 0.8508 & -- & -- & 0.3262 \\
  & F1     & 0.4919 & 0.4036 & 0.1417 & 0.2492 & -- & -- & 0.1242 & 0.0606 & 0.2675 \\
  & BLEU-1 & 0.4393 & 0.2934 & 0.1427 & 0.2205 & -- & -- & 0.0249 & 0.0273 & 0.2197 \\
  & LLM-J  & 0.6094 & 0.4061 & 0.2472 & 0.3793 & -- & -- & 0.2712 & 0.3126 & 0.3827 \\

\midrule
\multicolumn{11}{c}{\cellcolor{Secondary}\textit{\textbf{Multimodality}}} \\
\midrule

\multirow{4}{*}{\makecell{Full Context\\(without memory)}}
  & EM     & 0.1294 & 0.1108 & 0.1188 & 0.1662 & 0.1453 & 0.9087 & -- & -- & 0.2168 \\
  & F1     & 0.2811 & 0.2557 & 0.2226 & 0.3003 & -- & -- & 0.1768 & 0.0916 & 0.2451 \\
  & BLEU-1 & 0.2296 & 0.1959 & 0.1969 & 0.2658 & -- & -- & 0.0967 & 0.0491 & 0.1996 \\
  & LLM-J  & 0.1960 & 0.1527 & 0.2308 & 0.2593 & -- & -- & 0.0138 & 0.0028 & 0.1787 \\
\midrule
\multirow{4}{*}{\makecell{Universal\\-RAG}}
  & EM     & 0.1122 & 0.0250 & 0.3673 & 0.2167 & 0.3807 & 0.9243 & -- & -- & 0.2966 \\
  & F1     & 0.3375 & 0.2742 & \underline{0.4855} & 0.3263 & -- & -- & 0.1319 & 0.1537 & 0.3163 \\
  & BLEU-1 & 0.2718 & 0.1506 & \underline{0.4702} & 0.2817 & -- & -- & 0.0530 & 0.0622 & 0.2543 \\
  & LLM-J  & 0.6047 & 0.3319 & 0.4940 & 0.4395 & -- & -- & 0.3088 & 0.3079 & 0.4397 \\
\midrule
\multirow{4}{*}{\makecell{RAG\\-Anything}}
  & EM     & 0.1224 & 0.0375 & \textbf{0.4107} & 0.2750 & 0.4404 & 0.8664 & -- & -- & 0.3260 \\
  & F1     & 0.3539 & 0.2596 & \textbf{0.4963} & 0.4118 & -- & -- & 0.1198 & 0.1517 & \underline{0.3418} \\
  & BLEU-1 & 0.2839 & 0.1451 & \textbf{0.4889} & 0.3601 & -- & -- & 0.0374 & 0.0757 & \underline{0.2807} \\
  & LLM-J  & 0.5876 & 0.3378 & \underline{0.5196} & 0.4877 & -- & -- & 0.2672 & 0.3362 & 0.4534 \\
\midrule
\multirow{4}{*}{\makecell{Mem0\\(with vision)}}
  & EM     & 0.2227 & 0.2227 & 0.1933 & \textbf{0.4054} & 0.2630 & 0.9131 & -- & -- & 0.3418 \\
  & F1     & 0.4191 & 0.3807 & 0.1604 & \underline{0.4781} & -- & -- & 0.1523 & 0.0887 & 0.3242 \\
  & BLEU-1 & 0.3605 & \textbf{0.3143} & 0.1506 & \underline{0.4475} & -- & -- & 0.0416 & 0.0363 & 0.2764 \\
  & LLM-J  & 0.5886 & \underline{0.4253} & 0.2482 & \underline{0.5497} & -- & -- & 0.3086 & \underline{0.3420} & 0.4365 \\
\midrule
\multirow{4}{*}{\makecell{MemVerse}}
  & EM     & 0.1437 & 0.1157 & 0.2218 & 0.2634 & \textbf{0.4541} & 0.8330 & -- & -- & 0.3021 \\
  & F1     & 0.3615 & 0.3464 & 0.3058 & 0.3698 & -- & -- & \underline{0.2439} & 0.1313 & 0.3199 \\
  & BLEU-1 & 0.3002 & 0.2506 & 0.2968 & 0.3382 & -- & -- & 0.1436 & 0.0629 & 0.2671 \\
  & LLM-J  & 0.5427 & 0.3991 & 0.3805 & 0.5083 & -- & -- & \underline{0.3513} & 0.3096 & 0.4406 \\
\midrule
\multirow{4}{*}{\makecell{NGM}}
  & EM     & 0.2048 & 0.1330 & 0.0937 & \underline{0.3584} & 0.4312 & 0.8552 & -- & -- & 0.3155 \\
  & F1     & 0.3642 & 0.3269 & 0.1618 & \textbf{0.4855} & -- & -- & 0.2028 & \textbf{0.1787} & 0.3210 \\
  & BLEU-1 & 0.3184 & 0.2311 & 0.1577 & \textbf{0.4523} & -- & -- & 0.1008 & \underline{0.0861} & 0.2695 \\
  & LLM-J  & 0.5065 & 0.3666 & 0.2329 & 0.5306 & -- & -- & 0.2950 & 0.3249 & 0.4014 \\
\midrule
\multirow{4}{*}{\makecell{MIRIX}}
  & EM     & 0.2143 & 0.0250 & 0.3316 & 0.1000 & \underline{0.4511} & 0.8463 & -- & -- & 0.2797 \\
  & F1     & 0.4138 & 0.2408 & 0.4556 & 0.2610 & -- & -- & \textbf{0.2624} & 0.0844 & 0.3109 \\
  & BLEU-1 & 0.3594 & 0.1371 & 0.4348 & 0.2169 & -- & -- & \textbf{0.2125} & 0.0391 & 0.2599 \\
  & LLM-J  & 0.5105 & 0.3081 & 0.4994 & \textbf{0.5716} & -- & -- & \textbf{0.3828} & 0.2568 & \underline{0.4618} \\
\midrule
\multirow{4}{*}{\makecell{\small\textbf{\method}\\\scriptsize(Our Method)}}
  & \cellcolor{skyblue}EM & \cellcolor{skyblue}{0.3000} & \cellcolor{skyblue}{0.1597} & \cellcolor{skyblue}{0.3503} & \cellcolor{skyblue}{0.2175} & \cellcolor{skyblue}{0.4450} & \cellcolor{skyblue}{\underline{0.9310}} & \cellcolor{skyblue}-- & \cellcolor{skyblue}-- & \cellcolor{skyblue}{\textbf{0.3546}} \\
  & \cellcolor{skyblue}F1 & \cellcolor{skyblue}{0.5218} & \cellcolor{skyblue}{0.2876} & \cellcolor{skyblue}{0.4790} & \cellcolor{skyblue}{0.4087} & \cellcolor{skyblue}-- & \cellcolor{skyblue}-- & \cellcolor{skyblue}{0.1913} & \cellcolor{skyblue}{\underline{0.1753}} & \cellcolor{skyblue}{\textbf{0.3828}} \\
  & \cellcolor{skyblue}BLEU-1 & \cellcolor{skyblue}{0.4672} & \cellcolor{skyblue}{0.1920} & \cellcolor{skyblue}{0.4262} & \cellcolor{skyblue}{0.3267} & \cellcolor{skyblue}-- & \cellcolor{skyblue}-- & \cellcolor{skyblue}{\underline{0.1776}} & \cellcolor{skyblue}{\textbf{0.1156}} & \cellcolor{skyblue}{\textbf{0.3190}} \\
  & \cellcolor{skyblue}LLM-J & \cellcolor{skyblue}{\textbf{0.6450}} & \cellcolor{skyblue}{0.3425} & \cellcolor{skyblue}{\textbf{0.5410}} & \cellcolor{skyblue}{0.4975} & \cellcolor{skyblue}-- & \cellcolor{skyblue}-- & \cellcolor{skyblue}{0.3128} & \cellcolor{skyblue}{0.3305} & \cellcolor{skyblue}{\textbf{0.4764}} \\


\bottomrule
\end{tabularx}
\label{tab:QA performance qwen 3b}
\end{table*}
\begin{table*}[t!]
\small
\centering
\caption{QA performance of \benchmark on \textbf{GPT-5-Nano}. For each metric, the best and second-best scores are marked in \textbf{bold} and with \underline{underline}, respectively.}
\vspace{-2mm}

\newcolumntype{C}{>{\scriptsize\centering\arraybackslash}X}
\newcolumntype{M}{>{\scriptsize\centering\arraybackslash}X}
\newcolumntype{W}{>{\scriptsize\centering\arraybackslash}p{1.8cm}}
\setlength{\tabcolsep}{3pt}
\renewcommand{\arraystretch}{0.6}
\setlength{\fboxsep}{1pt}

\begin{tabularx}{\linewidth}{W|M|CCCCCCCC|C}

\toprule
\multicolumn{2}{c|}{\raisebox{-0.2ex}{\includegraphics[height=1.1em]{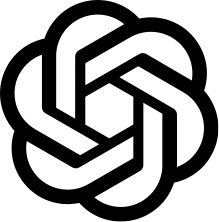}}\,\textbf{GPT-5-Nano}} & \textbf{SS} & \textbf{MS} & \textbf{TR} & \textbf{MR} & \textbf{FM} & \textbf{FJ} & \textbf{TH} & \textbf{II} & \textbf{Avg.}\\
\midrule
\multicolumn{11}{c}{\cellcolor{Secondary}\textit{\textbf{Text-Only}}} \\
\midrule

\multirow{4}{*}{\makecell{Full Context\\(without memory)}}
  & EM     & 0.1133 & 0.0973 & 0.1680 & 0.1774 & 0.0902 & 0.9644 & -- & -- & 0.2214 \\
  & F1     & 0.3896 & \underline{0.4041} & 0.2982 & 0.3083 & -- & -- & \underline{0.2636} & 0.1763 & 0.3205 \\
  & BLEU-1 & 0.3194 & 0.2929 & 0.2814 & 0.2675 & -- & -- & 0.1585 & 0.1376 & 0.2613 \\
  & LLM-J  & 0.4257 & 0.3713 & 0.2184 & 0.2432 & -- & -- & 0.3434 & 0.4037 & 0.3147 \\
\midrule
\multirow{4}{*}{\makecell{NaiveRAG}}
  & EM     & 0.2683 & 0.2042 & 0.3787 & 0.2478 & 0.5000 & 0.8953 & -- & -- & \textbf{0.3728} \\
  & F1     & 0.4082 & 0.3303 & 0.2197 & 0.2978 & -- & -- & 0.1218 & 0.0844 & 0.2716 \\
  & BLEU-1 & 0.3728 & 0.2238 & 0.2085 & 0.2723 & -- & -- & 0.0258 & 0.0426 & 0.2257 \\
  & LLM-J  & 0.5982 & 0.4012 & 0.4691 & 0.4738 & -- & -- & 0.2625 & 0.3613 & 0.4524 \\
\midrule
\multirow{4}{*}{\makecell{A-Mem}}
  & EM     & \textbf{0.3938} & \textbf{0.2965} & 0.2392 & 0.2152 & 0.4786 & 0.9109 & -- & -- & \underline{0.3718} \\
  & F1     & 0.4213 & 0.3205 & 0.1543 & 0.2705 & -- & -- & 0.1278 & 0.0742 & 0.2520 \\
  & BLEU-1 & 0.4567 & 0.2082 & 0.1495 & 0.2358 & -- & -- & 0.0307 & 0.0241 & 0.2163 \\
  & LLM-J  & 0.6695 & 0.4673 & 0.2715 & 0.4422 & -- & -- & 0.3078 & 0.3565 & 0.4329 \\
\midrule
\multirow{4}{*}{\makecell{Mem0\\(w/o vision)}}
  & EM     & \underline{0.3852} & 0.2692 & 0.2042 & 0.2147 & 0.4037 & 0.9421 & -- & -- & 0.3522 \\
  & F1     & \textbf{0.6038} & \textbf{0.4163} & 0.1681 & 0.2766 & -- & -- & 0.1432 & 0.0883 & 0.3072 \\
  & BLEU-1 & \textbf{0.5373} & \textbf{0.2986} & 0.1638 & 0.2391 & -- & -- & 0.0408 & 0.0421 & 0.2508 \\
  & LLM-J  & 0.6732 & 0.4533 & 0.2762 & 0.4255 & -- & -- & 0.3128 & 0.3838 & 0.4303 \\
\midrule
\multirow{4}{*}{\makecell{MemoryOS}}
  & EM     & 0.3771 & \underline{0.2740} & 0.2492 & 0.1952 & 0.2798 & 0.8530 & -- & -- & 0.3268 \\
  & F1     & \underline{0.5473} & 0.4019 & 0.1429 & 0.2477 & -- & -- & 0.1226 & 0.0623 & 0.2773 \\
  & BLEU-1 & \underline{0.4884} & \underline{0.2947} & 0.1442 & 0.2192 & -- & -- & 0.0233 & 0.0258 & 0.2285 \\
  & LLM-J  & 0.6602 & 0.4751 & 0.2632 & 0.4076 & -- & -- & 0.2945 & 0.3411 & 0.4183 \\

\midrule
\multicolumn{11}{c}{\cellcolor{Secondary}\textit{\textbf{Multimodality}}} \\
\midrule

\multirow{4}{*}{\makecell{Full Context\\(without memory)}}
  & EM     & 0.0838 & 0.0973 & 0.1670 & 0.2200 & 0.1514 & 0.8597 & -- & -- & 0.2257 \\
  & F1     & 0.3123 & 0.3440 & 0.3328 & 0.3438 & -- & -- & 0.2259 & \underline{0.2503} & 0.3153 \\
  & BLEU-1 & 0.2553 & 0.2580 & 0.3147 & 0.3037 & -- & -- & 0.1388 & \underline{0.2110} & 0.2642 \\
  & LLM-J  & 0.3707 & 0.3146 & 0.2187 & 0.3148 & -- & -- & 0.3084 & 0.3752 & 0.3102 \\
\midrule
\multirow{4}{*}{\makecell{Universal\\-RAG}}
  & EM     & 0.1633 & 0.0750 & 0.3776 & 0.2333 & 0.2798 & 0.9042 & -- & -- & 0.3013 \\
  & F1     & 0.4202 & 0.3465 & \underline{0.4112} & 0.3989 & -- & -- & 0.1554 & \textbf{0.2504} & 0.3583 \\
  & BLEU-1 & 0.3595 & 0.1875 & \textbf{0.4612} & 0.3531 & -- & -- & 0.0330 & 0.1936 & \underline{0.3020} \\
  & LLM-J  & \underline{0.6939} & \textbf{0.6375} & 0.4898 & 0.5750 & -- & -- & 0.4250 & \textbf{0.8250} & \underline{0.5913} \\
\midrule
\multirow{4}{*}{\makecell{RAG\\-Anything}}
  & EM     & 0.1122 & 0.0625 & 0.3673 & 0.2056 & 0.3593 & \textbf{0.9844} & -- & -- & 0.3018 \\
  & F1     & 0.4126 & 0.3204 & 0.3299 & 0.3800 & -- & -- & 0.1407 & 0.2035 & 0.3267 \\
  & BLEU-1 & 0.3406 & 0.1875 & 0.3724 & 0.3248 & -- & -- & 0.0152 & 0.1504 & 0.2679 \\
  & LLM-J  & 0.6721 & 0.5290 & 0.4809 & 0.5343 & -- & -- & 0.3986 & 0.5980 & 0.5378 \\
\midrule
\multirow{4}{*}{\makecell{Mem0\\(with vision)}}
  & EM     & 0.2208 & 0.2243 & 0.1948 & \textbf{0.4072} & 0.0612 & 0.9265 & -- & -- & 0.3137 \\
  & F1     & 0.4178 & 0.3793 & 0.1622 & \underline{0.4767} & -- & -- & 0.1508 & 0.0875 & 0.3235 \\
  & BLEU-1 & 0.3623 & 0.2858 & 0.1493 & \underline{0.4462} & -- & -- & 0.0432 & 0.0348 & 0.2719 \\
  & LLM-J  & 0.6212 & 0.4488 & 0.2562 & 0.5738 & -- & -- & 0.3258 & 0.3637 & 0.4580 \\
\midrule
\multirow{4}{*}{\makecell{MemVerse}}
  & EM     & 0.0816 & 0.0250 & \textbf{0.4617} & 0.2333 & 0.4801 & 0.9265 & -- & -- & 0.3277 \\
  & F1     & 0.2602 & 0.2541 & 0.3875 & 0.4307 & -- & -- & 0.0995 & 0.2340 & 0.3118 \\
  & BLEU-1 & 0.2185 & 0.1459 & \underline{0.3997} & 0.3707 & -- & -- & 0.1151 & 0.1740 & 0.2707 \\
  & LLM-J  & 0.6224 & 0.5312 & \underline{0.5255} & \underline{0.6167} & -- & -- & 0.3917 & \underline{0.7375} & 0.5706 \\
\midrule
\multirow{4}{*}{\makecell{NGM}}
  & EM     & 0.1020 & 0.1225 & 0.1224 & \underline{0.3667} & 0.3394 & 0.9443 & -- & -- & 0.2992 \\
  & F1     & 0.2824 & 0.2896 & 0.1959 & 0.4283 & -- & -- & 0.0666 & 0.2158 & 0.2781 \\
  & BLEU-1 & 0.2354 & 0.1651 & 0.1880 & 0.3830 & -- & -- & 0.0152 & 0.1559 & 0.2262 \\
  & LLM-J  & 0.4898 & 0.5188 & 0.1633 & 0.5889 & -- & -- & 0.1917 & 0.7000 & 0.4402 \\
\midrule
\multirow{4}{*}{\makecell{MIRIX}}
  & EM     & 0.1224 & 0.1043 & \underline{0.4031} & 0.3281 & \underline{0.5245} & 0.9666 & -- & -- & 0.3705 \\
  & F1     & 0.4374 & 0.2888 & 0.4032 & \textbf{0.4993} & -- & -- & \textbf{0.2929} & 0.2305 & \textbf{0.3940} \\
  & BLEU-1 & 0.3671 & 0.1621 & 0.3665 & \textbf{0.4481} & -- & -- & \textbf{0.2283} & 0.1576 & \textbf{0.3277} \\
  & LLM-J  & 0.6342 & 0.5208 & 0.4004 & \textbf{0.6497} & -- & -- & \underline{0.5119} & 0.6385 & 0.5626 \\
\midrule
\multirow{4}{*}{\makecell{\small\textbf{\method}\\\scriptsize(Our Method)}}
  & \cellcolor{skyblue}EM & \cellcolor{skyblue}{0.1800} & \cellcolor{skyblue}{0.2307} & \cellcolor{skyblue}{0.2307} & \cellcolor{skyblue}{0.3394} & \cellcolor{skyblue}{\textbf{0.5306}} & \cellcolor{skyblue}{\underline{0.9777}} & \cellcolor{skyblue}-- & \cellcolor{skyblue}-- & \cellcolor{skyblue}{0.3717} \\
  & \cellcolor{skyblue}F1 & \cellcolor{skyblue}{0.5292} & \cellcolor{skyblue}{0.3490} & \cellcolor{skyblue}{\textbf{0.4147}} & \cellcolor{skyblue}{0.3174} & \cellcolor{skyblue}-- & \cellcolor{skyblue}-- & \cellcolor{skyblue}{0.2155} & \cellcolor{skyblue}{0.2270} & \cellcolor{skyblue}{\underline{0.3617}} \\
  & \cellcolor{skyblue}BLEU-1 & \cellcolor{skyblue}{0.4537} & \cellcolor{skyblue}{0.2841} & \cellcolor{skyblue}{0.3710} & \cellcolor{skyblue}{0.2279} & \cellcolor{skyblue}-- & \cellcolor{skyblue}-- & \cellcolor{skyblue}{\underline{0.1623}} & \cellcolor{skyblue}{\textbf{0.2516}} & \cellcolor{skyblue}{0.3001} \\
  & \cellcolor{skyblue}LLM-J & \cellcolor{skyblue}{\textbf{0.7317}} & \cellcolor{skyblue}{\underline{0.5500}} & \cellcolor{skyblue}{\textbf{0.5350}} & \cellcolor{skyblue}{0.5975} & \cellcolor{skyblue}-- & \cellcolor{skyblue}-- & \cellcolor{skyblue}{\textbf{0.5500}} & \cellcolor{skyblue}{0.5875} & \cellcolor{skyblue}{\textbf{0.5969}} \\

\bottomrule
\end{tabularx}
\label{tab:QA performance gpt}
\end{table*}
\begin{table*}[t!]
\small
\centering
\caption{QA performance of \benchmark on \textbf{Doubao-Seed-2.0-Pro}. For each metric, the best and second-best scores are marked in \textbf{bold} and with \underline{underline}, respectively.}
\vspace{-2mm}

\newcolumntype{C}{>{\scriptsize\centering\arraybackslash}X}
\newcolumntype{M}{>{\scriptsize\centering\arraybackslash}X}
\newcolumntype{W}{>{\scriptsize\centering\arraybackslash}p{1.8cm}}
\setlength{\tabcolsep}{3pt}
\renewcommand{\arraystretch}{0.6}
\setlength{\fboxsep}{1pt}

\begin{tabularx}{\linewidth}{W|M|CCCCCCCC|C}

\toprule
\multicolumn{2}{c|}{\raisebox{-0.2ex}{\includegraphics[height=1.1em]{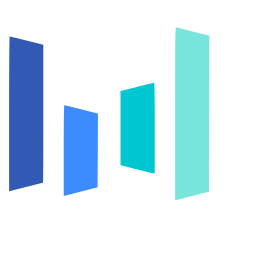}}\,\textbf{Doubao-Seed-2.0-Pro}} & \textbf{SS} & \textbf{MS} & \textbf{TR} & \textbf{MR} & \textbf{FM} & \textbf{FJ} & \textbf{TH} & \textbf{II} & \textbf{Avg.}\\
\midrule
\multicolumn{11}{c}{\cellcolor{Secondary}\textit{\textbf{Text-Only}}} \\
\midrule

\multirow{4}{*}{\makecell{Full Context\\(without memory)}}
  & EM     & 0.1211 & 0.1400 & 0.3925 & 0.2040 & 0.0979 & 0.9137 & -- & -- & 0.2718 \\
  & F1     & \underline{0.4871} & \underline{0.4459} & \textbf{0.5253} & 0.3094 & -- & -- & \underline{0.3313} & \underline{0.3289} & \underline{0.4085} \\
  & BLEU-1 & 0.3961 & 0.3373 & \underline{0.5028} & 0.2790 & -- & -- & \underline{0.2123} & \underline{0.2824} & 0.3454 \\
  & LLM-J  & 0.1977 & 0.1912 & 0.4929 & 0.2560 & -- & -- & 0.0391 & 0.0593 & 0.2420 \\
\midrule
\multirow{4}{*}{\makecell{NaiveRAG}}
  & EM     & \underline{0.2626} & 0.2008 & 0.3777 & 0.2463 & 0.4954 & 0.9020 & -- & -- & 0.3708 \\
  & F1     & 0.4069 & 0.3325 & 0.2181 & 0.2956 & -- & -- & 0.1168 & 0.0842 & 0.2701 \\
  & BLEU-1 & 0.3695 & 0.2257 & 0.2059 & 0.2714 & -- & -- & 0.0235 & 0.0415 & 0.2243 \\
  & LLM-J  & 0.5940 & 0.4014 & 0.4621 & 0.4753 & -- & -- & 0.2606 & 0.3645 & 0.4507 \\
\midrule
\multirow{4}{*}{\makecell{A-Mem}}
  & EM     & 0.2488 & \underline{0.2772} & 0.1187 & 0.1157 & 0.2018 & 0.8953 & -- & -- & 0.2539 \\
  & F1     & 0.3363 & 0.3837 & 0.0830 & 0.1559 & -- & -- & 0.0980 & 0.1273 & 0.2001 \\
  & BLEU-1 & 0.3108 & 0.3087 & 0.0756 & 0.1411 & -- & -- & 0.0284 & 0.0916 & 0.1680 \\
  & LLM-J  & 0.4448 & 0.4837 & 0.1250 & 0.2636 & -- & -- & 0.1694 & 0.4890 & 0.3090 \\
\midrule
\multirow{4}{*}{\makecell{Mem0\\(w/o vision)}}
  & EM     & \underline{0.2626} & 0.2676 & 0.1529 & 0.1847 & 0.5734 & 0.9488 & -- & -- & 0.3400 \\
  & F1     & 0.4829 & 0.4368 & 0.1430 & 0.2671 & -- & -- & 0.1788 & 0.1538 & 0.2894 \\
  & BLEU-1 & \underline{0.4177} & \underline{0.3422} & 0.1302 & 0.2303 & -- & -- & 0.0438 & 0.1268 & 0.2329 \\
  & LLM-J  & 0.6503 & \textbf{0.5839} & 0.2339 & 0.4337 & -- & -- & 0.3491 & 0.5048 & 0.4528 \\
\midrule
\multirow{4}{*}{\makecell{MemoryOS}}
  & EM     & 0.2056 & \textbf{0.2876} & 0.1901 & 0.1713 & 0.3135 & 0.8953 & -- & -- & 0.2919 \\
  & F1     & 0.4422 & 0.4142 & 0.1354 & 0.2723 & -- & -- & 0.1439 & 0.1644 & 0.2753 \\
  & BLEU-1 & 0.3761 & 0.3076 & 0.1250 & 0.2331 & -- & -- & 0.0244 & 0.1305 & 0.2179 \\
  & LLM-J  & 0.6326 & \underline{0.5537} & 0.1848 & 0.4358 & -- & -- & 0.3025 & 0.5158 & 0.4316 \\

\midrule
\multicolumn{11}{c}{\cellcolor{Secondary}\textit{\textbf{Multimodality}}} \\
\midrule

\multirow{4}{*}{\makecell{Full Context\\(without memory)}}
  & EM     & 0.1220 & 0.1311 & 0.2384 & \textbf{0.4994} & 0.1575 & \textbf{0.9777} & -- & -- & 0.3355 \\
  & F1     & 0.4869 & \textbf{0.4700} & 0.4491 & \textbf{0.6334} & -- & -- & \textbf{0.3384} & \textbf{0.3311} & \textbf{0.4896} \\
  & BLEU-1 & 0.4051 & \textbf{0.3628} & 0.4039 & \textbf{0.5876} & -- & -- & \textbf{0.2127} & 0.2762 & \textbf{0.4183} \\
  & LLM-J  & 0.2513 & 0.2181 & 0.4498 & 0.5561 & -- & -- & 0.0390 & 0.0893 & 0.3344 \\
\midrule
\multirow{4}{*}{\makecell{Universal\\-RAG}}
  & EM     & 0.1020 & 0.0500 & \textbf{0.4923} & 0.2444 & 0.4373 & 0.9510 & -- & -- & 0.3392 \\
  & F1     & 0.3816 & 0.3400 & \underline{0.5091} & 0.3921 & -- & -- & 0.1493 & 0.2502 & 0.3666 \\
  & BLEU-1 & 0.3082 & 0.2179 & \textbf{0.5057} & 0.3452 & -- & -- & 0.0260 & 0.2075 & 0.3037 \\
  & LLM-J  & 0.6362 & 0.5511 & 0.5015 & 0.5732 & -- & -- & 0.2978 & 0.5958 & 0.5377 \\
\midrule
\multirow{4}{*}{\makecell{RAG\\-Anything}}
  & EM     & 0.0816 & 0.0500 & 0.4337 & 0.2444 & 0.4404 & \underline{0.9666} & -- & -- & 0.3272 \\
  & F1     & 0.3605 & 0.2947 & 0.4738 & 0.3729 & -- & -- & 0.1618 & 0.2862 & 0.3479 \\
  & BLEU-1 & 0.2855 & 0.2007 & 0.4562 & 0.3304 & -- & -- & 0.0355 & 0.2532 & 0.2877 \\
  & LLM-J  & 0.6047 & 0.4874 & 0.5061 & 0.5395 & -- & -- & 0.3090 & \underline{0.6054} & 0.5159 \\
\midrule
\multirow{4}{*}{\makecell{Mem0\\(with vision)}}
  & EM     & 0.2186 & 0.2185 & 0.1964 & 0.4012 & 0.3609 & 0.9287 & -- & -- & 0.3562 \\
  & F1     & 0.4194 & 0.3833 & 0.1553 & 0.4780 & -- & -- & 0.1473 & 0.1868 & 0.3305 \\
  & BLEU-1 & 0.3596 & 0.2894 & 0.1454 & 0.4441 & -- & -- & 0.0420 & 0.1339 & 0.2779 \\
  & LLM-J  & 0.6198 & 0.4522 & 0.2521 & 0.5740 & -- & -- & 0.3228 & 0.3652 & 0.4572 \\
\midrule
\multirow{4}{*}{\makecell{MemVerse}}
  & EM     & 0.0816 & 0.0375 & \underline{0.4439} & 0.4000 & \underline{0.5994} & 0.9577 & -- & -- & \underline{0.3907} \\
  & F1     & 0.3062 & 0.2543 & 0.3946 & 0.4444 & -- & -- & 0.1544 & 0.2700 & 0.3346 \\
  & BLEU-1 & 0.2438 & 0.1811 & 0.3855 & 0.4017 & -- & -- & 0.0354 & 0.2327 & 0.2819 \\
  & LLM-J  & 0.5979 & 0.4902 & 0.4366 & 0.6587 & -- & -- & 0.3318 & 0.5865 & 0.5363 \\
\midrule
\multirow{4}{*}{\makecell{NGM}}
  & EM     & 0.0204 & 0.0375 & 0.1531 & \underline{0.4222} & 0.5612 & 0.9131 & -- & -- & 0.3235 \\
  & F1     & 0.1865 & 0.2596 & 0.1520 & \underline{0.5887} & -- & -- & 0.0941 & 0.2864 & 0.3011 \\
  & BLEU-1 & 0.1457 & 0.1553 & 0.1469 & \underline{0.5431} & -- & -- & 0.0112 & 0.2363 & 0.2509 \\
  & LLM-J  & 0.3827 & 0.4938 & 0.1633 & \textbf{0.7472} & -- & -- & 0.1250 & \textbf{0.6500} & 0.4500 \\
\midrule
\multirow{4}{*}{\makecell{MIRIX}}
  & EM     & \textbf{0.3265} & 0.0250 & 0.2449 & 0.1778 & 0.3777 & 0.9532 & -- & -- & 0.3036 \\
  & F1     & \textbf{0.5475} & 0.2780 & 0.3562 & 0.3427 & -- & -- & 0.2551 & 0.1794 & 0.3512 \\
  & BLEU-1 & \textbf{0.4921} & 0.1707 & 0.3344 & 0.2927 & -- & -- & 0.2031 & 0.1475 & 0.2983 \\
  & LLM-J  & \underline{0.6735} & 0.3750 & \textbf{0.5255} & \underline{0.6861} & -- & -- & \textbf{0.4250} & 0.3125 & \underline{0.5482} \\
\midrule
\multirow{4}{*}{\makecell{\small\textbf{\method}\\\scriptsize(Our Method)}}
  & \cellcolor{skyblue}EM & \cellcolor{skyblue}{0.1838} & \cellcolor{skyblue}{0.1893} & \cellcolor{skyblue}{0.3860} & \cellcolor{skyblue}{0.3973} & \cellcolor{skyblue}{\textbf{0.6743}} & \cellcolor{skyblue}{\underline{0.9666}} & \cellcolor{skyblue}-- & \cellcolor{skyblue}-- & \cellcolor{skyblue}{\textbf{0.4300}} \\
  & \cellcolor{skyblue}F1 & \cellcolor{skyblue}{0.4436} & \cellcolor{skyblue}{0.2171} & \cellcolor{skyblue}{0.4525} & \cellcolor{skyblue}{0.4657} & \cellcolor{skyblue}-- & \cellcolor{skyblue}-- & \cellcolor{skyblue}{0.2272} & \cellcolor{skyblue}{0.3039} & \cellcolor{skyblue}{0.3826} \\
  & \cellcolor{skyblue}BLEU-1 & \cellcolor{skyblue}{0.3698} & \cellcolor{skyblue}{0.2574} & \cellcolor{skyblue}{0.4353} & \cellcolor{skyblue}{0.3970} & \cellcolor{skyblue}-- & \cellcolor{skyblue}-- & \cellcolor{skyblue}{0.1940} & \cellcolor{skyblue}{\textbf{0.2927}} & \cellcolor{skyblue}{\underline{0.3476}} \\
  & \cellcolor{skyblue}LLM-J & \cellcolor{skyblue}{\textbf{0.7850}} & \cellcolor{skyblue}{0.4562} & \cellcolor{skyblue}{\underline{0.5100}} & \cellcolor{skyblue}{0.5775} & \cellcolor{skyblue}-- & \cellcolor{skyblue}-- & \cellcolor{skyblue}{\underline{0.3583}} & \cellcolor{skyblue}{0.4750} & \cellcolor{skyblue}{\textbf{0.5519}} \\


\bottomrule
\end{tabularx}
\label{tab:QA performance doubao}
\end{table*}

Each persona scenario in \benchmark{} is seeded from a \emph{persona}
(who the user is and what their interactions cover) and a
\emph{core event} (the storyline that the multi-session timeline
elaborates). Table~\ref{tab:persona-core} summarises both seeds for all
fifteen subsets; these are the inputs to the timeline-construction
prompts described in Section~\ref{sec:appendix-prompt-templates}.

\begingroup
\clearpage
\onecolumn
\setlength{\LTcapwidth}{\textwidth}
\scriptsize
\setlength{\tabcolsep}{4pt}
\renewcommand{\arraystretch}{1.05}
\begin{longtable}{@{}>{\centering\arraybackslash}m{2.3cm}|>{\raggedright\arraybackslash}m{6.7cm}|>{\raggedright\arraybackslash}m{6.7cm}@{}}
\caption{Persona and core-event seeds for the fifteen subsets of
\benchmark{}. The persona defines the user's identity and the scope of
their interactions; the core event defines the storyline that the
multi-session timeline elaborates.}
\label{tab:persona-core}\\
\toprule
\textbf{Subset} & \textbf{Persona} & \textbf{Core Event} \\
\midrule
\endfirsthead
\multicolumn{3}{l}{\footnotesize\itshape Table~\ref{tab:persona-core} (continued)}\\
\toprule
\textbf{Subset} & \textbf{Persona} & \textbf{Core Event} \\
\midrule
\endhead
\midrule
\multicolumn{3}{r}{\footnotesize\itshape continued on next page}\\
\endfoot
\bottomrule
\endlastfoot

\textbf{Alex} (Veterinarian)
& Head clinician of a pet hospital handling a broad caseload of dogs, cats, and exotics; takes his work seriously, documents cases photographically, teaches techniques to his techs, and is a caring cat owner and a big snacks fan.
& A full spectrum of cases---a parvovirus puppy, a blocked cat, an oxygen-dependent budgie, a dehydrated bearded dragon, a car-struck kitten, and surgical foreign-body and laceration management---interleaved with hard owner conversations about cost and end-of-life decisions while teaching and caring for his techs. \\
\midrule
\textbf{Carl} (Pilot)
& Chicago-based commercial airline pilot in his career-progression phase flying the ORD--LAX route; doting father, deep-dish-pizza regular, and casual flight-prep-plus-food-and-music conversationalist.
& A single early-October trip: flying an ORD--LAX leg, completing a multi-day captain-upgrade training course in Los Angeles, earning the training-completion certificate, and returning to Chicago to celebrate. \\
\midrule
\textbf{Claire} (Cat Cafe)
& Runs the Cat Loft, a small cat-cafe and kitten-rescue space; warm, practical, and patient, handling kitten care, pregnant-stray intake, vet visits, adoption matchmaking, and social media, baking cat-friendly catnip treats for adoptive families.
& Scaling up the rescue: a community TNR effort, raising and placing several kitten litters (including one from a pregnant stray she helps deliver), weathering a sick-kitten scare, and growing the cafe's social-media reach, all with a hands-on caretaking approach. \\
\midrule
\textbf{Ethan} (Violinist)
& Career violinist and visiting orchestra coach at a Seattle arts high school; gentle, patient, and documentation-minded, deliberate about his energy, a devoted Seattle ramen hunter who hosts Studio Ghibli nights with students and dotes on his niece Lily.
& A visiting-coach residency: preparing rehearsals, mediating a student conflict, and troubleshooting a stage-lighting failure right before the concert, closing with a framed certificate at the post-concert dinner, all in his gentle, student-first style. \\
\midrule
\textbf{Liam} (Woodworking Apprentice)
& Self-taught woodworking apprentice climbing a ladder of projects from chisel sharpening and a small box up to a dining table and wardrobe; humble, curious, and detail-oriented, asking practical ``any tips?'' questions and studying American hardwoods on the side.
& Climbing a deliberate apprenticeship ladder from a first box and bird carving through bookshelf and birdhouse builds, graduating to a solid-wood dining table and a full wardrobe, with woodworking-class study material complementing the bench work. \\
\midrule
\textbf{Lily} (Cooking Newbie)
& Enthusiastic first-time cook buying her first knife, board, and wok and working up from basic stir-fries to omelets, sauces, pasta, and a quiche; curious and process-oriented, sharing photos and asking ``did I do it right?'', loving the small wins of home cooking.
& Moving from a first market trip and basic stir-fry through knife skills, sauces, omelets, smoothie bowls, pasta, and a quiche, building a starter repertoire and the confidence to take on a community-cooking class studying nutrition and recipe guides. \\
\midrule
\textbf{Lin Zeyu} (CS Junior)
& Computer Science junior juggling math and systems coursework with backend and full-stack project work (SQL, Spring Boot, Vue, Docker, Git, Redis); technical, direct, and tenacious, levelling up toward an industry SDE role on late-night coding sessions and cheap fried rice.
& A full semester from foundational coursework into a capstone: slow SQL queries, Spring Boot startup failures, Vue bugs, CORS issues, and full backend--frontend integration, culminating in a deployed remote-server stack plus CS-paper self-study. \\
\midrule
\textbf{Lin} (Bike Repair)
& Runs a small, busy bike-repair shop in a tourist town; hands-on wear diagnostics, brake and bottom-bracket service, e-bike checks, vintage restoration, and trainee mentoring; practical and relationship-first, enjoys weekend group rides, and keeps studying Shimano manuals.
& A packed month at the start of tourist season: managing a repair surge while professionalising the shop with a safety-disclosure script, trainee training, a beginner workshop, restocking, and a vintage-bike restoration, keeping pricing honest and customers front and centre. \\
\midrule
\textbf{Lina} (Family)
& Busy mother of two (Emma on piano and presentation nerves, Brandon on calculus and track) managing her own hyperthyroidism and running the household; warm, careful, a touch anxious, putting family wellness first and always trying to cook a little better.
& A normal-life family stretch: supporting both kids through their worries, handling a hyperthyroidism clinic visit, troubleshooting a washing-machine overflow, smoothing over parent-meeting tensions, and reining in grocery costs without anyone feeling squeezed. \\
\midrule
\textbf{Marcus} (Travel)
& Unhurried solo traveller on a multi-country trip (Japan, Vietnam, Thailand, France, Spain); experiences-over-things style and a keen amateur photographer writing for a personal travel blog, hunting unusual local snacks and crafts for his niece Lily.
& A deliberate Asia-to-Europe arc: Osaka ryokans and temples, Hoi An street food, Bangkok and Chiang Mai markets, Montmartre wine-and-cheese, and Barcelona's Gaud\'i architecture, photographing each stop for his blog and gathering snacks and crafts for Lily. \\
\midrule
\textbf{Mathieu} (Reservoir Engineer)
& Reservoir engineer at TotalEnergies in Pau, France; well-log interpretation, 3D Petrel geomodels, Eclipse simulation, history matching, and waterflood optimisation, with a growing CCS and energy-transition interest; surfs the Atlantic coast on weekends.
& His structured first months at TotalEnergies: joining the asset team, learning integrated log-to-model interpretation, building first geomodels and simulations, and presenting field-development recommendations, later digging into the company's financial report and energy-transition strategy. \\
\midrule
\textbf{Mina} (Botany Student)
& Undergraduate botany student building a daily plant-identification practice (leaf arrangement, venation, stress diagnosis, herbarium, invasive-species project); careful and evidence-led, leaning on iNaturalist but cross-checking with textbooks and TAs, and loving weekend nature walks.
& A semester building a botany toolkit through campus identification, venation and trichome documentation, an invasive-species visual guide, and garden visits, then a deeper-reading phase on plant phylogenomics, marking her shift toward graduate-level research. \\
\midrule
\textbf{Noah} (Barista Apprentice)
& Barista apprentice at a specialty coffee shop methodically working through craft fundamentals (pour-over, espresso dial-in, grinder settings, microfoam, cold brew, TDS); technical and detail-led, geeking out over single-origin beans and supplementing with barista training manuals.
& A structured apprenticeship arc from pour-over and espresso basics through cold brew, microfoam, single-origin dial-in, water quality, and batch brewing, culminating in a signature-blend capstone, followed by deeper study of specialty-coffee training material. \\
\midrule
\textbf{Oliver} (PhD CS)
& CS PhD student working on Transformer architectures, RAG systems, and vector retrieval (FAISS, Milvus); mixes deep technical debugging with academic rhythms of ablations, drafts, and LaTeX revisions; a devoted campus-coffee-shop regular powering late-night writing pushes.
& A complete PhD research arc: Transformer reading, HuggingFace and training-pipeline debugging, building and tuning RAG and vector-retrieval systems on Milvus, running an ablation, and revising a paper draft, wrapping with foundational-paper study, often from his favourite coffee shop. \\
\midrule
\textbf{Wei} (Courier)
& Food-delivery courier in a tier-2 Chinese city working surge-driven app-dispatched shifts; route optimisation across modes, pickup pacing, and turnstile navigation; hardworking, resourceful, and budget-conscious, getting by on cheap eats and a rented single room, and studying labour-conditions papers.
& An intensive shift cluster running full courier rotations through rainy-brake scares, e-bike-vs-motorcycle cost decisions, office-turnstile gauntlets, marathon detours, and late-night surges, while managing rent, meals, and rest, later studying ergonomic and platform-governance papers. \\

\end{longtable}
\twocolumn
\endgroup

\section{Prompt Templates}
\label{sec:appendix-prompt-templates}

We list the prompt templates in the order they are invoked while
constructing the dataset. 

\paragraph{Timeline construction.}
Generation begins from the persona and core event. We first prompt the
LLM with the \emph{Timeline Generation} template
(Figure~\ref{fig:prompt-timeline-gen}) to produce the core-event line
of $N$ chronological events, each carrying a description, a
visual-grounded user query, an absolute date, and an image-search
keyword. We then call the \emph{Distractor Timeline Generation}
template (Figure~\ref{fig:prompt-distractor}) to add $M$ off-topic side
events drawn from orthogonal domains, which are interleaved into the
core line by date. The merged timeline is passed through one round of
the \emph{Timeline Self-Check} template
(Figure~\ref{fig:prompt-timeline-check}), which audits persona
violations, core-event drift, and internal contradictions and flags
any offending events for repair. The audited timeline is then released
to a human reviewer for a final spot-check before it enters dialogue
synthesis.

\begin{figure*}[t]
\begin{tcolorbox}[colback=gray!10,colframe=black,width=\linewidth,title=Timeline Generation]
You are generating a timeline of events for a synthetic user's life. Return ONLY a JSON array. No markdown. No trailing commentary.

\textbf{Persona:} <persona> \\
\textbf{Core event:} <core\_event> \\

\textbf{Hard requirements:}
\begin{itemize}\setlength\itemsep{0pt}
\item Output EXACTLY <N> event objects (no more, no fewer).
\item Every string field below must be NON-EMPTY. Never use \texttt{""} or omit a key.
\item Chronological \texttt{Event\_Time} across the story.
\end{itemize}
Each event MUST be an object with these keys in order:
(1)~\texttt{Event\_Index} (integer 1..N matching array position);
(2)~\texttt{Event\_Description} (one third-person sentence);
(3)~\texttt{Query\_Description} (one sentence using the protagonist name: ``[Name] shares photos/screenshots of [concrete visuals] and asks [one specific question]'');
(4)~\texttt{Event\_Time} (\texttt{YYYY-MM-DD});
(5)~\texttt{Keyword} (a 2--6 word image-search phrase).
\end{tcolorbox}
\caption{Prompt template for timeline generation. The event count \texttt{<N>} is set by the configuration.}
\label{fig:prompt-timeline-gen}
\end{figure*}

\begin{figure*}[t]
\begin{tcolorbox}[colback=gray!10,colframe=black,width=\linewidth,title=Distractor Timeline Generation]
You are generating DISTRACTOR life events for the same synthetic user, to be interleaved with their main story. These events must look like ordinary moments in this person's life but must NOT advance, mirror, or reference the core event arc. Return ONLY a JSON array.

\textbf{Persona:} <persona> \\
\textbf{Core event (DO NOT touch these themes):} <core\_event> \\
\textbf{Existing main-line events (date range only):} <mainline\_context> \\

\textbf{Hard requirements:}
\begin{itemize}\setlength\itemsep{0pt}
\item Output EXACTLY <M> event objects.
\item Each event is a believable, mundane moment ORTHOGONAL to the core event (food / errands / hobbies / minor health / weather / family chat / commute / app glitch).
\item \texttt{Event\_Time} values fall WITHIN the main-line date range and interleave plausibly (do not bunch on one date).
\end{itemize}
Keys per object: \texttt{Event\_Description}, \texttt{Query\_Description} (off-topic visual + unrelated question), \texttt{Event\_Time}, \texttt{Keyword} (off-topic).
\end{tcolorbox}
\caption{Prompt template for distractor (side-event) generation.}
\label{fig:prompt-distractor}
\end{figure*}

\begin{figure*}[t]
\begin{tcolorbox}[colback=gray!10,colframe=black,width=\linewidth,title=Timeline Self-Check]
You are auditing a generated timeline against the persona and the core event. Return ONLY a JSON object.

\textbf{Persona:} <persona> \quad \textbf{Core event:} <core\_event> \\
\textbf{Timeline (JSON array of events):} <timeline> \\

Check three things and report each finding with the offending \texttt{Event\_Index} values:
\begin{itemize}\setlength\itemsep{0pt}
\item \texttt{persona\_violations} --- events contradicting the persona (wrong name, profession, impossible behaviour).
\item \texttt{core\_event\_violations} --- main-line events that fail to express or contradict the core-event arc (events marked off-topic are exempt).
\item \texttt{contradictions} --- pairs of events that contradict each other (impossible date order, mutually exclusive states, name drift).
\end{itemize}
Output a JSON object with the three arrays above plus \texttt{"ok"} (true iff all are empty).
\end{tcolorbox}
\caption{Prompt template for the timeline self-check audit.}
\label{fig:prompt-timeline-check}
\end{figure*}

\paragraph{Dialogue synthesis.}
With the audited timeline in hand, we generate the conversation by
walking an event sliding window: each round feeds the current event
(and, after the first round, the immediately preceding one) into the
\emph{Thematic User Simulator} template
(Figure~\ref{fig:prompt-thematic-user}), which produces the user's
motivated query over the event's attached images. The simulator is
conditioned on the running conversation summaries so that follow-up
queries can revisit earlier context. On the \emph{first} round the
window contains only the opening event and there is no prior summary,
so the simulator emits a single scene-setting question that introduces
the persona's situation rather than a follow-up. When the shared
evidence for an event is a \emph{document} rather than a photo, the
same simulator role is used with two differences: the user is shown a
batch of freshly rendered PDF pages instead of images and must ask a
question that quotes a concrete number, label, or term from those
pages; and the assistant answers using only the visible pages,
saying so when the answer is not present. The user query is then
passed to the \emph{Thematic Assistant} template
(Figure~\ref{fig:prompt-thematic-agent}), which replies over the
same attached evidence. After each round, the exchange is condensed by
the \emph{Round Summary} template (Figure~\ref{fig:prompt-summary}) and
appended to the running memory that conditions the next round.

\begin{figure*}[t]
\begin{tcolorbox}[colback=gray!10,colframe=black,width=\linewidth,title=Thematic User Simulator]
You are the USER in a roleplay.

\textbf{Persona:} <persona> \\
\textbf{Relevant event(s) (JSON):} <events> \\
\textbf{Conversation summaries so far (JSON):} <summaries> \\
\textbf{Attached images:} <image\_desc> \\

Write TWO consecutive user messages that progress the conversation. Return ONLY a JSON array of TWO objects, each with keys \texttt{speaker} (= ``User''), \texttt{timestamp}, \texttt{clean\_text}, \texttt{image\_description}.
\end{tcolorbox}
\caption{Prompt template for the thematic user simulator (follow-up turn).}
\label{fig:prompt-thematic-user}
\end{figure*}

\begin{figure*}[t]
\begin{tcolorbox}[colback=gray!10,colframe=black,width=\linewidth,title=Thematic Assistant]
You are the ASSISTANT.

\textbf{Users said (JSON array of 2 user messages):} <user\_messages> \\
\textbf{Attached images:} <image\_desc> \\

Reply to each user message in order with TWO assistant messages. Return ONLY a JSON array of TWO objects, each with keys \texttt{speaker} (= ``Agent'') and \texttt{clean\_text}.
\end{tcolorbox}
\caption{Prompt template for the thematic assistant.}
\label{fig:prompt-thematic-agent}
\end{figure*}

\paragraph{Dialogue quality control.}
Once all rounds are generated, the dialogue is audited in two passes.
The \emph{Dialogue Self-Check (Per-Chunk)} template
(Figure~\ref{fig:prompt-dlg-chunk}) scans a contiguous slice of rounds
for persona violations, core-event drift, hallucinations, and
within-chunk contradictions; the \emph{Dialogue Self-Check
(Cross-Chunk)} template (Figure~\ref{fig:prompt-dlg-cross}) then
compares distant rounds to catch contradictions and invariant failures
that no single chunk reveals. Flagged rounds are repaired, and the
\emph{Round Summary} template (Figure~\ref{fig:prompt-summary})
provides the per-round summaries used both for cross-round memory
during generation and as the compact dialogue overview consumed by the
cross-chunk pass.

\begin{figure*}[t]
\begin{tcolorbox}[colback=gray!10,colframe=black,width=\linewidth,title=Dialogue Self-Check (Per-Chunk)]
You are auditing a contiguous slice of a generated dialogue against the persona and the core event. Return ONLY a JSON object.

\textbf{Persona:} <persona> \quad \textbf{Core event:} <core\_event> \\
\textbf{Timeline overview:} <timeline> \\
\textbf{Dialogue rounds in this chunk (each with a round id like \texttt{D1:7}):} <chunk> \\

Audit four categories, each reporting the offending round id:
\begin{itemize}\setlength\itemsep{0pt}
\item \texttt{persona\_violations} --- a turn contradicts the persona.
\item \texttt{core\_event\_violations} --- main-line rounds that fail to advance the core-event arc (off-topic rounds are fine).
\item \texttt{hallucinations} --- the assistant invents entities/data/numbers never grounded by persona, core event, timeline, or earlier rounds.
\item \texttt{internal\_contradictions} --- within this chunk, two rounds that contradict each other.
\end{itemize}
Output the four arrays plus \texttt{"ok"} (true iff all are empty).
\end{tcolorbox}
\caption{Prompt template for the per-chunk dialogue self-check.}
\label{fig:prompt-dlg-chunk}
\end{figure*}

\begin{figure*}[t]
\begin{tcolorbox}[colback=gray!10,colframe=black,width=\linewidth,title=Dialogue Self-Check (Cross-Chunk)]
You are doing a CROSS-CHUNK pass over a long generated dialogue, looking for contradictions that span chunks (not visible inside any single chunk). Return ONLY a JSON object.

\textbf{Persona:} <persona> \quad \textbf{Core event:} <core\_event> \\
\textbf{Per-chunk findings already collected:} <chunk\_findings> \\
\textbf{Dialogue overview (each round's user line summarised with its id):} <overview> \\

Flag problems where one round contradicts a much earlier or later round, or where a persona / core-event invariant fails when read end-to-end. Do NOT repeat per-chunk findings; report only NEW issues requiring distant comparison. Output \texttt{cross\_chunk\_contradictions}, \texttt{global\_persona\_violations}, \texttt{global\_core\_event\_violations}, and \texttt{"ok"}.
\end{tcolorbox}
\caption{Prompt template for the cross-chunk dialogue self-check.}
\label{fig:prompt-dlg-cross}
\end{figure*}

\begin{figure*}[t]
\begin{tcolorbox}[colback=gray!10,colframe=black,width=\linewidth,title=Round Summary]
Summarize the following USER+ASSISTANT exchange in 1--2 short sentences. Stay faithful to what was actually said; do not add facts or numbers not in the exchange. Paraphrase at the level of topics and numbers; do not rely on file names.

\textbf{Exchange:} <exchange> \\

Return ONLY a flat JSON object with exactly one string field \texttt{"summary"}.
\end{tcolorbox}
\caption{Prompt template for per-round summarisation (cross-round memory).}
\label{fig:prompt-summary}
\end{figure*}

\begin{figure*}[t]
\begin{tcolorbox}[colback=gray!10,colframe=black,width=\linewidth,title=LLM-as-a-Judge]
You are an impartial judge evaluating the memory capabilities of an AI assistant on a question-answering task. Compare the Assistant's Answer against the Ground Truth and assign a score of $0$, $0.25$, $0.5$, $0.75$, or $1$.

\textbf{Question:} <question> \quad \textbf{Gold Answer:} <gold> \quad \textbf{Model Answer:} <model\_answer> \\

\textbf{Scoring Rubric}
\begin{itemize}\setlength\itemsep{1pt}
\item \textbf{Score 0 (Incorrect / Miss):} the answer contradicts the Ground Truth; for Yes/No questions the polarity is wrong; for open-ended questions it is factually wrong or hallucinated; or it fails to provide the required information.
\item \textbf{Score 0.25 (Poor / Tangential):} touches the topic but misses the core entity or key value; mixes minor correct details with significant hallucinations; or is uselessly vague (e.g., ``a dog'' instead of ``a golden retriever'').
\item \textbf{Score 0.5 (Partial / Vague / Excessive):} technically correct but incomplete or unconfident; captures the main entity but misses required supporting details; or is over-informative. For Yes/No, the polarity is correct but reasoning is flawed or uncertain.
\item \textbf{Score 0.75 (Good / Minor Imperfection):} largely accurate and confident, missing only minor details, or correct but with unnecessary filler that reduces precision.
\item \textbf{Score 1 (Correct / Exact):} accurate, precise, and confident; for Yes/No the polarity matches exactly; for open-ended it contains all core information and necessary details without hallucinations.
\end{itemize}
Reply with ONLY the score digit. No other text.
\end{tcolorbox}
\caption{Prompt template for the LLM-as-a-Judge metric (five-level rubric).}
\label{fig:prompt-judge}
\end{figure*}

\paragraph{Retrieval-augmented generation.}
For all memory-based systems, at inference time the prompt is composed from the retrieved
information and any attached images or rendered PDF pages. The base
answering template (top of Figure~\ref{fig:prompt-answer}) requests a short, direct
answer and is shared by all question types. For find-matching
(\textsc{fm}) questions, where the gold answer is an image identifier
rather than free text, the template is extended with the additional
instruction shown below the rule, which constrains the model to choose
among the candidate file names actually attached from the retrieved
rounds, preventing it from inventing or emitting the placeholder
\texttt{img\_<number>}.

\begin{figure*}[t]
\begin{tcolorbox}[colback=gray!10,colframe=black,width=\linewidth,title=Retrieval-Augmented Generation]
You are answering a memory-QA question. Use ONLY the information below (and any attached images / rendered PDF pages).

\textbf{=== INFORMATION ===} \\
<information> \\
\textbf{=== END ===} \\

\textbf{Question:} <question> \\

Give a short, direct answer (a few words or a brief phrase). Do not add explanations or repeat the question.

\tcbline
\textit{\textbf{Additional instruction for find-matching (\textsc{fm}) questions:}}

The candidate images attached below come from the retrieved rounds. If the question asks for image file name(s), reply with one to three matching file names from this list (comma-separated): <candidate\_filenames>. Do not invent file names that are not in this list. Do not respond with the placeholder \texttt{img\_<number>}.
\end{tcolorbox}
\caption{Answering prompt used by \method{}. The base template (top) is
shared by all question types; the part below the rule is appended only
for find-matching-image (\textsc{fm}) questions.}
\label{fig:prompt-answer}
\end{figure*}
\section{Configuration Reference}
\label{sec:appendix-config}

All models are evaluated with deterministic generation whenever the endpoint supports it. The default settings are temperature 0.0, top-\(p=1.0\), and a maximum of 100 LLM calls per turn.

All target models are exposed to the benchmark through the same OpenAI-compatible function-calling scaffold. Local vLLM services use tensor parallelism with two to four A100 PCIe 80GB GPUs per model, bfloat16 inference, a 32,768-token maximum context unless explicitly overridden, prefix caching, chunked prefill, up to 32 concurrent sequences, and a maximum batched-token budget of 65,536. For local vLLM models, model-family-specific parsers and chat templates are used only inside the serving layer to translate model-native outputs into this common interface: Qwen uses the Qwen tool-call parser with thinking disabled, while Gemma and GLM use Gemma- and GLM-compatible chat templates and tool-call parsers.

\end{document}